\documentclass{article}
\usepackage{arxiv}

\usepackage[backend=bibtex, mincitenames=1, maxcitenames=2, style=ieee, citestyle=numeric, natbib=true, doi=true, isbn=false, url=false, eprint=true, date=year]{biblatex}
\AtEveryBibitem{\clearlist{language}} 
\AtEveryBibitem{\clearfield{note}} 
\AtEveryBibitem{\clearfield{primaryclass}} 
\usepackage{amsmath,amssymb,amsfonts} 
\usepackage[T1]{fontenc} 
\usepackage{authblk} 
\usepackage{xcolor} 
\usepackage[super]{nth} 
\usepackage{indentfirst} 
\usepackage{textcomp} 

\usepackage{algorithm}
\usepackage{algpseudocode}
\usepackage[english]{babel} 
\addto\captionsenglish{} 

\usepackage{siunitx} 
\DeclareSIUnit\rpm{RPM} 
\newcommand{\rads}{\radian\per\second} 

\usepackage[inline]{enumitem} 

\usepackage{graphics} 
\usepackage{graphicx} 
\usepackage{tikz} 
\usepackage{pgfplots} 
\usepackage{varwidth} 
\usepackage{layout} 
\usepackage[section]{placeins} 
\usepackage{float} 
\usepackage{subcaption} 
\usepackage{caption} 
\usepackage{lscape}

\usepackage{booktabs} 
\usepackage{tabularx,tabulary} 
\usepackage{multirow} 
\usepackage{adjustbox} 
\usepackage{array} 
\usepackage[flushleft]{threeparttable} 
\newcolumntype{L}[1]{>{\raggedright\let\newline\\\arraybackslash\hspace{0pt}}m{#1}} 
\newcolumntype{C}[1]{>{\centering\let\newline\\\arraybackslash\hspace{0pt}}m{#1}} 
\newcolumntype{R}[1]{>{\raggedleft\let\newline\\\arraybackslash\hspace{0pt}}m{#1}} 

\usepackage{hyperref} 
\usepackage[nameinlink,noabbrev,capitalise]{cleveref} 
\hypersetup{
  final=true,
  plainpages=false,
  pdfstartview=FitV,
  pdftoolbar=true,
  pdfmenubar=true,
  bookmarksopen=true,
  bookmarksnumbered=true,
  breaklinks=true,
  linktocpage,
  colorlinks=true,
  allcolors=black
}

\usepackage{needspace} 
\usepackage{bm} 
\usepackage{fancyhdr} 
\newcommand\blfootnote[1]{
  \begingroup
  \renewcommand\thefootnote{}\footnote{#1}%
  \addtocounter{footnote}{-1}%
  \endgroup
}
\usepackage{chngcntr} 
\usepackage{adjustbox} 
\DeclareCaptionLabelFormat{noprefix}{#2} 
\usepackage{soul} 

\newcommand{\treg}{\textsuperscript{\textregistered}}
\newcommand{\ttm}{\textsuperscript{\texttrademark}}

\usepackage{amsmath}
\usepackage{mhchem}
\usepackage{adjustbox}
\usepackage{multirow}
\usepackage{pbox}

\Crefformat{figure}{#2Fig.~#1#3}
\Crefmultiformat{figure}{Figs.~#2#1#3}{ and~#2#1#3}{, #2#1#3}{ and~#2#1#3}

\usepackage{setspace}
\renewcommand\sffamily{}

\title{Learning characteristic parameters and dynamics of centrifugal pumps under multiphase flow using physics-informed neural networks}
\author[1,2]{Felipe de Castro Teixeira Carvalho}
\author[3]{Kamaljyoti Nath}
\author[1,2]{Alberto Luiz Serpa\textsuperscript{*}}
\author[3,4]{George Em Karniadakis}

\affil[1]{Faculdade de Engenharia Mecânica (FEM), Universidade Estadual de Campinas (UNICAMP), Campinas, SP, Brazil}
\affil[2]{Centro de Estudos de Energia e Petróleo (CEPETRO), Universidade Estadual de Campinas (UNICAMP), Campinas, SP, Brazil}
\affil[3]{Division of Applied Mathematics, Brown University, Providence, RI, USA}
\affil[4]{School of Engineering, Brown University, Providence, RI, USA}

\begin{document}


\maketitle
 \thispagestyle{fancy}
 	\cfoot{}
 	\rhead{}
    \chead{}
    \lhead{}
 	\renewcommand{\headrulewidth}{0pt}
 	\rfoot{\large \today}
 	\blfootnote{\textsuperscript{*}Corresponding author: Alberto Luiz Serpa (alserpa@unicamp.br)\\
    \textit{E-mail addresses:} fdcastro@unicamp.br (Felipe de Castro Teixeira Carvalho), kamaljyoti\_nath@brown.edu (Kamaljyoti Nath), alserpa@unicamp.br (Alberto Luiz Serpa), george\_karniadakis@brown.edu (George Em Karniadakis)}

\begin{abstract}
  Electrical submersible pumps (ESPs) are prevalently utilized as artificial lift systems in the oil and gas industry. These pumps frequently encounter multiphase flows comprising a complex mixture of hydrocarbons, water, and sediments. Such mixtures lead to the formation of emulsions, characterized by an effective viscosity distinct from that of the individual phases. Traditional multiphase flow meters, employed to assess these conditions, are burdened by high operational costs and susceptibility to degradation. To this end, this study introduces a physics-informed neural network (PINN) model designed to indirectly estimate the fluid properties, dynamic states, and crucial parameters of an ESP system. A comprehensive structural and practical identifiability analysis was performed to delineate the subset of parameters that can be reliably estimated through the use of intake and discharge pressure measurements from the pump. The efficacy of the PINN model was validated by estimating the unknown states and parameters using these pressure measurements as input data. Furthermore, the performance of the PINN model was benchmarked against the particle filter method utilizing both simulated and experimental data across varying water content scenarios. The comparative analysis suggests that the PINN model holds significant potential as a viable alternative to conventional multiphase flow meters, offering a promising avenue for enhancing operational efficiency and reducing costs in ESP applications.
\end{abstract}


\section{Introduction} \label{sec:intro}

\noindent Multistage centrifugal pumps are essential across various industries, including water supply, chemical processes, power generation, and oil and gas. Among the different types of multistage centrifugal pumps, electrical submersible pumps (ESP) are the second most widely utilized artificial lift equipment in the oil and gas sector \cite{machado2019}, along with other industries. These pumps frequently have to handle multiphase flows consisting of gas, hydrocarbons, water, and sediment. A typical case of such flows is the two-phase liquid-liquid flow, where oil and water are transported simultaneously \cite{angeli2000}. These flows often lead to the formation of colloidal dispersion-like emulsions due to the chemical properties of the liquids involved. Notably, emulsions exhibit non-Newtonian behavior, which can lead to operational instabilities in ESP systems \cite{honorio2015,pastre2022}.

Additionally, another common multiphase flow condition is the two-phase gas-liquid flow. In this condition, the ESP performance may also experience unstable operation due to phenomena such as intermittent gas lock  \cite{gamboa2011}. Overall, the ESP system's unstable operation under multiphase flow conditions is undesirable and may cause early failures of the ESP \cite{pastre2022,zhou2010}. Thus, it is critical when operating ESP under multiphase flow conditions to avoid these unstable regions by changing the operational point of the ESP system. However, adjusting the ESP operation leveraging current models requires knowledge of fluid properties, phase fractions, and volumetric flow rates. Moreover, wax deposition in the ESP system pipelines changes their characteristics within the field operation \cite{martinez-palou2011}. 

Furthermore, as mentioned by \citet{falcone2010}, in the oil industry, obtaining information regarding the volumetric flow rates without the need for separation tanks could significantly enhance reservoir management, production allocation, and monitoring. Thus, this study explores a machine-learning approach that uses pressure measurements to monitor fluid properties, volumetric flow rate, and system properties. This method aims to facilitate the development of more effective monitoring and optimize oil production.

In the context of multiphase flows, multiphase flow meters (MPFMs) are designed to estimate volumetric flow rates, the fraction of each phase, and fluid properties. However, MPFMs are costly, necessitating direct intervention for repair, and are susceptible to degradation from factors such as sand erosion or partial blockages, which can compromise measurement accuracy \cite{falcone2010, bikmukhametov2020}. In this context, an alternative approach known as Virtual Flow Metering (VFM) has been developed, particularly within the oil and gas industry. VFMs are categorized into three types: physics-based, data-driven, and hybrid, which combine both approaches \cite{hotvedt2022b}.

Physics-based VFMs typically model the entire production line, including the reservoir, ESP, valves, pipelines, and fluid properties, using multiphase flow models like the two-fluid and the drift-flux model. Subsequently, an optimization problem is formulated to minimize the discrepancy between the measured values and those predicted by the model, a process commonly referred to as data reconciliation \cite{bikmukhametov2020}. In contrast to physics-based VFM, hybrid VFM often relies less on physics. For instance, \citet{hotvedt2020,hotvedt2022b} discuss enhancing valve models using field data combined with neural networks (NN). On the other hand, in data-driven VFM, several studies have been developed to estimate the volumetric flow rate of phases using different machine-learning techniques. Examples include the use of long short-term memory neural networks (LSTM) as demonstrated by \cite{andrianov2018}, as well as neural networks and regression trees (RT) in \cite{al-qutami2017}, and gradient boosting techniques in \cite{bikmukhametov2019}.

In the context of ESP models operating under two-phase flows, \citet{gamboa2011} conducted a detailed review of various models for gas-liquid flows available in the literature. The authors argued that a poor understanding of the underlying physical mechanisms of gas lock phenomena limits the accuracy of these mechanistic models. In the case of two-phase liquid-liquid flows, one of the primary challenges associated with ESPs is the non-Newtonian behavior of emulsions that form within the ESP system due to turbulence and shear stress induced by the ESP, valves, and fittings \cite{bulgarelli2021}. Moreover, the effective viscosity of these emulsions can significantly exceed that of oil or water in a single-phase state. This viscosity is influenced by factors such as the oil and water phases' individual viscosity, the oil-water mixture's water content, temperature, and the size and distribution of droplets \cite{plasencia2013}. Recently \citet{bulgarelli2021} proposed a model for relative viscosity inside the ESP that can be used to estimate the effective viscosity.

Researchers are continuously focusing on improving models for both effective viscosity and viscous flow inside the ESP. In this context, \citet{paternost2015} developed a correlation to estimate the pressure gain of ESPs operating under viscous conditions, while \citet{patil2018} provided a simplified methodology for predicting changes in centrifugal pump performance due to fluid viscosity variations. Similarly, \citet{zhu2020} proposed a mechanistic model for estimating pressure gain in ESPs under viscous fluid flow conditions. Furthermore, \citet{zhu2019a} introduced an ESP model for two-phase liquid-liquid flows based on the \citet{brinkman1952} model to account for the emulsion viscosity. Most recently, \citet{fctc2024} proposed a dynamic ESP system model that utilizes a bond-graph approach and incorporates the \citet{brinkman1952} model for effective viscosity in liquid-liquid flows, demonstrating reasonable agreement with experimental data.

The problem of estimating unknown properties based on indirect measurements is referred to as the inverse problem and is common in various fields. As discussed above, in the context of VFM, it is known as data reconciliation. However, when solving an inverse problem, one of the difficulties is the non-uniqueness of the solution, i.e., multiple values of the parameters may potentially provide satisfactory fits to the given data \cite{groetsch1993, aster2013}. We need to conduct an identifiability analysis of the system to determine whether a set of parameters or states can be accurately estimated from a given set of inputs and states. This analysis can be either structural or practical. Structural identifiability analysis assumes an ideal scenario where the observed states are noise-free, and the model is error-free. This analysis, also known as prior identifiability, can be performed without actual experimental data \cite{bellman1970,ljung1994,tuncer2018}.

Various methods have been developed to analyze both local and global structural identifiability of systems. \citet{chis2011} and \citet{raue2014} conducted comparative studies of these methods, mainly focusing on biological systems. \citet{castro2020} introduced a simple scaling method that exploits the invariance of equations under parameter scaling transformations. More recently, \citet{dong2022} proposed a method based on differential elimination. In the present study, we consider the method proposed by \citet{dong2022} for structural identifiability analysis to determine the set of identifiable parameters.

In addressing the inverse problem within the context of Ordinary Differential Equations (ODEs), the Nonlinear Least Squares (NLS) method is frequently utilized due to its versatility across various ODE systems. However, NLS is computationally demanding, and inaccuracies in the numerical approximations of derivatives can pose significant challenges, especially in stiff systems \cite{ramsay2007, qi2010, bradley2021}. Beyond NLS, alternative methods include collocation techniques \cite{ramsay2007}, Gaussian process-based approaches \cite{dondelinger2013}, Bayesian methods \cite{putter2002}, and a novel two-stage approach using Neural ODEs \cite{bradley2021}. Additionally, for state estimation, the Kalman filter is widely used in linear systems, with variations such as the Extended Kalman filter (EKF) and the Unscented Kalman filter (UKF) addressing nonlinear systems. These filters can also be applied to state and parameter estimation \cite{simon2006}. In cases of high non-linearity, the particle filter (PF) offers an effective alternative to Kalman filters \cite{simon2006, ristic2004}.

\citet{raissi2019} introduced Physics-Informed Neural Networks (PINNs). This methodology involves training neural networks by minimizing the residuals of differential equations within a multi-loss function (data and physics) framework, enabling the solution of both forward and inverse problems. The approach has been extended to various variants, including Conservative PINNs (cPINNs) \cite{Jagtap_2020_CPINN}, Extended PINN (XPINN) \cite{Jagtap_2020_XPINN}, hp-VPINNs \cite{Kharazmi2021}, Parareal PINN (PPINN) \cite{Meng_2020}, and Separable PINN \cite{cho_2022_separable}. These variants have been applied to complex inverse problems such as unsaturated groundwater flow \cite{Depina_2022_unsaturated}, diesel engine optimization \cite{Nath_2023_Physics}, and supersonic flows \cite{Jagtap_20221_supersonic_flows}. Additionally, \citet{mcclenny2022selfadaptive} proposed a self-adaptive weights technique to adjust the weights associated with the different PINN losses automatically.

As previously discussed, despite recent advancements in MPFMs these meters incur high operational costs and suffer from accuracy degradation over time. In contrast, physics-based VFMs have emerged as a cost-effective alternative, capable of estimating fluid density, phase fractions, and volumetric flow rates. However, these VFMs depend on detailed descriptions of the ESP system and complex and computationally intensive multiphase flow models. On the other hand, a common strategy to enhance the tractability of complex systems involves employing a lumped-element model, such as the model developed by \citet{fctc2024} for ESP systems.

Nonetheless, these lumped-element models still contain several unknown parameters, including the fluid bulk modulus, the emulsion's effective viscosity, and the pipeline's variable equivalent resistance, which may change over the operational lifespan of the well due to factors such as wax deposition. In this context, the PINNs offer a promising alternative for estimating unknown states and parameters. The PINNs can integrate data with models, which is advantageous for addressing the simplifications in the lumped-element model of the ESP system. Consequently, combining PINNs and a lumped-element model could facilitate the development of a less computationally intensive, physics-based VFM. This approach allows for monitoring fluid properties, system parameters, and states.

In this paper, we investigate the application of PINNs for estimating key characteristic parameters (such as flow rate, fluid properties, and pump parameters) and predicting the dynamics of state variables using pressure measurements at the inlet and outlet of an ESP. This analysis is based on the model proposed by \citet{fctc2024}. We identified sets of unknown parameters through structural and practical identifiability analysis conducted on the \citet{fctc2024} model. Our study utilizes simulated and experimental datasets of two-phase liquid-liquid flows with different water fractions. Therefore, the key contributions of this paper can be briefly summarized as: \begin{enumerate*}[label=(\roman*)]
  \item Conducting structural and practical identifiability analysis on the lumped-element model of the ESP system;
  \item Applying PINNs for estimating key parameters and dynamic states of the ESP system, including fluid properties and pump parameters;
  \item Comparative analysis of PINNs and PF, focusing on parameter estimation;
  \item Implementing a combination of PINNs and a lumped-element model to serve as a cost-effective VFM;
  \item Experimental validation of the PINN model using both simulated and experimental datasets.
\end{enumerate*}

The remainder of this research paper is organized as follows: In \cref{sec:methodology}, we discuss the methodology, we provide an overview of the experimental setup (\S \ref{subsec:exp}), the dynamic model of the ESP system (\S \ref{subsec:esp-model}), the structure of the PINN model (\S \ref{subsec:pinn}), and the details regarding data acquisition and generation (\S \ref{subsec:data-generation}). We also discuss the structural and practical identifiability analysis within the section (\S \ref{subsec:structiden} and \S \ref{subsec:practiden}, respectively). In \cref{sec:results_iden}, we present the results obtained from the structural local and practical identifiability analyses. Subsequently, in \cref{sec:results_pinn}, we present the results of the PINN and PF. Finally, in \cref{sec:conclusion}, we present the conclusions drawn from this study, summarizing the key findings and discussing the results. A brief discussion on the particle filter method is included in \cref{subsec:pf}.

\section{Methodology}\label{sec:methodology}

\subsection{Experimental study}\label{subsec:exp}

\subsubsection{Experimental setup}\label{subsec:exp-setup}
\begin{figure}[!htb]
 \centering
 \includegraphics{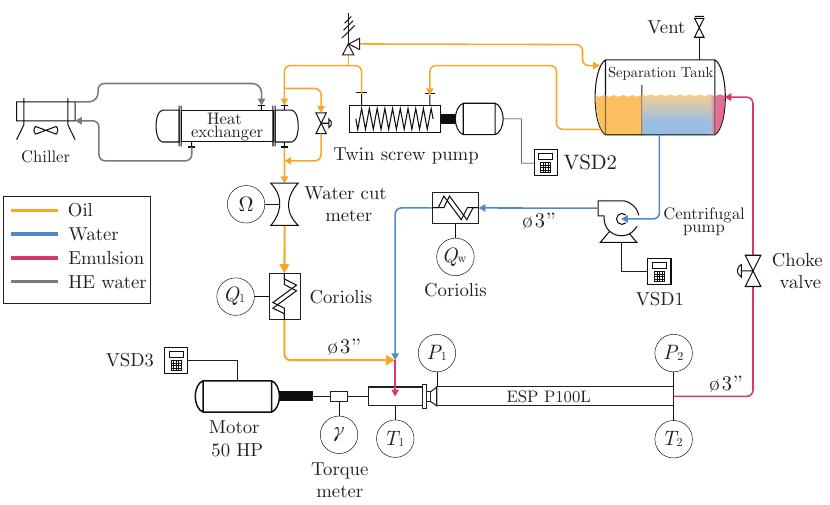}
 \caption{\textbf{Schematic diagram of the experimental setup:} A schematic diagram of the experimental setup indicates all the piping system's components and dimensions. It has four distinct flow lines, an oil flow line, a water flow line, a two-phase flow line carrying the oil-water mixture, and a closed loop of water used in the heat exchanger (HE). The separation tank is open to the atmosphere and separates the oil and water phases from the emulsion by gravity. It also stores the oil and water phases. The fluids are pumped independently from the separator tank and operate in a closed loop. The twin-screw pump pumps the oil phase from the separator tank. Then, the oil phase flows through a shell and tube heat exchanger, a water fraction meter (measures the water fraction inside the oil phase), and a Coriolis meter (measures the mass flow rate and density). A single-stage centrifugal pump pumps the water phase from the separator tank that flows through a Coriolis meter and a remotely controlled valve. Then, the oil and water phases mix in a ``T'' joint, forming an emulsion. Then, the emulsion is pumped by the ESP, flows through a remotely controlled valve, and returns to the separation tank.}
 \label{fig:explayout}
\end{figure}

\noindent The experimental setup was designed to investigate the ESP performance under various flow conditions, including single-phase and water/oil two-phase flows. The main equipment is an eight-stage ESP model P100L 538 series from Baker Hughes\textsuperscript{\textregistered}, located downstream of the oil and water mixture point. A schematic diagram of the experimental setup is shown in \cref{fig:explayout}, indicating all the components and dimensions of the piping system.

The ESP impellers have mixed flow geometry with an outer diameter of \qty{108}{\milli\metre} and have \num{7} blades. Pressure gauges and temperature transmitters were installed at the ESP inlet and outlet. They measure the suction and discharge pressure and temperature, respectively. The pressure transducers are capacitive and manufactured by Emerson Rosemount\ttm, series 2088. The temperature transducers are four-wire resistance temperature detectors type PT100 with 1/10 DIN accuracy. The pump shaft torque and rotational speed are measured using the sensor model T21WN manufactured by HBM\treg. Each pump on the experimental setup is driven by an electric three-phase induction motor supplied by a Variable Speed Driver (VSD) on each motor.

The fluids used in the experiments are water and a blend of mineral oil. The oil viscosity was characterized by a rotational rheometer model HAAKE MARS III (see \cref{apdx:oil_visc} for the viscosity measurements in different temperatures). The water fraction inside the oil phase is measured by the water fraction meter model Nemko 05 ATEX 112 manufactured by Roxar\treg. The Coriolis meters used to measure the density and the mass flow rates of the oil and water flow lines are model F300S355 manufactured by Micro Motion\treg.

As the system operates in a closed loop, it tends to heat up and change the viscosity. Thus, the heat exchanger controls the fluid's temperature and the viscosity inside the flow line. The temperature is regulated by monitoring the ESP inlet temperature ($T_1$). This temperature measurement is also used to calculate the working fluid viscosity. The heat exchanger (HE) has an independent water flow line comprised of a chiller, a heater, and a water tank, allowing it to cool and heat the fluid. Also, as the oil viscosity strongly depends on the temperature, it is possible to conduct tests with different viscosities with the same oil by adjusting the fluid's temperature.

The operational ranges and uncertainties of the instruments are presented in \cref{tab:sensor_spec}. The pumps, motors, variable speed drive, valve, tank, and heat control system specifications are presented in \cref{tab:device_spec}.

\subsubsection{Experimental procedure}\label{subsec:exp-proc}
\noindent In this section, we detail the experimental procedure utilized to evaluate the dynamic behavior of the ESP system when varying the ESP shaft angular velocity while operating under a two-phase flow of oil and water. For simplicity, when referring to ESP angular velocity, it refers to the ESP shaft angular velocity. The procedure is divided into two stages; the first stage is the start-up stage, and the objective of this stage is to obtain a stable water fraction. The second stage is the transient acquisition stage, in which we collect the experiment data.

\begin{enumerate}[label=\textbf{Stage \arabic*:}, leftmargin=*]
 \item Start-up:
 \begin{enumerate}[leftmargin=0.0cm]
 \item Begin by slowly starting the oil pump and water pump at a low speed (e.g., \qty{62}{\rads}) to ensure system stability and prevent sudden pressure surges.
 \item Gradually increase the rotation speed of the ESP while monitoring the suction pressure. Maintain the ESP angular velocity below \qty{105}{\rads} and ensure the suction pressure remains above \qty{100}{\kPa} to prevent cavitation in the ESP first stages.
 \item \label{enum:wc_def} Check the measured water fraction and adjust the water pump angular velocity if necessary. Repeat this step until the desired water fraction is achieved.
 \item Stop the water pump.
 \end{enumerate}
 \item Transient acquisition:
 \begin{enumerate}[leftmargin=0.0cm]
 \item Increase the ESP angular velocity to the desired initial value.
 \item Increase the oil pump angular velocity while closing the ESP downstream valve to maintain the suction pressure within the range of \qtyrange{100}{600}{\kPa}. The upstream pressure of the ESP should not exceed \qty{600}{\kPa} for a safe operation.
 \item Gradually increase the ESP angular velocity to reach the desired final value.
 \item Increase the oil pump angular velocity while closing the downstream valve to maintain the suction pressure within the range of \qtyrange{100}{600}{\kPa}.
 \item Return the ESP angular velocity to the desired initial speed.
 \item \label{enum:stab} Allow sufficient time for the suction, discharge pressure, and volumetric flow rate to stabilize. This stabilization period allows capturing only the dynamics of the ESP angular velocity change.
 \item Start data acquisition.
 \item Wait \qty{5}{\second}.
 \item Adjust the ESP angular velocity on the VSD to the final desired value.
 \item Allow another stabilization period for the suction, discharge pressure, and volumetric flow rate. This period allows the capture of the complete dynamic system response to the rotation change.
 \item \label{enum:end} End data acquisition.
 \item Return to the \ref{enum:wc_def}
 \end{enumerate}
\end{enumerate}

We considered the system stable when the coefficient of variation of the last \qty{20}{\second} of measurements of the suction and discharge pressure and oil volumetric flow rate was smaller than \qty{0.8}{\percent}. These criteria are the same as used and described by \citet{figueiredo2020}. Additionally, for the temperature, we considered a tolerance of \qty{\pm 0.25}{\celsius} during the acquisition of the experimental time. The Supervisory Control and Data Acquisition (SCADA) system automatically monitors for the system stabilization and performs the steps \ref{enum:stab} to \ref{enum:end} automatically.
\subsubsection{Experimental investigation}\label{subsec:ps-var}
\noindent Our experimental investigations focused on two different water-in-oil emulsion conditions. We varied parameters such as water fraction, temperature, and angular velocity step size. The specific values used in each experiment are presented in \cref{tab:test-matrix}. By employing this approach, we were able to evaluate the system dynamics across different density and viscosity conditions while also assessing the influence of emulsion formation on the ESP.
%
\begin{table}[H]
  \sisetup{round-mode = places, round-precision = 1}
  \centering
  \caption{\textbf{Test matrix for assessing the dynamics of the ESP system in oil/water two-phase flows.} The table includes the experimental parameters varied in the study, such as the initial angular velocity ($\omega_i$), final angular velocity ($\omega_f$), water cut ($\Omega$), temperature, and twin-screw pump angular velocity ($\omega_t$).}
    \begin{tabular}{cccccc} 
    \toprule
    Investigation (\#) & $\omega_i$ (\qty{}{\rads}) & $\omega_f$ (\qty{}{\rads}) & $\Omega$ (\qty{}{\percent})      & Temperature (\qty{}{\celsius}) & $\omega_t$ (RPM)   \\ \midrule
    1 \label{testmatrix:1}              & \num{272.27}         & \num{314.16}       & \num{14}         & \num{33.5} & \num{1280} \\
    2 \label{testmatrix:2}              & \num{251.32}         & \num{314.16}        & \num{5}         & \num{26}   & \num{1160} \\ \bottomrule
  \end{tabular}
  \label{tab:test-matrix}
\end{table}


\subsubsection{Data collection}\label{subsec:daq}
\noindent We developed the SCADA system and implemented it using LabVIEW\ttm software. It gathers real-time data to monitor and control the ESP. We used a data acquisition system from the National Instruments model cDAQ9188 to read and generate analog signals. The generated analog signals are sent to the pumps' VSDs and valve actuators. We defined a sampling rate of \qty{250}{\hertz} for the signal acquisition. The signals from different measuring instruments are collected simultaneously to ensure synchronized data acquisition.

An anti-aliasing filter was employed as a post-processing step on the acquired signals to remove high-frequency noise and ensure accurate signal acquisition. This low-pass filter is an 8$^{\text{th}}$ order Butterworth filter with a cut-off frequency of \qty{10}{\hertz}. This frequency was chosen to comply with the Nyquist-Shannon sampling theorem for the maximum frequency specified by the pressure transmitter manufacturer. Then, the signal was downsampled to the same frequency.

\subsubsection{Limitations of the experimental study}\label{subsec:limitations}

\noindent The experiments were conducted under a laboratory-scale ESP system with a single ESP in the flow loop, while oil extraction production lines typically involve multiple wells with ESPs operating in parallel with significantly higher pressure and volumetric flow rates. Furthermore, the experimental setup was constrained to a pipeline length of approximately \qty{60}{\meter}, whereas actual flow lines and risers in field conditions can extend over one kilometer. Therefore, this experimental setup does not assess the interaction of multiple ESPs operating in parallel in long pipelines. We considered two different water fraction conditions. However, these are only water-in-oil emulsions. On the other hand, oil-in-water emulsions may present different dynamic behaviors due to considerably different emulsion viscosity, which has not been studied in the present study.

\subsection{Electrical submersible pump system dynamic model}\label{subsec:esp-model}
\noindent In this section, we introduce the dynamic model of the ESP system, which is utilized in conjunction with the PINN to estimate unknown states and parameters. We adopt the model proposed by \citet{fctc2024}, which encompasses both the dynamics of the pipeline and the ESP. The model was validated with varying water fraction conditions, ESP rotations, and different ESP models. It specifically addresses the mechanical and hydraulic domains, explicitly excluding the electrical motor from consideration. Consequently, the shaft torque is the input to the ESP system model proposed by \citet{fctc2024}. The state variable describes the ESP system model,
\begin{equation}
    \bm{X} = \{Q_p, \omega, Q_1, Q_2, P_1, P_2 \},
\end{equation}
where $Q_p$ is the pump volumetric flow rate, $\omega$ is the shaft angular velocity, $Q_1$ and $Q_2$ are the volumetric flow rate in the upstream and downstream pipeline respectively, $P_1$ and $P_2$ are the intake and discharge pressure respectively. The governing equations describing the system's dynamics depend on the input vector $\gamma(t)$. These are of the form
\begin{equation}
    \bm{\dot{X}} = F(\bm{X}, \gamma(t)).
\end{equation}
\par The ESP system is governed by the following set of ODEs:
\begin{subequations}
  \begin{align} \label{eq:esp-model1}
    & \frac{dQ_p}{dt} = \frac{(P_1 - P_2 + k_3 \, \mu \, Q_p)\,A_p}{\rho \, L_p} + \frac{A_p\,(k_{1p} \, \omega \, Q_p + k_{2p} \, \omega^2 + k_{4p} \, {Q_p}^2)}{L_p},\\
    \label{eq:esp-model2}
    & \frac{d\omega}{dt} = \frac{\gamma(t) - k_{1s}\,\rho\,{Q_p}^2 - k_{2s}\,\rho\,\omega\,Q_p - k_{3s}\,\mu\,\omega - k_{4s}\,\omega - k_{5s}\,\omega^2}{I_s},\\
    \label{eq:esp-model3}
    & \frac{dQ_1}{dt} = \frac{(k_{bd}\omega_t - Q_1)\,k_{bl}\,\mu + P_{in} - P_1 - f_f(Q_1, \mu, L_u, d_u)\, {Q_1}^2\,A_u}{\rho\,L_u} - \frac{k_u\,{Q_1}^2}{2\,L_u\,A_u},\\
    \label{eq:esp-model4}
    & \frac{dQ_2}{dt} = \frac{(P_2 - P_{out} - f_f(Q_2, \mu, L_d, d_d)\, {Q_2}^2)\,A_d}{\rho\,L_d} - \frac{k_d\,{Q_2}^2}{2\,L_d\,A_d} - \frac{{Q_2}^2\, A_d}{L_d\,C_v(a)^2\,\rho_0},\\
    \label{eq:esp-model5}
    & \frac{dP_1}{dt} = \frac{(Q_1 - Q_p) \, B}{A_u \, L_u},\\
    \label{eq:esp-model6}
    & \frac{dP_2}{dt} = \frac{(Q_p - Q_2) \, B}{A_d \, L_d},
  \end{align} \label{eq:esp-model}
\end{subequations}
\noindent where the parameters are described in detail in \cref{apdx:model}. The values of these parameters were established by \citet{fctc2024}, and in this study, we utilize them without modification. The function $C_v(a)$ represents the valve coefficient as a function of the valve aperture ($a$), and \(\rho_0\) is the water density at \qty{15}{\celsius}. For simplicity, as we are not varying the valve aperture, we considered a single equivalent resistance coefficient obtained experimentally.

\newcommand{\numberofps}{26}
\newcommand{\numberofstates}{6}

Considering the high viscosity of the emulsion, we assumed a laminar flow in the emulsion and oil flow lines. Therefore, the friction function $f_f(Q, \mu, L, d)$ is derived using the Darcy-Weisbach equation and the friction factor for laminar flow, resulting in the following equation:
\begin{equation}
f_f(Q, \mu, L, d) = \frac{128 \, L \, \mu}{\pi \, Q \, d^{4}}.
\end{equation}
The ESP system model takes a single input variable, $\gamma(t)$, representing the torque applied to the ESP shaft. In the experimental setup, the intake pressure of the twin-screw pump $P_{in}$ and the pressure at the end of the pipeline $P_{out}$ refers to the same separation tank that is open to the atmosphere. Therefore, for the sake of simplicity, these pressures are assumed to be constant and are set to \qty{0}{\pascal}.

It is important to mention that in both experimental and real-world conditions, various factors, such as turbulence and shear generated by the ESP, valve, and other devices installed in the flow line, contribute to emulsion formation, leading to changes in emulsion viscosity. Thus, the viscosity of the emulsion varies as the fluids traverse the system due to these effects. Additionally, the emulsion formation is influenced by the emulsion stability, chemical properties of the fluid, and the presence or addition of emulsifiers and surfactants. Moreover, in practical scenarios, temperature gradients along the pipeline significantly impact viscosity. In such cases, the assumption of a single viscosity is no longer valid, and appropriate assumptions must be made to account for these complexities.

In our specific case, the ESP operates within a closed-loop system. Consequently, the same emulsion continuously flows through the ESP and valves while the fluid temperature is controlled through a heat exchanger. Given the tank volume, the operating volumetric flow rate, temperature control measures, and the transient experiment duration, we assumed that viscosity remains approximately constant.

Although the \citet{brinkman1952} model is widely used in the oil industry for emulsion effective viscosity, numerous alternative models for emulsion viscosity have been proposed (e.g., \citet{taylor1932} and \citet{pal1989}). Recently, \citet{bulgarelli2021} introduced a model specifically for effective emulsion viscosity inside ESPs. However, this model exclusively applies to ESPs and does not hold for pipelines. Thus, in this study, we employed the \citet{brinkman1952} model to determine the emulsion viscosity for the entire system. It is expressed as:
\begin{equation}\label{eq:brinkman}
\mu = \mu_c \left(\frac{1}{1 - \Omega} \right)^{2.5} \quad \quad 0<\Omega<1 ,
\end{equation}
where $\Omega$ represents the water fraction and $\mu_c$ is the viscosity of the continuous phase

\subsection{Physics-informed neural networks}\label{subsec:pinn}

\begin{figure}[!htb]
  \centering
  \includegraphics{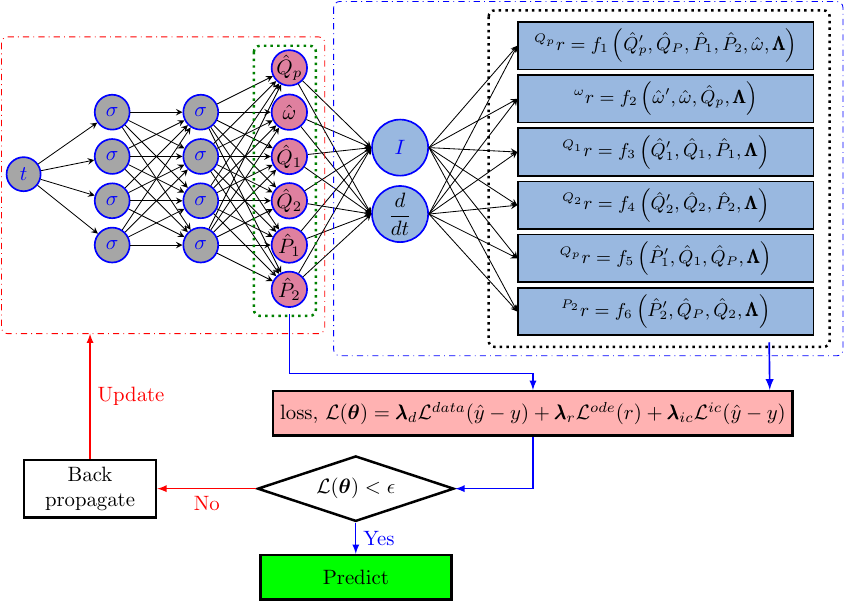}
  \caption{\textbf{Schematic representation of the Physics-informed neural network for the ESP system:} The deep neural network, shown in the red dashed-dotted rectangle on the left, is considered to approximate the solution of the ODEs system described in \cref{subsec:esp-model} (\crefrange{eq:esp-model1}{eq:esp-model6}). The input to the neural network is time, denoted by $t$, and the output is the ESP system states, highlighted in the green dotted rectangle of the figure. Each differential equation of the ESP system has residues at certain collocation points that must be minimized. We indicate them in the black dotted region on the right. The time derivative of each state (DNN output) is computed using automatic differentiation. The total loss, denoted as $\mathcal{L}(\bm{\theta})$, includes data loss, physics (ODE) loss, and initial condition loss. The data loss ($\mathcal{L}^{data}(\hat{y}-y))$ is the loss between the DNN output and measured data, the physics (ODE) loss is the loss of the residue of the equations at the collocation points, and the initial condition loss is a loss between the initial condition and the DNN output at $t=0$. $\bm{\lambda}_d$, $\bm{\lambda}_r$, and $\bm{\lambda}_{ic}$ are the weights to the data loss, physics loss, and initial condition loss, respectively, while calculating the total loss. These may be fixed or adaptive. $\bm{\Lambda}$ are the unknown parameters in the ODE system. The activation function of the DNN is represented by $\sigma$.}
  \label{fig:pinn}
\end{figure}

The physics-informed neural network (PINN) comprises a neural network (NN) used to approximate the solution of a given differential equation (DE). It is comprised of two losses: a PINN loss and a data loss. The data loss is a function of the difference between the labeled and predicted output. It may include the initial and boundary condition loss and any other known data loss for an inverse problem. The PINN loss is comprised of the model residue at the domain. The objective of PINN is to minimize both the data and PINN loss in the entire domain. In the context of inverse problems, the unknown parameters are also considered trainable parameters along with the NN parameters. The computation of loss gradients with respect to the NN and DE parameters, as well as the time derivative in the context of an ODE system, are facilitated by automatic differentiation \cite{Baydin_2018}. For readers interested in a more detailed discussion, we refer them to \citet{raissi2019,karniadakis2021}. The schematic diagram of the PINN considered in this study is presented in \cref{fig:pinn}.

\subsubsection{PINN for ESP system}\label{subsec:pinn:arch}

\noindent The architecture of the neural network used in this study is a fully connected feed-forward deep neural network (DNN) to approximate the state variables (i.e. $Q_p$, $\omega$, $Q_1$, $Q_2$, $P_1$, $P_2$). The DNN architecture comprises one input layer, $h$ hidden layers, and one linear output layer and it is structured as follows:
\begin{subequations}
  \begin{align}
      \bm{y}_0 & = t \hspace{6.35cm} && \text{Input}\\
      \bm{y}_i & = \sigma\left(\bm{y}_{i-1}\bm{W}_i + \bm{b}_i\right) \hspace{1cm} i\;\forall \;1\le i\le h - 1 \hspace{1cm} && \text{Hidden layers}\\
      \bm{\hat{y}} = \bm{y}_h & = \bm{y}_{h-1}\bm{W}_h + \bm{b}_{h-1} && \text{Output layer}
  \end{align}
\end{subequations}
where $\bm{W} \in \mathbb{R}^{n \times m}$ and $\bm{b} \in \mathbb{R}^{1 \times m}$ are the weights and biases of the neural network. The $n$ and $m$ are the sizes of the previous and current layers, respectively. The function $\sigma(.)$ is the activation function, which in this study we consider as a $\text{tanh}(.)$ function. For the remainder of this work, the weights and biases are referred to collectively as parameters of the neural network $\bm{\theta} = \{\bm{W}, \bm{b}\}$.

The optimization of the PINN model is driven by a composite loss function, which comprises by three different losses: the physics loss ($\mathcal{L}^{ode}(\bm{\theta}, \bm{\Lambda}, \bm{\lambda}_r$)), the data loss ($\mathcal{L}^{data}(\bm{\theta}, \bm{\lambda}_d)$) and the initial condition loss ($\mathcal{L}^{ic}(\bm{\theta}, \bm{\lambda}_{ic}$)). The total loss is given as:
\begin{equation} 
  \mathcal{L}(\bm{\theta}, \bm{\Lambda}, \bm{\lambda}_d, \bm{\lambda}_r, \bm{\lambda}_{ic}) = \mathcal{L}^{ode}(\bm{\theta}, \bm{\Lambda}, \bm{\lambda}_r) + \mathcal{L}^{data}(\bm{\theta}, \bm{\lambda}_d) + \mathcal{L}^{ic}(\bm{\theta}, \bm{\lambda}_{ic}).
  \label{eq:total loss}
\end{equation}
The physics loss is given by the weighted sum (self-adaptive weights) of physics loss due to individual equations,
\begin{subequations}
    \begin{align}
    \mathcal{L}^{ode}(\bm{\theta}, \bm{\Lambda}, \bm{\lambda}_r) & = \sum_{s \in \Phi} m(\lambda_r^s) \mathcal{L}^{ode}_s, \qquad \Phi =\{Q_p,\; \omega, \; Q_1, \; Q_2, \; P_1, \; P_2 \}, \\
    & = \sum_{s\in\Phi} m(\lambda_r^s) \left[ \frac{1} { {^s}N^{ode}} \sum_{i=1}^{{^s}N^{ode}} {{^s}r_{i}}^2 \right],\\
    & = \sum_{s\in\Phi} m(\lambda_r^s) \left[ \frac{1}{{^s}N^{ode}} \sum_{i=1}^{{^s}N^{ode}} \left(\frac{d\hat{y_s}}{dt}\middle|_{t_i}  - f_s \left(\hat{y}(t_i; \bm{\theta}), \gamma(t_i); \bm{\Lambda} \right) \right)^2 \right],
    \end{align}
    \label{eq:loss:ode}
\end{subequations}

\noindent where $\mathcal{L}^{ode}_s$, $s\in\Phi$,  $\Phi =\{Q_p,\; \omega, \; Q_1, \; Q_2, \; P_1, \; P_2 \}$ are the physic-informed loss for the differential equation describing the dynamics of the state variables (\crefrange{eq:esp-model1}{eq:esp-model6}). $^sN^{ode}$ is the number of residual points for each of the equations, $\lambda_r^s$ are the self-adaptive weights associated with each physics loss, and $m(.)$ is a mask function which is considered as softplus function in the present study. $^sr_n$ are the residual of the equation. We discuss the calculation of the residual in \cref{apdx:loss-ode}. In the present study, the residual points for all the equations are considered the same. Thus, all $^sN^{ode}$ has the same points.

Similarly, the data loss is given by the weighted sum (self-adaptive weights) of data loss due to $P_1$ and $P_2$.
\begin{subequations}
    \begin{align}
        \mathcal{L}^{data}(\bm{\theta}, \bm{\lambda}_d) & = \sum_{s \in \phi} m(\lambda_d) \mathcal{L}^{data}_s, \qquad \phi =\{P_1, \; P_2 \}\\
        & = \sum_{s\in\phi} m(\lambda_d) \left[ \frac{1}{^sN^{data}} \sum_{i=1}^{^sN^{data}} \left[y_s(t_i) - \hat{y}_s(t_i; \bm{\theta})\right]^2 \right],
    \end{align}
\end{subequations}
where $^sN^{data}$ is the number of points at which the measured data are available for each of the known quantities, $y_s(t_i)$ is the known value of the $s$ quantity at time $t_i$ and the corresponding approximated value from the neural network is $\hat{y}_s(t_i;\bm{\theta})$. Similar to physics loss, $\lambda_d^s$ is the self-adaptive weight associated with each data loss, and $m(.)$ is a mask function taken as the softplus function in the present study.

The initial condition loss
\begin{subequations}
    \begin{align}
        \mathcal{L}^{ic}(\bm{\theta}, \bm{\lambda}_{ic}) & = \sum_{s\in\Phi} m(\lambda_{ic}^s) \mathcal{L}^{ic}_s, \qquad \Phi =\{Q_p,\; \omega, \; Q_1, \; Q_2, \; P_1, \; P_2 \} \\
        & = \sum_{s\in\Phi} m(\lambda_{ic}^s) \left[\left(y_s(t_0) - \hat{y}_c(t_0; \bm{\theta})\right)^2 \right],
    \end{align}
\end{subequations}
where $y_s(t_0)$ is the known value of the $s$ quantity at time zero, and the corresponding approximated value from the neural network is $\hat{y}_s(t_0;\bm{\theta})$. $m(.)$ is a mask function taken as the softplus function in the present study.

\subsubsection{Input and output scaling}

The neural network architecture considered in the present study is  $[1, 20, 20, 20, 6]$, where the input layer consists of one neuron that takes time $t$ as input and is scaled between $[-1,1]$. In the output layer, we have six neurons that approximate the six state variables discussed in \cref{subsec:esp-model} (\cref{eq:esp-model1} to \cref{eq:esp-model6}). It is important to note that these state variables differ significantly in their scales, which could adversely affect the neural network's approximation capabilities. To address this issue, the outputs of each neural network undergo a transformation to standardize the magnitude of the different state variables into similar values. The form of this transformation varies according to the specific requirements of the problem. In this study, we employ the following transformation:
\begin{equation}\label{eq:nn:transform}
  f(x) = \frac{(x + 1)(x_{\text{max}} - x_{\text{min}})}{2} + x_{\text{min}},
\end{equation}
where $f(x)$ represents the state variable in the physical domain, and $x$ is the output from the neural network. For known states (i.e., $P_1$ and $P_2$), $x_{\text{min}}$ and $x_{\text{max}}$ are the minimum and maximum measured values, respectively.

However, for the unknown states, the $x_{\text{min}}$ and $ x_{\text{max}}$ are estimated by solving a system of equations at two specific time points $t_1$ and $t_2$. The system of equations, derived from simplifications of \cref{eq:esp-model1,eq:esp-model2}, is given by:
\begin{equation}\label{eq:sysinit}
  \begin{cases}
    k_{1p}\,\rho\,\omega(t_i)\,Q_p(t_i) + k_{2p}\,\rho\,\omega(t_i)^2 + k_{4p}\,\rho Q_p(t_i)^2 - (P_2(t_i) - P_1(t_i)) = 0, \\
    k_{1s}\,\rho\,Q_p(t_i)^2 - k_{2s}\,\rho\,\omega(t_i)\,Q_p(t_i) + \gamma(t_i) = 0, \quad \quad  t_i \in \{t_1, t_2\}
  \end{cases}
\end{equation}
where $k_{1p}$, $k_{2p}$, $k_{4p}$, $k_{1s}$, and $k_{2s}$ are the ESP system parameters, $\rho$ represents the fluid density, and $t_i$ denotes the time instant for the state variables ($\omega$, $Q_p$, $P_1$, $P_2$) and system input ($\gamma$). Furthermore, the time points $t_1$ and $t_2$ are determined based on the pressure difference between $P_2(t)$ and $P_1(t)$ as follows:
\begin{equation}
  t_{1} = \arg\min_{t} \left( P_2(t) - P_1(t) \right), \quad t_{2} = \arg\max_{t} \left( P_2(t) - P_1(t) \right).
\end{equation}
Solving the system of equations at the time point $t_1$ provides the estimation for $x_{\text{min}}$, and at the time point $t_2$ for $x_{\text{max}}$. We would also like to mention that to solve the above \cref{eq:sysinit}, we need to know the parameters ($k_{1p}$, $k_{2p}$, $k_{4p}$, $k_{1s}$ and $k_{2s}$)  of the equation, which are unknown in this case. Thus, we assumed some initial value in the parameters. In this study, while solving the \cref{eq:sysinit}, we considered $\rho = \qty{931.51}{\kilogram\per\cubic\meter}$, which represents a water fraction of \qty{50}{\percent}, as an initial guess, and for the remaining pump parameters we considered a value of $+15\%$ of the true value for all parameters. Additionally, we considered the same values obtained for $Q_p$ for the $Q_1$ and $Q_2$ states (pipeline flow rates). It is important to note that, in the present study, the bounds \( x_{\text{min}} \) and \( x_{\text{max}} \) vary across the different experimental investigations as they depend on $P_2(t)$ and $P_1(t)$.

\subsubsection{Training of neural network}

\noindent In the previous section, we discussed PINN formulation for the present ESP problem and the scaling of the input and output. The scaling helps in converging the network faster. The next step is to obtain the optimal parameters of the network and the unknown parameters. This is done using a gradient-based optimization strategy that minimizes the loss function discussed earlier. As discussed in the previous section, the trainable parameters ($\bm{\theta}$, $\bm{\Lambda}$, $\bm{\lambda}_d$, $\bm{\lambda}_r$ and $\bm{\lambda}_{ic}$) are optimized using the Adam optimizer. The automatic differentiation for the derivatives of the outputs with respect to input is evaluated using the Python library JAX. At the same time, the Adam optimization is carried out using the Optax library. As discussed earlier, we have considered self-adaptive weight \cite{mcclenny2022selfadaptive}. Thus, we considered different optimizers (three) for the neural network parameters (weights and biases), unknown parameters, and self-adaptive weights. For each numerical example, we discuss the number of training epochs and learning rate scheduler in detail in \cref{apdx:trn-cfg}. Furthermore, the unknown parameters transformations will be discussed in \cref{sec:results_pinn}. We have used \num{100} collocation points for all cases examined.

The PINN's hyperparameters were manually fine-tuned to improve model performance using empirical observations and domain-specific knowledge. Furthermore, a hyperparameter optimization study was conducted using the Tree-structured Parzen Estimator (TPE) method, as described by \citet{bergstra2011algorithms}, with the Optuna Python library \cite{optuna_2019}. However, this manuscript mainly focuses on the manual tuning results, which presented better performance. The hyperparameter optimization study details are provided in \cref{apdx:ho-results} for reference.

\FloatBarrier

\subsection{Data generation}\label{subsec:data-generation}

\noindent We evaluate the performance and applicability of the present PINN method using both experimental and simulated data. The simulated data were obtained from the system described in \cref{subsec:esp-model}. This approach provides a synthetic dataset for assessing the performance and reliability of PINNs in solving inverse problems, circumventing the challenges of uncertainty, and modeling approximations when considering experimental data. This section details the steps involved in generating the simulated data.

In this study, we employed the Tsitouras 5/4 Runge-Kutta method to solve the ESP system set of ODEs presented in \cref{subsec:esp-model}. We set both relative and absolute tolerances to \num{1e-8}. The solver operates with a variable time step, which was subsequently resampled to a fixed time step of \(\Delta t = \qty{0.0001}{\second}\). We selected this value to adequately capture the system's dynamics while ensuring compatibility with the sampling rate utilized in experimental tests. While for training the PINN and used as measurements in the PF, we considered a fixed time step of \(\Delta t = \qty{0.5}{\second}\).

The model input is the torque ($\gamma(t)$), which we directly measured during each experimental investigation. After post-processing the torque signal, as described in \cref{subsec:daq}, we observed that the torque signal still contained relatively high-frequency components, which would be unrealistic as an input for the ESP system. Therefore, we applied a Butterworth low-pass filter with a cut-off frequency of \qty{2}{\hertz} to the torque signal to obtain a more representative input signal.

For fluid viscosity, we assumed a single value for the entire system. We calculated the average water fraction and temperature observed during the corresponding experimental investigation and estimated the emulsion viscosity using the \citet{brinkman1952} model. As for fluid density, we directly measured it using a Coriolis meter. Then, we calculated the average density obtained from the measurements conducted during each experimental investigation.

Furthermore, to evaluate the capability of PINNs in handling measurement uncertainties while excluding potential influences of missing physics or idealizations of the ESP system model, we incorporated Gaussian noise into the simulated signals $P_{1}$ and $P_{2}$. The noise values used in these signals were determined based on the manufacturer's specifications for the pressure transmitters' uncertainties, as detailed in \cref{tab:sensor_spec}.

\subsection{Structural identifiability analysis}\label{subsec:structiden}

In the \cref{subsec:esp-model}, we introduced a set of ODE that describes the ESP system dynamics that has \num{\numberofps} parameters and \num{\numberofstates} states variables. Before trying to estimate the unknown parameters of the system using the PINN, it is important to assess whether the parameters can be uniquely determined from a given set of data.

Consider a dynamical system in the following format
\begin{align}
  & \dot{\bm{x}}(t) = \bm{f}(\bm{x}(t), \bm{u}(t), \bm{\Theta}), \\
  & \bm{y}(t) = \bm{h}(\bm{x}(t), \bm{u}(t), \bm{\Theta}),
\end{align}
where $\bm{x}(t)$ is an $m$-dimensional state vector, $\bm{u}(t)$ is an $n$-dimensional input signal, $\bm{y}(t)$ is an $r$-dimensional output signal or the measurable output, and $\bm{\Theta}$ is $k$-dimensional vector of parameters. A parameter set $\bm{\Theta}$ is said to be structurally globally identifiable if the following property holds:
\begin{equation}\label{eq:identifiable-ps}
  \bm{h}(\bm{x}(t), \bm{u}(t), \bm{\Theta}) = \bm{h}(\bm{x}(t), \bm{u}(t), \bm{\beta}) \implies \bm{\Theta} = \bm{\beta},
\end{equation}
where $\bm{\beta}$ is a $k$-dimensional vector of parameters. Furthermore, if the property in \cref{eq:identifiable-ps} holds within a neighborhood of $\bm{\Theta}$, it is referred to as structurally locally identifiable. 

Thus, the identifiability property serves as a prerequisite for practical parameter estimation. Additionally, as stated by \citet{daneker2022}, if a parameter is locally identifiable, it implies that the search range for that parameter should be limited before attempting its estimation. On the other hand, for globally identifiable parameters, there is no need to define a search range.

\subsection{Practical identifiability analysis}\label{subsec:practiden}

\noindent In \cref{subsec:structiden}, it is discussed that the circumstances under a system are structurally locally identifiable. However, it is worth noting that this analysis is conducted assuming that the measured variables have no noise and that the model is error-free. Hence, we must also assess if we can estimate parameters accurately based on the quantity and quality of the available data. This analysis is known as practical or posterior identifiability analysis, and it is possible that a structurally identifiable system may not be practically identifiable.

The practical identifiability can be assessed using either Monte Carlo simulations or sensitivity analysis. The Monte Carlo approach rigorously determines the practical identifiability of the model but comes with a high computational cost due to the requirement of multiple model fits. On the other hand, sensitivity analysis offers a faster computation method and provides information about the correlation structure among the parameters. This correlation structure can guide the fixing of parameters when practical identifiability is not achieved \cite{tuncer2018,miao2011}.

In this study, we considered only the sensitivity approach to practical identifiability. Then, we used the Fisher information matrix (FIM) to compute the correlation matrix of all parameters to determine their practical identifiability. We estimate the sensitivities of the ESP system model with respect to the parameters with the Julia package DiffEqSensitivity.jl. For the sensitivity analysis, we considered the simulated case, and similarly to \citet{daneker2022}, we considered a noise level of the measurements of \qty{1}{\percent}.

Then, to obtain the correlation matrix, we first estimate the covariance matrix ($\bm{C}$), which can be approximately obtained from the FIM by
\begin{equation}
  \bm{C} = FIM^{-1}.
\end{equation}
We then estimate the correlation matrix from ($\bm{C}$) with
\begin{equation}
  \begin{cases}
    r_{ij} = \frac{\bm{C}_{ij}}{\sqrt{\bm{C}_{ii}\,\bm{C}_{jj}}}, & \text{if }  i \neq j, \\
    r_{ij} = 1,                                    & \text{if }  i = j.\\
  \end{cases}
\end{equation}
When $|r_{ij}|$ is close to \num{1}, the parameters $i$ and $j$ are strongly correlated and cannot be individually estimated. Therefore, the parameters are practically unidentifiable \cite{tuncer2018,miao2011}. 

\section{Results and discussions for identifiability analysis}\label{sec:results_iden}

\subsection{Local structural identifiability}\label{subsec:locstruct}

\noindent We utilized the method of differential elimination for dynamical models via projections proposed by \citet{dong2022} to evaluate the structural identifiability of the ESP system model. It is important to note that our analysis focused specifically on assessing local identifiability due to out-of-memory issues when attempting to test for structural global identifiability. Additionally, our investigation considered the suction and discharge pressure ($P_1$ and $P_2$) as available measurements, as they are typically measured in ESP deployments within oil fields.

From \cref{tab:model-ps,tab:model-ps2}, we observe that the ESP system model initially contains a total of \num{\numberofps} parameters. However, it is worth noting that \num{8} of these parameters are geometrical parameters, such as pipe diameter, cross-sectional area, and pipeline length. They can be readily obtained, and they were fixed for both the upstream and downstream pipelines. Furthermore, \num{3} parameters refer to the twin-screw pump, which serves as a pressure booster for the flow line. In actual oil field extraction, the pressure is a characteristic of the well. Thus, we reduce the number of unknown parameters to \num{15}. Subsequently, we assessed the local identifiability of the ESP system model based on these \num{15} parameters. The corresponding results are displayed in the first row of \cref{tab:struc}.

\newcommand{\avl}{\checkmark}
\newcommand{\un}{$\times$}
\newcommand{\slf}{$-$}
\begin{table}[!htb]
  \centering
  \caption{\textbf{Local structural identifiability results of the ESP system model, considering that the intake pressure ($\bm{P_1}$) and discharge pressure ($\bm{P_2}$) are known.} The symbols \un indicate locally unidentifiable parameters, \avl indicate locally identifiable parameters, and \slf $\,$ represent fixed parameters.}
  \label{tab:struc}
  \begin{tabular}{llllllllllllllll}
  \toprule
   $k_{1p}$ & $k_{2p}$ & $k_{3p}$ & $k_{4p}$ & $k_{1s}$ & $k_{2s}$ & $k_{3s}$ & $k_{4s}$ & $k_{5s}$ & $I_{s}$ & $\rho$  & $\mu$ & $B$ & $k_{u}$ & $k_{d}$ \\ \midrule
       \un  &     \un  &     \avl &     \avl &     \un  &     \un  &     \un  &     \un  &     \un  &     \un &    \avl &  \avl & \avl&    \avl &   \avl\\
       \avl &     \avl &     \avl &     \avl &     \avl &     \avl &     \un  &     \un  &     \avl &     \slf&    \avl &  \avl & \avl&    \avl &   \avl\\
       \avl &     \avl &     \avl &     \avl &     \avl &     \avl &    \slf  &     \slf&     \avl &     \slf&    \avl &  \avl & \avl&    \avl &   \avl \\ \bottomrule
  \end{tabular}
\end{table}

The ESP system model with \num{15} unknown parameters is found to be structurally locally unidentifiable, and it is impossible to estimate all the parameters simultaneously. As discussed by \citet{daneker2022}, there are several approaches to address the identifiability issue, such as fixing specific parameters or introducing additional measured variables. For simplicity, when we refer to fixing a parameter, we consider that the parameter is known. In this study, we opted to work with the available measurements from the actual field and consider only the suction and discharge pressures ($P_1$ and $P_2$). Therefore, we decided to fix the shaft inertia ($I_s$). Despite this adjustment, the model remains structurally locally unidentifiable, with only $k_{3s}$ and $k_{4s}$ being locally unidentifiable. The corresponding results are shown in the second row of \cref{tab:struc}.

Finally, we solved the issue of the model being locally unidentifiable by fixing both $k_{3s}$ and $k_{4s}$. The results are shown in the third row of \cref{tab:struc}. We found that fixing only one of them did not solve the problem. In summary, the ESP system model can only be structurally locally identifiable with the suction and discharge pressure measurements if the shaft inertia ($I_s$), viscous damping coefficient ($k_{4s}$), and disk friction coefficient ($k_{3s}$) are known.

\subsection{Practical identifiability}\label{subsec:practstruct}

\begin{figure}[H]
  \centering
  \begin{subfigure}[b]{0.475\textwidth}
    \centering
    \includegraphics[trim={1cm 0cm 1cm 0cm}, scale=1.35]{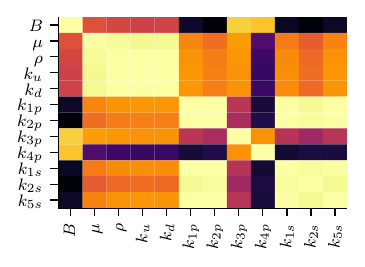}
    \caption{Twelve unknown parameters.} 
    \label{fig:practical_corr_12}
  \end{subfigure}
  \hfil
  \begin{subfigure}[b]{0.475\textwidth}
    \centering
    \includegraphics[trim={1cm 0cm 1cm 0cm}, scale=1.35]{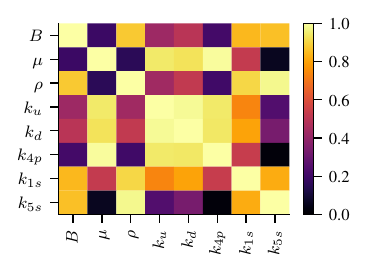}
    \caption{Eight unknown parameters.}
    \label{fig:practical_corr_8}
  \end{subfigure}
  \hfil
  \caption{\textbf{Absolute values of the estimated correlation matrices for ESP system model parameters.} The matrices are derived based on the FIM. The matrix indicates whether parameters are identifiable or not. Parameters with absolute correlation values near 1 are practically unidentifiable. The \cref{fig:practical_corr_12} represents twelve structurally locally identifiable parameters, while the \cref{fig:practical_corr_8} focuses on eight parameters, particularly keeping fluid properties and pipeline resistances unknown.}
\end{figure}

\noindent In \cref{subsec:locstruct}, we discussed the circumstances under which the ESP system is structurally locally identifiable. In this section, we further assess the ESP system's identifiability by considering the practical identifiability analysis discussed in \cref{subsec:practiden}. The absolute of correlation matrix of the parameters structurally locally identifiable of the ESP system is shown in \cref{fig:practical_corr_12}. For the practical identifiability analysis, we considered the simulated case for the first investigation of the test matrix, \cref{tab:test-matrix}.

The correlation analysis shown in \cref{fig:practical_corr_12} demonstrates a significant correlation among multiple parameters. Notably, the fluid density ($\rho$) and the pipeline resistances ($k_u$ and $k_d$) exhibit a strong correlation. Consequently, although the initial analysis indicated structural local identifiability of these parameters, they are not practically identifiable within the context of the conducted investigation.

As many parameters are strongly correlated, we need to fix some parameters to make the system practically identifiable. We set that the main variables of interest are the fluid parameters and the pipeline resistances since the fluid parameters ($B$, $\mu$, and $\rho$) change within the water fraction, which changes with the well's life, and the pipeline resistance ($k_u$ and $k_d$) changes with the actuation of valves and wax deposition for instance. Thus, we preferably fixed the pump and shaft parameters. We started by fixing the parameter $k_{1p}$, then the $k_{3s}$, $k_{2p}$ and $k_{3p}$ successively. The resultant parameter correlation matrix is presented in \cref{fig:practical_corr_8}.
It is noticeable from \cref{fig:practical_corr_8} that after fixing some parameters, they are less correlated. However, some parameters are still correlated, such as the viscosity ($\mu$) and the pump equivalent resistance ($k_{4p}$). However, this correlation is not as strong as the correlation between density ($\rho$) and pipeline resistance ($k_{u}$) observed in \cref{fig:practical_corr_12}. In \cref{fig:practical_corr_8}, the correlation coefficient between $\mu$ and $k_{4p}$ is \num[round-mode=figures, round-precision=4]{0.988021}, while in \cref{fig:practical_corr_12}, the correlation coefficient between $\rho$ and $k_{u}$ is \num[round-mode=figures, round-precision=4]{0.999359}.

\section{Results and discussions for inverse problem using PINN}\label{sec:results_pinn}

\begin{figure}[H]
    \centering
    \includegraphics[width=0.85\textwidth]{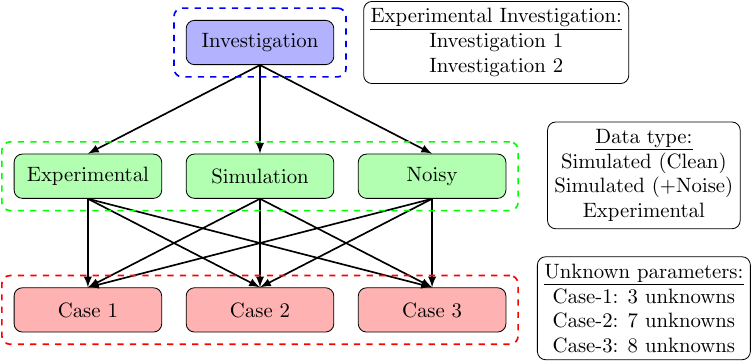}
      \caption{\textbf{Cases and scenarios evaluated.} As presented in \cref{subsec:ps-var}, two experimental investigations (blue) are performed, considering different water fractions and initial angular velocities. Each experimental investigation is analyzed using three different data sources (scenarios): simulation data, simulation data with added Gaussian noise, and experimental data collected from the instruments (green). For each scenario, three sets of unknown parameters, denoted as cases, are evaluated  (red). Thus, we have \num{9} different results for each experimental investigation.}
    \label{fig:study}
\end{figure}

\noindent From the structural identifiability analysis presented in \cref{sec:results_iden}, we have defined three cases to evaluate the PINN for parameter and state estimation in the ESP system. Consistent with \cref{sec:results_iden}, we considered only the suction and discharge pressures as the measured variables. The three cases are as follows:
\begin{enumerate}[label=Case \arabic*]
  \item Flow parameters: In this case, we assumed as known all parameters of the ESP system model except for the flow properties, namely, density ($\rho$), viscosity ($\mu$), and the bulk modulus ($B$).
  \item Flow, pipeline, and impeller parameters: Based on the local identifiability analysis results presented in the first row of \cref{tab:struc}, we assumed as known the parameters marked as locally unidentifiable and estimate the remaining identifiable ones. The unknown parameters considered are the flow properties ($B$, $\mu$, $\rho$), the upstream pipeline equivalent resistances ($k_u$ and  $k_d$) and the pump parameters $k_{3p}$ and $k_{4p}$.
  \item Flow, pipeline, impeller, and shaft parameters: Considering the findings from the practical identifiability analysis outlined in \cref{subsec:practstruct}, we estimate the parameters depicted in \cref{fig:practical_corr_8} while assuming as known the remaining parameters. The unknown parameters are fluid properties ($B$, $\mu$, $\rho$), the upstream pipeline equivalent resistances ($k_u$ and $k_d$), the pump parameter $k_{4p}$, and the shaft parameters $k_{1s}$ and $k_{5s}$.
\end{enumerate}
For each case, we will evaluate the performance and effectiveness of PINN in estimating the unknown parameters and non-measured system states in three distinct data scenarios:
\begin{enumerate}
 \item Simulated data: This scenario represents an ideal condition without any disturbances or modeling errors, allowing us to assess the baseline performance of the proposed method.
 \item Simulated data with Gaussian noise: In this scenario, we introduce Gaussian noise to the simulated data to evaluate the sensitivity of the proposed method to noisy data.
 \item Experimental data: This scenario is closer to actual oil field conditions, incorporating noisy measurements, modeling errors, and missing information about certain parameters.
\end{enumerate}
Furthermore, to evaluate the PINN performance, we have implemented the particle filter (PF) method for each scenario and case, running it \num{30} times with varying random seeds. Similarly, the PINN was trained \num{30} times for each study to assess the impact of neural network weights initialization. The results are presented and discussed in each case for the two experimental conditions defined in \cref{tab:test-matrix}. \cref{fig:study} presents a schematic representation of the cases and scenarios evaluated.

During the training of the PINN, we observed that the significant difference in magnitude between the states, with pressure in the order of $10^5$ and flow rates in the order of $10^{-2}$, affected the training and tuning process despite the scaling in the physics loss and the usage of self-adaptive weights on the loss function. Therefore, we converted the units of the states from SI units to \qty{}{\cubic\meter\per\hour} for the volumetric flow rates and to \qty{}{MWC} (meters of water column) for the pressure. After obtaining the results, we converted them back into SI units.

Furthermore, the transformation of the unknown parameters contributes to the convergence and accuracy of the PINN by restricting the search space to the local neighborhood of the parameter. We first tried to keep the linear scaling for all parameters and cases during the development. However, in Case 3, we could not achieve satisfactory results without restricting the parameter search space using bounded transformations such as $\tanh$. We also took advantage of that in our experiments. It is not possible to have a density higher than the water. Then, we set an upper bound for it. It helped to improve the results in all cases. The transformation scheme for each parameter is shown in \cref{tab:transformps}, and the scaling values are presented in \cref{tab:scalingval}.
\begin{table}[!htb]
  \centering
  \caption{\textbf{Transformations of the unknown parameters for each case evaluated.} The transformations applied to the unknown parameters are detailed. The softplus function $S_p(\cdot)$ and the $S_m(\cdot)$, defined as $S_m(x) = x - \text{softplus}(x) + 1$, are utilized along with a scaling constant $\Lambda$. The $L(\cdot)$ is a linear transformation given by $L(x) = x \Lambda$, which considers only a scaling constant $\Lambda$. For Case 3, the transformation for an unknown parameter is expressed as $\Lambda_{\text{sc}}(x) = \Lambda_{\text{true}}\left(\tanh(x)\alpha + 1\right)$, where $x$ represents the PINN estimated value of the unknown parameter, $\Lambda_{\text{true}}$ is the parameter true value, and $\alpha$ denotes the span percentage around the true value. For most parameters, we set the span to $\pm 50\%$ ($\alpha = 0.5$), except for the $B$ parameter, which spans $\pm 15\%$ ($\alpha = 0.15$).}
  \label{tab:transformps}
  \setlength{\tabcolsep}{5.5pt}
  \begin{tabular}{lcccccccccc}
    \toprule
    \textbf{Case} &   $B$                     & $\mu$  & $\rho$       & $k_u$  & $k_d$  & $k_{3p}$ & $k_{4p}$ & $k_{1s}$ & $k_{5s}$ \\
    \midrule
    \textbf{Case 1} & $(S_p(x) + 0.9)\Lambda$ & $L(x)$ & $S_m(x)\Lambda$ &     &     &       &       &       &  \\
    \textbf{Case 2} & $(S_p(x) + 0.9)\Lambda$ & $L(x)$ & $S_m(x)\Lambda$ & $L(x)$ & $L(x)$ & $L(x)$   & $L(x)$   &       &  \\
    \textbf{Case 3} & $\Lambda_{\text{sc}}(x)$ & $\Lambda_{\text{sc}}(x)$ & $S_m(x)\Lambda$ & $\Lambda_{\text{sc}}(x)$& $\Lambda_{\text{sc}}(x)$ & & $\Lambda_{\text{sc}}(x)$ & $\Lambda_{\text{sc}}(x)$ & $\Lambda_{\text{sc}}(x)$\\
    \bottomrule
  \end{tabular}
\end{table}
\begin{table}[!htb]
  \centering
  \caption{\textbf{Scaling constants for the unknown parameters.} The table presents the scaling constants used in linear parameter transformations and the scaling parameter for the liquid bulk modulus ($B$) and density ($\rho$) transformations.}
  \label{tab:scalingval}
  \begin{tabular}{lcccccccc}
    \toprule
    \textbf{Case} &   $B$                     & $\mu$  & $\rho$       & $k_u$  & $k_d$  & $k_{3p}$ & $k_{4p}$\\
    \midrule
    \textbf{Case 1} & \num{1e9} & \num{1} & \num{1000} &     &     &       & \\
    \textbf{Case 2} & \num{1e9} & \num{1} & \num{1000} & \num{3.e+2}  & \num{1.e+1}    & \num{-1.0e+7} & \num{-1.0e+2}\\
    \textbf{Case 3} &  &  & \num{1000} &  &  &  &  \\
    \bottomrule
  \end{tabular}
\end{table}

\FloatBarrier
\subsection{State estimation results}\label{subsec:states}

\noindent We begin by assessing the state estimation performance of the proposed PINN in comparison to the Particle Filter (PF) when applied to the ESP system model. The proposed PINN and PF are utilized to estimate the unknown system states using the known suction and discharge pressure signals ($P_1$ and $P_2$). In the simulated cases, the unknown states include the pipeline and ESP volumetric flow rates ($Q_1$, $Q_2$, $Q_p$) and the ESP angular velocity ($\omega$). However, in the experimental case, measurements of the downstream pipeline and ESP volumetric flow rates ($Q_2$ and $Q_p$) are unavailable, and therefore, they were excluded from the accuracy analysis. We would like to mention that these results are from inverse problems with unknown parameters discussed above, not forward problems. First, we present the state estimation results for all the cases considered, as shown in \cref{fig:study}. Then, we discuss the prediction of the unknown parameters for all the cases in \cref{subsec:inv}.

It is worth mentioning that in PINN, it is common practice to use second-order optimization algorithms, such as L-BFGS, after training with the first-order algorithm. This additional step usually aims to fine-tune the PINN's performance. However, we observed that using L-BFGS after Adam adversely affected the parameter estimation in the noisy and experimental case. Therefore, we decided to rely solely on the first-order optimization algorithm, as our focus is on unknown parameter estimation in more realistic scenarios. It is important to note that the simulated cases serve as a baseline performance for assessing the PINN's performance, and we expect that performance will be comparatively reduced under the more challenging conditions posed by experimental data and noise.

\subsubsection{State estimation results for simulated data}\label{subsec:fw:simulated}

\begin{figure}[!htb]
  \centering
  \includegraphics{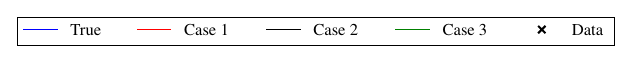}\\
  \begin{subfigure}[b]{0.47\textwidth}
    \centering
    \includegraphics{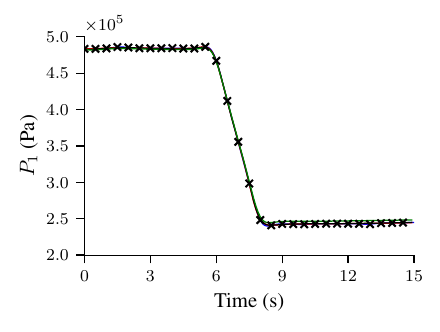}
    \caption{Intake pressure.}
    \label{fig:sim_time:p1}
  \end{subfigure}
  \begin{subfigure}[b]{0.47\textwidth}
    \centering
    \includegraphics{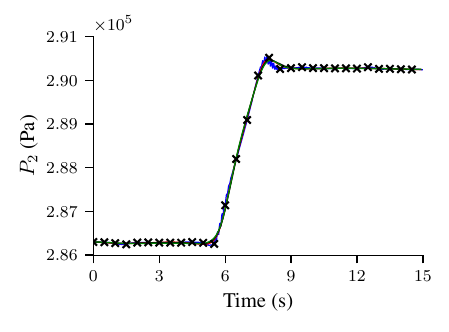}
    \caption{Discharge pressure.}
    \label{fig:sim_time:p2}
  \end{subfigure}\\
  \begin{subfigure}[b]{0.47\textwidth}
    \centering
    \includegraphics{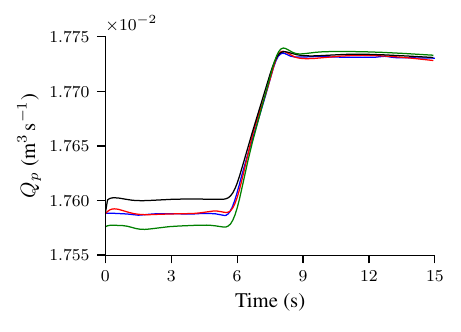}
    \caption{Impeller flow rate.}
    \label{fig:sim_time:qi}
  \end{subfigure}
  \begin{subfigure}[b]{0.47\textwidth}
    \centering
    \includegraphics{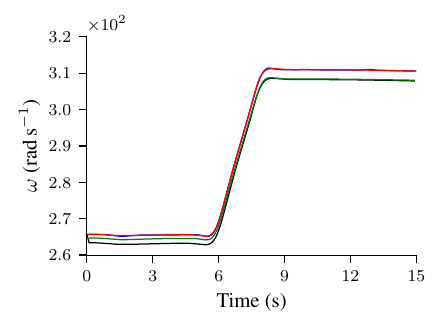}
    \caption{ESP angular velocity.}
    \label{fig:sim_time:omega}
  \end{subfigure}
  \caption{\textbf{Comparison of predicted and true states for the simulated scenario of the first experimental investigation with unknown parameters.} The blue line indicates the true values, while the $\times$ markers denote the data points used for training the PINN. The red, black, and green lines represent the mean values of the predicted states calculated at each time instant across the \num{30} trained PINNs for Cases 1 to 3, as defined earlier in this section (\cref{sec:results_pinn}). Across all cases, the training dataset consists of \num{30} data points for the data loss in $P_1$ and $P_2$ and \num{100} collocation points for the physics loss. For brevity, the results for $Q_1$ and $Q_2$ are presented in \cref{apdx:pinn_results}, as they are similar to $Q_p$.}
  \label{fig:sim_time}
\end{figure}

\noindent As a baseline performance assessment, we begin with the simulated scenario. The predicted states from the PINN are shown in \cref{fig:sim_time}, and in \cref{tab:mape_sim}, we have presented the Mean Absolute Percentage Error (MAPE) for estimated states for both methods, i.e. PINN and PF. The MAPE is defined as
\begin{align}\label{eq:state:mape}
  \text{MAPE} = \frac{1}{N^{mape}} \sum_{i=1}^{N^{mape}} \left| \frac{y_s(t_i) - \bar{y}_s(t_i; \bm{\theta})}{y_s(t_i)} \right|,
\end{align}
where $N^{mape}$ is the total number of temporal data points, the data samples used for calculating the MAPE are the same ones employed for training the PINN and as a measurement in the PF. They have a sampling interval of \(\Delta t = \qty{0.5}{\second}\), as previously discussed in \cref{subsec:data-generation}. The variable $y_s(t_i)$ represents the true value of the state $s$ at time $t_i$, and $\bar{y}_s(t_i; \bm{\theta})$ denotes the mean of the predicted value of the state $s$ from the PINN and PF across 30 different realizations of the results. These results are evaluated using different initializations of the parameters of the network and unknown parameters. For brevity, we restrict our discussion to the state prediction results and the MAPE table for the first experimental investigation. Further results from the second investigation are presented in \cref{apdx:pinn_results}.

\begin{table}[H]
  \caption{\textbf{MAPE for the simulated scenario.} The table presents the results for the simulation data considering three different cases of unknown parameters. The MAPE values represent the error of the simulation results for the states: $P_1$, $P_2$, $Q_1$, $Q_2$, $Q_p$, and $\omega$. As the number of unknown parameters increased from Case 1 to Case 3, the MAPE values also increased, indicating a slight performance loss for the PINN and PF. However, the PINN MAPE values remain relatively small, and visual inspection of \cref{fig:sim_time} reveals a reasonable agreement with the actual values. While for PF the MAPE values for all cases and states were higher than the PINN. Nonetheless, in the PF, it is noticeable that PF was unable to estimate accurately the measured states ($P_1$ and $P_2$) for Cases 2 and 3, with MAPE values higher than \qty{42}{\percent}. On the other hand, the MAPE values for the volumetric flow rates ($Q_1$, $Q_2$ and $Q_p$) and ESP rotation ($\omega$) were consistently smaller than the measured states ($P_1$ and $P_2$), although significantly higher than the PINN.}
  \sisetup{round-mode = places, round-precision = 3}
  \centering
  \begin{tabular}{llllllll}
    \toprule
           & Method &                                $P_1$ &                                  $P_2$ &                                 $Q_1$ &                                $Q_2$ &                                  $Q_p$ &                            $\omega$ \\
    \midrule
    \multirow{2}{*}{Case 1} & PINN &  \qty{0.12904223209751683}{\percent} &  \qty{0.0075823928206989405}{\percent} &  \qty{0.003741529995816875}{\percent} &  \qty{0.00594839797634069}{\percent} &  \qty{0.0071214185964283755}{\percent} &   \qty{0.023362301318421}{\percent} \\
           & PF &   \qty{1.6050748787015992}{\percent} &     \qty{1.3615950897650138}{\percent} &   \qty{0.08913079662138146}{\percent} &  \qty{0.08380652572729595}{\percent} &    \qty{0.03970013925967904}{\percent} &  \qty{0.9751976532396905}{\percent} \\
    \cmidrule{1-8}
    \multirow{2}{*}{Case 2} & PINN &   \qty{0.1265006517745895}{\percent} &    \qty{0.00874120848185776}{\percent} &   \qty{0.03818304237084766}{\percent} &  \qty{0.03667918579210787}{\percent} &    \qty{0.03764721063299241}{\percent} &  \qty{0.8488063527330059}{\percent} \\
           & PF &    \qty{48.24956276274835}{\percent} &      \qty{52.77644031500316}{\percent} &    \qty{0.6967805083384502}{\percent} &   \qty{0.6909366040432701}{\percent} &     \qty{0.6517110613429108}{\percent} &  \qty{0.9147812830180911}{\percent} \\
    \cmidrule{1-8}
    \multirow{2}{*}{Case 3} & PINN &   \qty{0.9153789672980568}{\percent} &     \qty{0.0098352464522773}{\percent} &   \qty{0.04183493259777283}{\percent} &  \qty{0.03955150155314627}{\percent} &   \qty{0.041433789970329885}{\percent} &   \qty{0.612252995280974}{\percent} \\
           & PF &   \qty{42.521611940583206}{\percent} &      \qty{51.89648750794882}{\percent} &    \qty{0.7202383213652941}{\percent} &   \qty{0.7145583745362174}{\percent} &     \qty{0.6750001572814034}{\percent} &  \qty{1.5200015526815545}{\percent} \\
    \bottomrule
    \end{tabular}
    \label{tab:mape_sim}
\end{table}
The comparison between the predicted dynamics of the states by the PINN model and the actual values, as shown in \cref{fig:sim_time}, demonstrates a good overall agreement across all states and cases. Notably, a small error is observed for the $P_1$ and $P_2$ states in all cases. However, a slightly larger error is observed for the volumetric flow rate states ($Q_1$, $Q_2$, and $Q_p$) {and the ESP angular velocity state ($\omega$)}. Among the different cases, Case 1 (red line) exhibits the best agreement across the entire simulation range in both investigations. This is also observed in \cref{tab:mape_sim}, where Case 1 presented the lower MAPE in all states except $P_1$.

In the first investigation, the flow rates in Case 2 (black line) and Case 3 (green line) show a reasonable agreement with the simulation approximately from the \num{6}$^{th}$ second. However, a slight offset from the actual value is observed from the beginning to the \num{6}$^{th}$ second. As for the second investigation (\cref{fig:sim3_time}), a slightly higher offset is observed in the flow rate states and in the ESP angular velocity across the entire simulation time for Case 3, while for Case 2, we no longer observe the offset.

In \cref{tab:mape_sim}, the differences in the MAPE values for the PF measured state estimations compared to other estimated states in Cases 2 and 3 are likely attributable to sample impoverishment and degeneracy. After examining the multiple PF runs, we observed that resampling typically began within the first few seconds and persisted until the end of the simulation, suggesting a low diversity of particles for Cases 2 and 3. It indicates a low diversity of the particles along the simulation with a few particles, below \num{50}, with high importance. To improve particle diversity, we attempted to increase the resampling dispersion. However, most of the particles resulted in the solver being interrupted after reaching the maximum number of iterations. Additionally, as Case 1 was not affected and the difference between the other cases is the number of unknown parameters, the increase of unknown parameters negatively affected the method, which might indicate an issue related to the number of particles used. Furthermore, this issue in Cases 2 and 3 occurred in every data scenario, as will be detailed in \cref{subsec:fw:simulatedWnoise,subsec:fw:exp}.

\FloatBarrier
\subsubsection{State estimation results for simulated data with noise}\label{subsec:fw:simulatedWnoise}

\begin{figure}[!htb]
  \centering
  \includegraphics{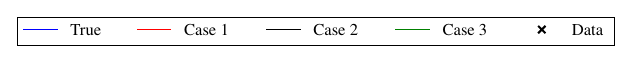}
  \\
  \begin{subfigure}[b]{0.47\textwidth}
    \centering
    \includegraphics{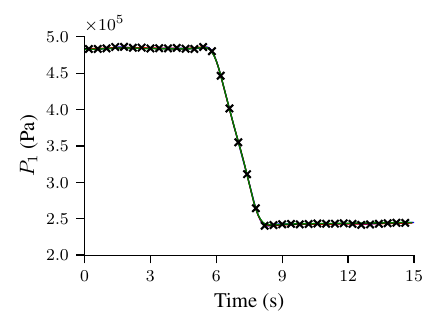}
    \caption{Intake pressure.}
    \label{fig:sim_noi_time:p1}
  \end{subfigure}
  \begin{subfigure}[b]{0.47\textwidth}
    \centering
    \includegraphics{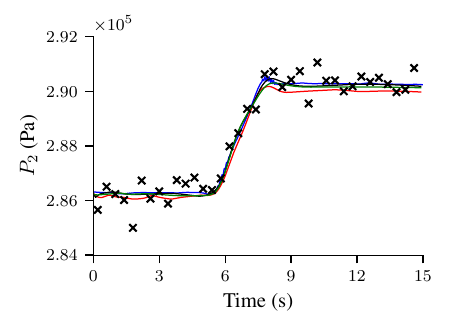}
    \caption{Discharge pressure.}
    \label{fig:sim_noi_time:p2}
  \end{subfigure}\\
  \begin{subfigure}[b]{0.47\textwidth}
    \centering
    \includegraphics{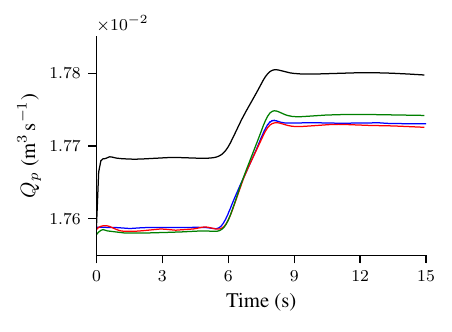}
    \caption{Impeller flow rate.}
    \label{fig:sim_noi_time:qi}
  \end{subfigure}
  \begin{subfigure}[b]{0.47\textwidth}
    \centering
    \includegraphics{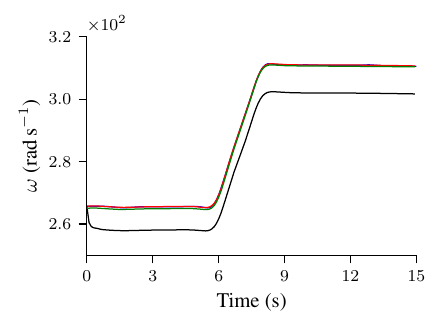}
    \caption{ESP angular velocity.}
    \label{fig:sim_noi_time:omega}
  \end{subfigure}\\
  \caption{\textbf{Comparison of predicted and true states for the simulated case with the noise of the first experimental investigation and unknown parameters.} In \cref{fig:sim_noi_time:p1} and \cref{fig:sim_noi_time:p2}, the blue line indicates the signal with added Gaussian noise. In other figures, the same blue line represents the noise-free simulated state. This differentiation aids in evaluating the PINN's accuracy by comparing its predictions with the actual state values and also presents the signal used for PINN training. Each PINN training used a different random seed. Therefore, the Gaussian noise in the signal varied for every PINN training and case. Specifically, the noisy signals for $P_1$ and $P_2$ in \cref{fig:sim_noi_time:p1} and \cref{fig:sim_noi_time:p2} correspond to the first training for Case 1, and they were used as a reference. The added Gaussian noise considered the uncertainty from the pressure transducers as specified in \cref{tab:sensor_spec}. For all cases, the training dataset consisted of \num{37} data points for data loss in $P_1$ and $P_2$ and \num{100} collocation points for physics loss. The plots show the mean value of results for 30 realizations of PINNs. The noisy data shown in the plots are one of the noisy data shown as a representative sample. For brevity, the results for $Q_1$ and $Q_2$ are presented in \cref{apdx:pinn_results}, as they are similar to $Q_p$.}
  \label{fig:sim_noi_time}
\end{figure}

\noindent To further evaluate the robustness of the PINN model, we extend our analysis to include a simulated scenario with Gaussian noise. This section aims to assess the performance of the PINN model and compare it with PF when subjected to noise alone without considering missing physics or errors in the model parameter estimation. For each PINN realization, a different noise realization is also considered. Similarly to the previous section, we will focus on presenting the result corresponding to the first experimental investigation, while the result for the second experimental investigation can be found in \cref{apdx:pinn_results} for reference. The MAPE for the different predicted states is shown in \cref{tab:mape_sim_noi}, and the output of the predicted states from the neural network, along with the addition of noise on the pressure signals, are shown in \cref{fig:sim_noi_time}.
\begin{table}[H]
  \sisetup{round-mode = places, round-precision = 3}
  \caption{\textbf{MAPE for the simulated scenario with added noise.} The table presents the MAPE for state prediction in the simulated case with added noise data, considering three cases of unknown parameters. Similarly to \cref{tab:mape_sim}, increasing the number of unknown parameters (Case 1 to Case 3) leads to higher MAPE values for all states, indicating a performance loss. Nevertheless, the MAPE values remain relatively small, and visual inspection of \cref{fig:sim_noi_time} shows a reasonable agreement with the actual values, except for Case 2. Similar to findings from the simulated data scenario (\cref{subsec:fw:simulated}), the MAPE values for the measured states ($P_1$ and $P_2$) in Cases 2 and 3 are notably higher, despite the low MAPE values for the other states. Notably, in Case 2, the MAPE values for volumetric flow rates ($Q_1$, $Q_2$, and $Q_p$) and ESP rotation ($\omega$) were comparable between the PINN and PF. Conversely, in the other cases, the PINN consistently demonstrated lower MAPE values compared to those of the PF.}
  \centering
  \begin{tabular}{llllllll}
    \toprule
            & Method &                                $P_1$ &                                $P_2$ &                                 $Q_1$ &                                 $Q_2$ &                                 $Q_p$ &                             $\omega$ \\
    \midrule
    \multirow{2}{*}{Case 1} & PINN &  \qty{0.18250453710171932}{\percent} &   \qty{0.1271231144036381}{\percent} &  \qty{0.017266866293912577}{\percent} &  \qty{0.016902615161262926}{\percent} &  \qty{0.017701240452358963}{\percent} &  \qty{0.02598153988909484}{\percent} \\
            & PF &   \qty{1.6306309924676194}{\percent} &   \qty{0.7180009306440753}{\percent} &    \qty{0.1328022964984469}{\percent} &   \qty{0.12238666175067323}{\percent} &   \qty{0.08436631908901708}{\percent} &   \qty{0.6015769661632667}{\percent} \\
    \cmidrule{1-8}
    \multirow{2}{*}{Case 2} & PINN &  \qty{0.20741437048547043}{\percent} &  \qty{0.10535721207434572}{\percent} &   \qty{0.44341081635479024}{\percent} &    \qty{0.4439459082274542}{\percent} &   \qty{0.44338042125176896}{\percent} &    \qty{2.747291618758699}{\percent} \\
            & PF &     \qty{48.5003496228037}{\percent} &    \qty{52.57294621343982}{\percent} &    \qty{0.6543616444784626}{\percent} &    \qty{0.6462940846179008}{\percent} &    \qty{0.6080869162244529}{\percent} &   \qty{1.0458916144117323}{\percent} \\
    \cmidrule{1-8}
    \multirow{2}{*}{Case 3} & PINN &   \qty{0.2256368540598666}{\percent} &  \qty{0.10024333684370594}{\percent} &   \qty{0.04770056999815104}{\percent} &   \qty{0.04440522384225304}{\percent} &   \qty{0.04833744279312106}{\percent} &  \qty{0.16233675278350793}{\percent} \\
            & PF &    \qty{42.95874754402584}{\percent} &      \qty{52.072619467269}{\percent} &    \qty{0.6735315459849626}{\percent} &    \qty{0.6682675496181358}{\percent} &    \qty{0.6284395834883577}{\percent} &    \qty{1.660004150512305}{\percent} \\
    \bottomrule
    \end{tabular}
  \label{tab:mape_sim_noi}
\end{table}
The comparison between the predicted dynamics of the states by the PINN and the actual values, as depicted in \cref{fig:sim_noi_time}, demonstrates a good overall agreement in all states and investigations for almost all cases. However, in Case 2 (black line), the PINN was unable to accurately estimate the flow rate states ($Q_1$, $Q_2$, and $Q_p$,). They exhibit relatively higher errors. However, the model is able to predict the general shape of the time series. Remarkably, these higher errors were observed exclusively in the first experimental investigation (\cref{fig:exp_time}) (higher water fraction), whereas in the second investigation (\cref{fig:exp3_time}) (with low water fraction), Case 2 demonstrated performance similar to that of Case 1 (red line). 

Notably, minor errors are observed in the $P_1$ and $P_2$ states for all cases and investigations. Specifically, in the discharge pressure state (\cref{fig:sim_noi_time:p2}), we can observe that the neural network did not overfit and could capture the signal trend and magnitude. Similarly to the simulated data without noise (\cref{subsec:fw:simulated}), Case 1 (red line) exhibits the best agreement across the entire simulation range in both investigations. Additionally, as shown on \cref{tab:mape_sim_noi}, Case 1 had a lower MAPE than the other cases for the unknown states and a higher MAPE for the known state $P_2$.

Despite the addition of noise, the state estimation for Cases 1 and 3 provides good accuracy without significant errors. These results were similar to those obtained in noise-free conditions, with a slight increase in MAPE values. In Case 3, there is a noticeable offset in the volumetric flow rates ($Q_1$, $Q_2$ and $Q_p$) approximately after the $\num{8}^{\text{th}}$ second, which is different from the observations made in \cref{subsec:fw:simulated}, where the offset occurred only in the beginning. Additionally, the angular velocity, in this case, remains closely aligned with the actual values throughout the simulation, exhibiting no offset. It is observed that results for Case 2 deviate from the true values, particularly for flow rates ($Q_1$, $Q_2$, $Q_p$). It may be possible that, though Case 2 is structurally identifiable, it is not practically identifiable.

\FloatBarrier
\subsubsection{State estimation results for experimental data}\label{subsec:fw:exp}

\begin{figure}[!htb]
  \centering
  \includegraphics{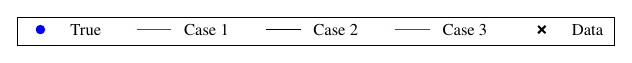}\\
  \begin{subfigure}[b]{0.47\textwidth}
    \centering
    \includegraphics[width=\textwidth, keepaspectratio]{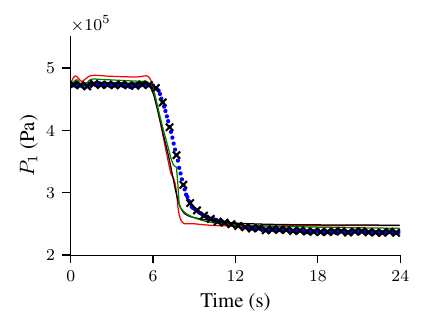}
    \caption{Intake pressure.}
    \label{fig:exp_time:p1}
  \end{subfigure}
  \begin{subfigure}[b]{0.47\textwidth}
    \centering
    \includegraphics[width=\textwidth, keepaspectratio]{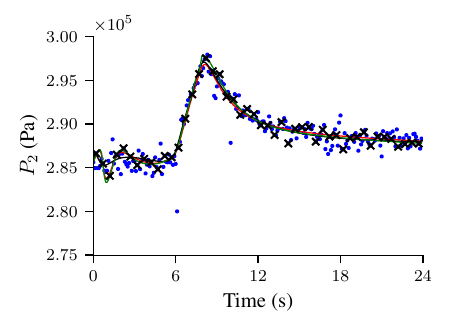}
    \caption{Discharge pressure.}
    \label{fig:exp_time:p2}
  \end{subfigure}\\
  \begin{subfigure}[b]{0.47\textwidth}
    \centering
    \includegraphics[width=\textwidth, keepaspectratio]{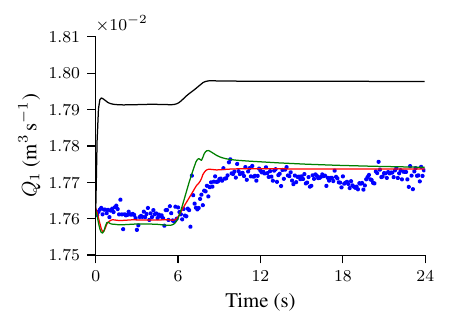}
    \caption{Upstream flow rate.}
    \label{fig:exp_time:q1}
  \end{subfigure}
  \begin{subfigure}[b]{0.47\textwidth}
    \centering
    \includegraphics[width=\textwidth, keepaspectratio]{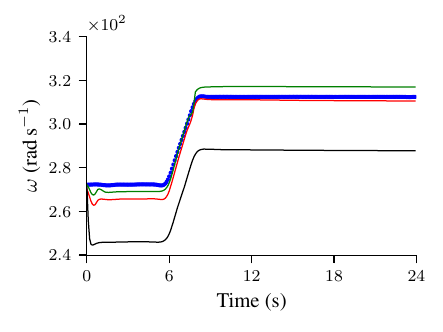}
    \caption{ESP angular velocity.}
    \label{fig:exp_time:omega}
  \end{subfigure}\\
  \caption{\textbf{Comparison of predicted and true states for the experimental case of the first experimental investigation with unknown parameters.} Due to constraints in our experimental setup, we could only directly measure the states $Q_1$, $P_1$, $P_2$, and $\omega$. Therefore, the PINN results analysis will primarily concentrate on these four states, denoted by the graphs' blue lines. Moreover, for all cases, the training dataset consists of \num{48} data points for data loss with \num{100} collocation points for physics loss. The plots show the mean value of results for 30 realizations of PINNs.}
  \label{fig:exp_time}
\end{figure}

\noindent In this section, we assess the performance of the PINN when subjected to experimental data, which is a more representative scenario of real-world oil field conditions. In this case, noise in data and model uncertainties exist, posing a more challenging scenario for the PINN model. Similarly to the previous sections, the results from the first experimental investigation are provided in this section, while results from the second investigation can be found in \cref{apdx:pinn_results}. The  MAPE values for these predictions are presented in \cref{tab:mae_exp}. Meanwhile, the predicted outputs of the neural network are shown in \cref{fig:exp_time}.
\begin{table}[!htb]
  \sisetup{round-mode = places, round-precision = 3}
  \caption{\textbf{MAPE for the experimental data scenario.} The table presents the PINN and PF MAPE values for state predictions using experimental data across the three cases with varying unknown parameters. Unlike the results from the simulated and noise-added scenarios, there was no significant variation among the cases for the PINN. However, while the PF estimated the volumetric flow rate ($Q_1$) with relatively small MAPE values across all cases, it failed to estimate the ESP rotation ($\omega$). Additionally, similar to other data scenarios, the MAPE values for the intake and discharge pressures ($P_1$ and $P_2$) were high in Cases 2 and 3. In contrast, for the PINN, Case 1 showed the best agreement regarding the upstream volumetric flow rate ($Q_1$) and the ESP angular velocity $\omega$. However, similarly to the noisy data scenario, Case 2 presented errors for $\omega$ and $Q_1$ that are considerably higher than the other states, as evident in \cref{fig:exp_time:q1,fig:exp_time:omega}. Nonetheless, the MAPE values remain relatively small for the other cases and states, and visual inspection of \cref{fig:exp3_time} shows a reasonable agreement with the actual values.
  }
  \centering
  \begin{tabular}{llllll}
    \toprule
           & Method &                               $P_1$ &                                $P_2$ &                                $Q_1$ &                            $\omega$ \\
    \midrule
    \multirow{2}{*}{Case 1} & PINN &   \qty{4.489809654082104}{\percent} &  \qty{0.26811930067640444}{\percent} &  \qty{0.14113054866715527}{\percent} &  \qty{1.0637595643280762}{\percent} \\
           & PF &   \qty{5.062628848020504}{\percent} &   \qty{0.7445399264895389}{\percent} &  \qty{0.20299022536416933}{\percent} &   \qty{78.75607706305048}{\percent} \\
    \cmidrule{1-6}
    \multirow{2}{*}{Case 2} & PINN &  \qty{3.8126667843811184}{\percent} &  \qty{0.24555227752123235}{\percent} &    \qty{1.545346534214616}{\percent} &   \qty{8.162325725343337}{\percent} \\
           & PF &   \qty{38.03907014987848}{\percent} &    \qty{38.71307977654926}{\percent} &   \qty{0.5827937003571948}{\percent} &   \qty{78.87141068457973}{\percent} \\
    \cmidrule{1-6}
    \multirow{2}{*}{Case 3} & PINN &  \qty{3.0149922960998126}{\percent} &   \qty{0.2779481717917182}{\percent} &  \qty{0.21167987858096124}{\percent} &   \qty{1.286701131585555}{\percent} \\
           & PF &   \qty{38.27414076570722}{\percent} &   \qty{44.683398323652916}{\percent} &   \qty{0.6142302848090686}{\percent} &   \qty{78.87720534417178}{\percent} \\
    \bottomrule
    \end{tabular}
    \label{tab:mae_exp}
\end{table}

By comparing the PINN predicted dynamics with the actual values (\cref{fig:exp_time}), a good overall agreement is evident across all states and cases, even in the presence of noise and model errors. However, similarly to the noisy data scenario, in Case 2 (black line), the PINN was unable to accurately estimate the flow rate ($Q_1$) and angular velocity ($\omega$) states. While the predicted states follow the signal's shape, they exhibit relatively high errors. Remarkably, these higher errors were observed exclusively in the first experimental investigation (\cref{fig:exp_time}) (high water fraction), whereas in the second investigation (\cref{fig:exp3_time}) (low water fraction), Case 2 demonstrated performance similar to that of Case 3 (green line). These high errors are observed in \cref{tab:mae_exp}, where the Case 2 values for $Q_1$ and $\omega$ had the highest MAPE values. 

For the $P_1$ state (\cref{fig:exp_time:p1}), we observed a minor deviation in the PINN prediction's shape compared to the experimental data. We can observe that the PINN estimation is settling to a new steady-state condition slightly faster than the experimental data. These differences could be attributed to the influence of the fluid's bulk modulus, which affects the system's settling time and, therefore, the shape, as observed by \citet{fctc2024}. The PINN is likely weighting the ESP system model more than the experimental data. It is important to note that the weights are automatically defined by the self-adaptive weights, which could have resulted in these minor discrepancies in the state shape and settling time.

Furthermore, in the parameter estimation section, we will discuss the challenges encountered in estimating the bulk modulus for the experimental data, as it was not experimentally measured. Moreover, in the discharge pressure state (\cref{fig:exp_time:p2}), it is clear that the PINN did not overfit and could capture the signal trend and magnitude, including the signal overshoot between \qtyrange{6}{12}{\second}, which was considerably smaller in the simulations. This discrepancy between the experimental data and simulation is likely due to the term associated with the fluid's bulk modulus ($B$), as discussed by \citet{fctc2024}.

A systematic error is also noticeable at the beginning of the $\omega$ state (\cref{fig:exp_time:omega}) for Cases 1 and 3. Despite these deviations, the predicted $\omega$ signal aligns well with the experimental data regarding shape and magnitude. When comparing the simulated and measured angular velocity ($\omega$), we also observed a systematic error, indicating a difference between the actual and the obtained model parameters. However, despite this difference, the PINN could reasonably estimate the angular velocity state. The observed discrepancies between the experimental data and model predictions could be attributed to multiple factors. These include model uncertainty arising from approximations, simplifications, assumptions in the model, and the influence of noise in the experimental data.

\subsection{Parameter estimation results}\label{subsec:inv}

\noindent The results for parameter estimation in the experimental investigations \num{1} and \num{2} across three cases are presented in the following sections. It should be noted that we did not measure the emulsion bulk modulus during the experiments. As a result, we cannot assess the accuracy of the values determined by the PINN and PF in the experimental data scenario. The viscosity values obtained were compared to the \citet{brinkman1952} model, described in \cref{subsec:esp-model}. For density estimation, we relied on measurements from the Coriolis meter. This section discusses absolute percentage error; additional tables presenting actual values are available in \cref{apdx:pinn_val}.

In this section, the mean of the absolute percentage error (MAPE) for the parameter estimation results is defined as
\begin{align}
  \text{MAPE}_{\Lambda} = \frac{1}{N^{init}} \sum_{i=1}^{N^{init}} \left| \frac{\Lambda - \hat{\Lambda}_i}{\Lambda} \right|,
\end{align}
where $N^{init}$ is the number of PINN and PF random initializations, considered \num{30} in this study, the variable $\Lambda$ represents the true value of the unknown parameter, and $\hat{\Lambda}_i$ is the estimated unknown parameter for the $i$-th PINN initialization. Additionally, to assess the variability in parameter estimation, we calculate the standard deviation (std) of these absolute percentage errors as follows:
\begin{equation}\label{eq:sdmape}
  \sigma_{\Lambda} = \sqrt{\frac{1}{N^{init}} \sum_{i=1}^{N^{init}} \left(\left| \frac{\Lambda - \hat{\Lambda}_i}{\Lambda} \right| - \text{MAPE}_{\Lambda}\right)^2}.
\end{equation}
While we only discuss the MAPE and standard deviation values in this section, we have shown the mean value of the predicted parameters and standard deviation for all the cases in \cref{tab:exp1_meanstd,tab:exp3_meanstd}.

\subsubsection{Case 1: Three unknown parameters $B$, $\mu$  and $\rho$ for different types of data}\label{subsec:case1}

\noindent In this section, we focus on Case 1, where the bulk modulus ($B$), viscosity ($\mu$), and density ($\rho$) were considered unknown, while the remaining parameters were known. The \cref{tab:inv_case1} presents the MAPE along with the corresponding standard deviations (in parentheses) for these parameters across two experimental investigations and all data scenarios.
\begin{table}[!htb]
  \sisetup{round-mode=places, round-precision=2}
  \caption{\textbf{Comparative mean absolute percentage error (with standard deviations) for Case 1 using PINN and PF.}}
  \centering
  \begin{tabular}{llllll}
    \toprule
      &              &    & \multicolumn{3}{c}{MAPE and std. ($\%$)} \\
      \cmidrule{4-6}
      Investigation & Data type & Method &                                        $B$ &                                                   $\mu$ &                                                   $\rho$ \\
    \midrule
    \multirow{6}{*}{1} & \multirow{2}{*}{Simulated} & PINN &   \num{0.3123914334679204} (\num{0.19303666922700458}) &  \num{0.03048262039075415} (\num{0.021972139301937466}) &  \num{0.019583041849378462} (\num{0.012707899791126381}) \\
      &              & PF &     \num{13.862802245167323} (\num{18.99623312745626}) &       \num{2.756598830307547} (\num{4.884721050646626}) &       \num{2.9902668146576534} (\num{5.527789225926375}) \\
    \cmidrule{2-6}
      & \multirow{2}{*}{Noisy} & PINN &   \num{0.5758474778631687} (\num{0.21045811311958046}) &     \num{0.1246478464937965} (\num{0.0753653424894653}) &    \num{0.09468425026741538} (\num{0.05260687862254404}) \\
      &              & PF &    \num{15.439700659466602} (\num{15.647714384456545}) &      \num{1.3313479001304862} (\num{2.679720687720072}) &        \num{1.529109970181961} (\num{3.659999414800815}) \\
    \cmidrule{2-6}
      & \multirow{2}{*}{Experimental} & PINN &    N/A &      \num{1.802230806257376} (\num{0.4008577575960764}) &     \num{0.7958879926832861} (\num{0.20873381758243328}) \\
      &              & PF &    N/A &      \num{9.390523896739618} (\num{12.046903195886808}) &        \num{5.406819651686946} (\num{8.568963336606613}) \\
    \cmidrule{1-6}
    \cmidrule{2-6}
    \multirow{6}{*}{2} & \multirow{2}{*}{Simulated} & PINN &     \num{0.6118124586223436} (\num{1.164159174886659}) &     \num{0.3155249354959103} (\num{1.3287162986502057}) &     \num{0.29123451847482407} (\num{1.2067855711457436}) \\
      &              & PF &     \num{9.815524224067868} (\num{18.550132042901538}) &      \num{1.5154057521001798} (\num{2.416624237026107}) &      \num{1.8716402386799942} (\num{3.8245466663156518}) \\
    \cmidrule{2-6}
      & \multirow{2}{*}{Noisy} & PINN &  \num{0.48019770476369555} (\num{0.20170805277503845}) &        \num{0.2143052567605567} (\num{0.6038020530427}) &     \num{0.21391718172044438} (\num{0.5817864869230065}) \\
      &              & PF &       \num{19.831677795249} (\num{17.700953921482565}) &      \num{1.336989314738137} (\num{3.4697034591386315}) &       \num{1.9545498532974332} (\num{5.741315691898129}) \\
    \cmidrule{2-6}
      & \multirow{2}{*}{Experimental} & PINN &     N/A &     \num{4.302495983974393} (\num{0.26041427984745963}) &     \num{2.3983552112728317} (\num{0.22275076534145866}) \\
      &              & PF &     N/A  &      \num{8.206780791956906} (\num{11.148561762322158}) &       \num{5.673850844447273} (\num{7.4906542927619615}) \\
    \bottomrule
    \end{tabular}
    \label{tab:inv_case1}
\end{table}

As presented in \cref{tab:inv_case1}, the PINN effectively estimated the fluid parameters across all scenarios and investigations, showing its capability to address the inverse problem for the ESP system problem. The PINN obtained relatively low MAPE values under both simulated and noisy cases, with errors remaining below \qty{1}{\percent}. In contrast, for the experimental data, while errors in Investigation \num{1} remained below \qty{2}{\percent}, Investigation \num{2} presented errors below \qty{4.5}{\percent}. 

In contrast, the PF method consistently exhibited higher MAPE values than the PINN, with MAPE values reaching up to \qty{19.83}{\percent} for the bulk modulus ($B$) in the noisy scenario, whereas the PINN maintained MAPE values as low as \qty{0.48}{\percent}. It is worth noting that for the simulated and noisy scenarios, the PINN error stayed below \qty{0.7}{\percent} for all parameters and investigations, while the lowest MAPE value for the PF was \qty{1.33}{\percent}. The experimental data scenario further highlighted PINN's consistently lower MAPE values.

The standard deviations of the absolute percentage error, presented within parentheses in \cref{tab:inv_case1}, indicate consistent results across experimental investigations, data scenarios, and fluid properties for PINN, with most standard deviations below \qty{1}{\percent}. However, Investigation \num{2} under simulated data scenario presented higher dispersion, exceeding \qty{1}{\percent}. After a more detailed analysis, we identified the presence of outlier points whose values were higher than 1.5 times the interquartile range (IQR) above the third quartile.

Furthermore, when analyzing the PF's standard deviations, we can observe that they are consistently high across all scenarios and investigations and slightly higher in the experimental data scenario. However, compared to the PINN, the PF standard deviations are significantly higher, suggesting that the PINN provides more accurate and consistent estimates.

It is crucial to note that the higher MAPE values observed in the experimental scenarios for both methods can be related to the model's underlying assumptions. Specifically, this includes the usage of a single effective viscosity, as outlined in \cref{subsec:esp-model}, and comparing viscosity with the \citet{brinkman1952} model. This model might not perfectly represent the complexity of the actual system, potentially skewing the error metrics.

\subsubsection{Case 2: Seven unknown parameters $B$, $\mu$, $\rho$, $k_u$, $k_d$, $k_{3p}$ and $k_{4p}$ for different types of data}\label{subsec:case2}

\noindent In this section, we examine Case 2. We treated the fluid's bulk modulus $B$, viscosity $\mu$, and density $\rho$, along with the pump's viscous flow loss coefficient $k_{3p}$, the pump equivalent resistance $k_{4p}$, and the equivalent resistances of the upstream and downstream pipelines $k_u$ and $k_d$, respectively, as unknowns. The remaining parameters were considered known. The MAPE values and the corresponding standard deviations (in parentheses) from the two experimental investigations across all data scenarios are provided in \cref{tab:inv_case2}.
\begin{table}[!htb]
  \sisetup{round-mode=places, round-precision=1}
  \setlength{\tabcolsep}{4pt}  
  \small
  \caption{\textbf{Comparative mean absolute percentage error (with standard deviations) for Case 2 using PINN and PF.}}
  \centering
  \begin{tabular}{lllccccccc}
    \toprule
      &      &    & \multicolumn{7}{c}{MAPE and standard deviation ($\%$)} \\
      \cmidrule{4-10}
      Inv. & Data T. & Mtd. &                                                  $B$ &                                                $\mu$ &                                                $\rho$ &                                                $k_u$ &                                                $k_d$ &                                             $k_{3p}$ &                                             $k_{4p}$ \\
    \midrule
    \multirow{6}{*}{1} & \multirow{2}{*}{Sim.} & PINN &   \num{2.5213836327651036} (\num{4.387909342162803}) &   \num{13.102961533897536} (\num{27.19205369881111}) &    \num{1.3105662092164576} (\num{2.720901664315951}) &   \num{24.223342389290032} (\num{48.35094650639312}) &   \num{6.151542597588386} (\num{13.244094835165122}) &    \num{4.280495346543093} (\num{5.432141161902463}) &    \num{3.80066181276497} (\num{7.2083173224836825}) \\
      &      & PF &     \num{4.03807387570199} (\num{8.504889013384119}) &   \num{2.5542680848122177} (\num{4.545469252656806}) &    \num{2.491670270659405} (\num{2.9310324919475477}) &    \num{2.402737891954507} (\num{3.883871154288147}) &    \num{2.722032593773199} (\num{4.397604388129908}) &    \num{2.529795372263163} (\num{3.863838041834853}) &   \num{2.027506835621069} (\num{2.5495873606284163}) \\
    \cmidrule{2-10}
      & \multirow{2}{*}{Noisy} & PINN &  \num{1.3429040186514873} (\num{1.7908819122742459}) &    \num{36.42589635565102} (\num{39.41820724248472}) &     \num{5.681855304227679} (\num{5.613470473518714}) &    \num{40.57632313663549} (\num{40.72381315607422}) &   \num{19.066711108882664} (\num{20.71153646571213}) &      \num{9.788220859167755} (\num{9.2254125819255}) &    \num{10.505141326103331} (\num{10.5251128539542}) \\
      &      & PF &    \num{9.255907070548494} (\num{15.53026431038057}) &     \num{6.52474945410555} (\num{10.02968086332609}) &     \num{4.138514993795417} (\num{6.354835065696077}) &   \num{6.145908075188957} (\num{10.249017411099116}) &     \num{5.78366121070701} (\num{8.260789725668948}) &    \num{8.72529318540169} (\num{15.408448111897698}) &    \num{3.345797303849969} (\num{4.580757741119477}) \\
    \cmidrule{2-10}
      & \multirow{2}{*}{Exp.} & PINN & N/A &  \num{120.18706717064303} (\num{28.996414837031224}) &    \num{9.718875912541977} (\num{3.5541846452975263}) &    \num{77.35400849627331} (\num{56.46106547734819}) &    \num{56.30507022450421} (\num{10.65026318390637}) &   \num{27.676174129055294} (\num{4.885772644947763}) &    \num{29.262093899680988} (\num{4.42942275098544}) \\
      &      & PF &   N/A &   \num{12.918928396659267} (\num{18.53152850257908}) &   \num{13.533461888580891} (\num{17.092301252972295}) &    \num{5.985160401827235} (\num{6.241265502820436}) &  \num{10.112815368862714} (\num{18.035044563907764}) &   \num{9.650397569379411} (\num{14.826550216722607}) &   \num{9.031852963650982} (\num{14.952803317398665}) \\
    \cmidrule{1-10}
    \cmidrule{2-10}
    \multirow{6}{*}{2} & \multirow{2}{*}{Sim.} & PINN &  \num{0.9823345510017085} (\num{1.1995361381298189}) &   \num{1.3793030201642582} (\num{2.477931388377458}) &  \num{0.17681157226145608} (\num{0.1428153207295647}) &  \num{0.2789054854944502} (\num{0.2691537002417715}) &  \num{1.2046161021431805} (\num{2.1478117911958234}) &  \num{3.9751433250910284} (\num{1.2686959833608065}) &  \num{2.2655190369602156} (\num{1.2433632526814615}) \\
      &      & PF &   \num{5.302863740378356} (\num{11.400051332599853}) &    \num{4.519832068375908} (\num{10.45895930810271}) &    \num{3.1305221412657698} (\num{4.934669251347136}) &    \num{2.626474187651748} (\num{4.725513731806687}) &   \num{4.771929595395402} (\num{10.492828931763343}) &   \num{3.598466216605731} (\num{7.7607353013035185}) &   \num{3.7342820732305104} (\num{7.977822358368949}) \\
    \cmidrule{2-10}
      & \multirow{2}{*}{Noisy} & PINN &  \num{1.2875478059261767} (\num{1.3055661331966981}) &   \num{13.010727545257499} (\num{24.10933721489435}) &    \num{3.1028807904373883} (\num{4.927883570221072}) &    \num{3.624120697748016} (\num{7.052904086646576}) &  \num{12.618720004236595} (\num{22.212623875796833}) &    \num{6.448300466498422} (\num{6.497575101824763}) &     \num{6.56768429052131} (\num{8.680814029054691}) \\
      &      & PF &   \num{11.57931273161333} (\num{15.497960666342431}) &  \num{10.727100522050266} (\num{16.261846040600396}) &     \num{7.607611155603845} (\num{9.285237592891319}) &   \num{9.889203113685568} (\num{13.135483017166731}) &  \num{10.800591940880558} (\num{16.579728826831033}) &   \num{8.404201907098843} (\num{12.140500035779139}) &     \num{6.6587268482773565} (\num{9.7252237315603}) \\
    \cmidrule{2-10}
      & \multirow{2}{*}{Exp.} & PINN &  N/A &    \num{14.49782816158458} (\num{20.93775780898413}) &     \num{7.994798599578138} (\num{5.118943537830622}) &   \num{4.818419157758497} (\num{4.3363164923481055}) &  \num{15.393800325496779} (\num{14.845421842893636}) &     \num{9.72735121273275} (\num{5.922143286957163}) &    \num{5.734447186165114} (\num{7.987802584070322}) \\
      &      & PF &    N/A &    \num{5.978617539283189} (\num{8.758262224170489}) &     \num{8.982345114186755} (\num{12.03786935099393}) &   \num{7.677165432517409} (\num{11.235119877288664}) &   \num{7.550778280223338} (\num{11.251304837170762}) &    \num{6.507999933425948} (\num{8.606920514327095}) &    \num{6.137110330914043} (\num{8.128209207209826}) \\
    \bottomrule
    \end{tabular}
  \label{tab:inv_case2}
\end{table}

The results in \cref{tab:inv_case2} indicate that the PINNs estimated flow parameters in most scenarios with low MAPE value. However, exceptions were noted in the first investigation's simulated-with-noise and experimental data scenarios, where the MAPE values were significantly higher than those of Investigation 2. Notably, in the experimental data scenario, the viscosity ($\mu$) MAPE value reached as high as \qty{120.19}{\percent}, with other parameters exhibiting MAPE values greater than \qty{27}{\percent}. In contrast, for the second investigation, the MAPE values in all scenarios remained below \qty{15.5}{\percent}. Moreover, in the simulated case of the first investigation, the equivalent resistance of the upstream pipeline ($k_u$) exhibited high MAPE values, a discrepancy not observed in other parameters or the second investigation.

Contrary to Case 1, discussed in \cref{subsec:case1}, the PF generally exhibited MAPE values similar to those of the PINN, except in the data scenarios previously noted where the PINN's MAPE values were relatively high. In these scenarios, the PF presented consistently lower MAPE values than the PINN. Furthermore, it is noteworthy that in the simulated data scenario of Investigation 2, the PINN showed slightly smaller values than the PF. Similarly to the PINN, for the experimental data, the results for the PF were better in the second investigation than in the first. However, the differences were not as pronounced as those observed for the PINN, with the highest MAPE value being approximately \qty{13.5}{\percent} in the first investigation and \qty{9.0}{\percent} in the second. It is also evident that PF performance deteriorated when noise was added to the experimental data: for simulated data, the MAPE values were below \qty{5.5}{\percent}, whereas for the noisy and experimental scenarios, they were \qty{11.6}{\percent} and \qty{13.5}{\percent}, respectively.

The standard deviations of the absolute percentage errors for the PINN indicated within parentheses in \cref{tab:inv_case2} were significantly higher than Case 1, discussed in \cref{subsec:case1}. For example, the equivalent resistance of the upstream pipeline ($k_u$) in the first investigation reached almost \qty{49}{\percent}, whereas most other values were smaller than \qty{5}{\percent} for the Case 1. However, particularly for the second investigation under the simulated data scenario, the values were below \qty{2.5}{\percent}. 

Similar to the findings in \cref{subsec:case1}, a detailed analysis revealed the presence of relatively large outliers—values that lay more than 1.5 times the IQR above the third quartile. When examining the median values of the absolute percentage errors for simulated scenarios in both investigations, they were predominantly found to be below \qty{1}{\percent}. This indicates that outliers skewed the results, suggesting that the PINN is not as consistent in Case 2 as it was in Case 1.

Furthermore, when analyzing the standard deviations of the PF, we observe that in both experimental investigations, the values increased from the simulated to the noisy scenarios. It is also noteworthy that although the PF's MAPE values for the experimental data scenario of the first investigation were smaller than those of the PINN, the standard deviations were relatively high, with most parameters exhibiting values higher than \qty{14.5}{\percent}. This suggests that, despite being more accurate, the PF results were not consistent. Additionally, compared to Case 1, the overall standard deviation values increased. This indicates that the introduction of more unknown parameters adversely affected the PF's performance.

It is worth noting that the parameter selection for this case did not include a practical identifiability analysis, which in this study would have assessed parameter sensitivity to noise. This could explain the observed decrease in PINN and PF performance. Furthermore, the performance of both models was poorer in the first experimental investigation compared to the second, especially in the experimental data scenarios. This observation is supported by the higher standard deviations recorded for the PF in the first investigation's experimental data.

\FloatBarrier
\subsubsection{Case 3: Eight unknown parameters $B$, $\mu$, $\rho$, $k_u$, $k_d$, $k_{4p}$, $k_{1s}$ and $k_{5s}$ for different types of data}\label{subsec:case3}

\noindent In this section, we present the results of the inverse problem for Case 3, specifically evaluating the performance of the parameter estimation. The evaluation is based on the MAPE and the corresponding standard deviations (in parentheses), as shown in \cref{tab:inv_case3}. The considered parameters, in this case, include the bulk modulus ($B$), viscosity ($\mu$), and density ($\rho$) of the fluid, as well as the pump equivalent resistance ($k_{4p}$), the first impeller-fluid coupling coefficient ($k_{1s}$), the shaft second-order friction coefficient ($k_{5s}$), and the equivalent resistance of the pipeline upstream and downstream ($k_u$ and $k_d$).
\begin{table}[!htb]
  \sisetup{round-mode=places, round-precision=1}
  \setlength{\tabcolsep}{2pt}  
  \small
  \caption{\textbf{Comparative mean absolute percentage error for Case 3 using PINN and PF.}}
  \centering
  \begin{tabular}{lllcccccccc}
    \toprule
      &      &    & \multicolumn{8}{c}{MAPE and standard deviation ($\%$)} \\
      \cmidrule{4-11}
      Inv. & Data T. & Mtd. &                                                   $B$ &                                                $\mu$ &                                               $\rho$ &                                                $k_u$ &                                                $k_d$ &                                             $k_{4p}$ &                                             $k_{1s}$ &                                             $k_{5s}$ \\
    \midrule
    \multirow{6}{*}{1} & \multirow{2}{*}{Sim.} & PINN &      \num{14.948393812003522} (\num{0.1391289229468987}) &   \num{11.307134906412633} (\num{7.293375557134734}) &   \num{3.024805096842583} (\num{2.0768225473905195}) &   \num{26.26571308093323} (\num{15.696354879782545}) &   \num{3.2450799503453087} (\num{2.745995319765503}) &  \num{1.7870117025733796} (\num{1.7088466565532832}) &    \num{6.915994745528418} (\num{4.549545752180401}) &  \num{12.238812343580292} (\num{11.161682685434615}) \\
      &      & PF &       \num{2.6223428950334653} (\num{2.812365077584566}) &    \num{2.740982278756182} (\num{6.625646506092324}) &  \num{1.1169298312379348} (\num{1.4783306825181188}) &   \num{3.2923516201376684} (\num{9.749117989238362}) &   \num{2.8661286010586693} (\num{7.370765732836711}) &    \num{2.2913908704096144} (\num{5.11027035507782}) &  \num{1.6083464079401522} (\num{1.8434668493287447}) &   \num{2.296122266799559} (\num{3.9134060340535486}) \\
    \cmidrule{2-11}
      & \multirow{2}{*}{Noisy} & PINN &       \num{14.506395405134997} (\num{2.694645461986955}) &   \num{10.410154579252955} (\num{5.446287735472595}) &  \num{1.8768204707887575} (\num{1.4279313717772126}) &   \num{25.96860009670603} (\num{14.273776146505766}) &   \num{4.988127891202737} (\num{3.0763038908609475}) &   \num{2.057263822405098} (\num{1.2366811858478661}) &   \num{6.878481075721019} (\num{4.4072746908221205}) &     \num{6.000178140461784} (\num{5.17994719438864}) \\
      &      & PF &       \num{7.545996798360313} (\num{14.389433904699514}) &  \num{7.1218772508421475} (\num{11.560524138687686}) &    \num{4.230532969555363} (\num{10.80469895854611}) &    \num{6.841633419058934} (\num{13.63186731895848}) &    \num{6.77323942876557} (\num{12.801886971188534}) &    \num{3.219514832813439} (\num{5.018571659620556}) &    \num{7.052672647179097} (\num{15.81034055022838}) &   \num{7.079095942790262} (\num{12.831436577308756}) \\
    \cmidrule{2-11}
      & \multirow{2}{*}{Exp.} & PINN &  N/A &   \num{13.052403078428917} (\num{6.413757522137849}) &   \num{9.255173033011172} (\num{2.1193718172997436}) &   \num{34.50834145300853} (\num{15.817923386210257}) &    \num{16.00668914180382} (\num{2.009829555813336}) &    \num{5.87474192124567} (\num{1.1945235473269455}) &    \num{12.3686559173505} (\num{6.0897642254101125}) &   \num{7.759101924557011} (\num{3.6060770944265523}) \\
      &      & PF &       N/A &   \num{7.238743086047252} (\num{11.965519788639615}) &   \num{6.540665845768402} (\num{10.653902743729205}) &   \num{7.848396247471131} (\num{13.073608434309685}) &      \num{4.97105516703516} (\num{9.84880296972412}) &   \num{7.140532319745466} (\num{11.465366891967813}) &     \num{6.1592408204334435} (\num{11.179762089327}) &   \num{7.206891362982573} (\num{12.832600972157781}) \\
    \cmidrule{1-11}
    \cmidrule{2-11}
    \multirow{6}{*}{2} & \multirow{2}{*}{Sim.} & PINN &    \num{14.999129436145031} (\num{0.004026205563665935}) &    \num{14.336275224657454} (\num{14.3134550186948}) &   \num{4.119317099646765} (\num{2.8218806983709612}) &     \num{6.900439504079831} (\num{8.05322948904196}) &    \num{6.655014737929068} (\num{7.667611831514188}) &   \num{3.4281873280701864} (\num{3.494575683664538}) &  \num{12.706454478948592} (\num{11.072490429870568}) &     \num{6.96440678514521} (\num{3.929910827578125}) \\
      &      & PF &       \num{7.730572690568638} (\num{14.930697694819406}) &    \num{6.9668758675513995} (\num{9.99481273937604}) &  \num{6.4293308637421145} (\num{11.413590565043469}) &  \num{10.713553092862249} (\num{21.374813210278326}) &   \num{4.9772324038537885} (\num{8.668533884631145}) &    \num{5.030114066250476} (\num{9.649297054225341}) &   \num{8.995660164878378} (\num{19.535351678212464}) &   \num{8.737432903753488} (\num{13.883187935347355}) \\
    \cmidrule{2-11}
      & \multirow{2}{*}{Noisy} & PINN &    \num{14.99978715592548} (\num{0.0007848589279220654}) &   \num{13.658673159887488} (\num{8.003053918764916}) &  \num{3.7653651802310963} (\num{3.0649208275370228}) &     \num{4.91554888888324} (\num{8.583009052116708}) &     \num{13.63428123347513} (\num{8.10408638898512}) &    \num{7.042093176433951} (\num{4.000652476670067}) &   \num{11.863419587322072} (\num{9.215161224716065}) &   \num{12.215725252776759} (\num{9.053991032948176}) \\
      &      & PF &       \num{9.780181962431095} (\num{14.702263661110855}) &    \num{8.306697794777506} (\num{10.72619929581033}) &    \num{8.20002153546969} (\num{10.798517138736539}) &   \num{13.037991364942707} (\num{22.14472492482943}) &    \num{6.314975166897598} (\num{7.825659668048877}) &   \num{6.633444694395772} (\num{11.054140958044488}) &  \num{11.095410598786271} (\num{20.963519585794963}) &  \num{11.208168220964653} (\num{18.624735906891935}) \\
    \cmidrule{2-11}
      & \multirow{2}{*}{Exp.} & PINN &   N/A &  \num{1.5967748423505703} (\num{1.2239031997119865}) &  \num{10.158213110725832} (\num{1.2534241241532886}) &  \num{1.9600652119559312} (\num{0.5339941889585855}) &  \num{12.052400641918984} (\num{2.0248664840645176}) &   \num{5.281842999816811} (\num{0.8925806099063434}) &    \num{7.00733773341248} (\num{3.3526932721643163}) &   \num{9.246945308406612} (\num{3.0564103097400515}) \\
      &      & PF &       N/A &  \num{16.662637615768986} (\num{12.762840625291945}) &   \num{16.27705426595193} (\num{13.756906636650541}) &   \num{27.356689108740795} (\num{29.23421039377853}) &  \num{15.153415967029028} (\num{13.609422817142724}) &   \num{13.77723114089023} (\num{14.246818838717585}) &   \num{17.387751131864025} (\num{16.77147896821076}) &  \num{13.727834829470389} (\num{15.018331690090962}) \\
    \bottomrule
    \end{tabular}
    
  \label{tab:inv_case3}
\end{table}

The results in \cref{tab:inv_case3} indicate that the PINN could estimate the parameters for Case 3. However, exceptions were noted in the first investigation across all data scenarios. Notably, the equivalent resistance of the upstream pipeline ($k_u$) exhibited MAPE values higher than \qty{26}{\percent} in different data scenarios, a discrepancy not observed in the second investigation. Additionally, the PINN was unable to estimate the bulk modulus parameter ($B$) for any scenario or investigation. The \qty{15}{\percent} value noted in \cref{tab:inv_case3} represents the upper and lower bounds imposed on the output scaling transformation. This finding is further supported by the overall low standard deviations, which were lower than \qty{0.15}{\percent}, indicating that the estimated value reached these bounds for most initializations.

Similar to Case 2, discussed in \cref{subsec:case2}, the PF generally exhibited MAPE values comparable to those of the PINN. However, in the experimental data scenario of the first investigation, the PINN presented mostly MAPE values higher than the PF, while in the second investigation, it was the PF that demonstrated higher MAPE values, with the PINN performing better across all parameters. The lowest MAPE value for the experimental scenario of the second investigation was \qty{13.8}{\percent} for the PF, whereas the highest for the PINN was \qty{12.1}{\percent}. However, in the simulated scenario of the first investigation, the PF slightly outperformed the PINN. Moreover, unlike the PINN, the PF did not encounter issues in estimating the bulk modulus ($B$) or the equivalent resistance of the upstream pipeline ($k_u$) in the first investigation.

Overall, the PINN demonstrated significantly lower standard deviations of the absolute percentage error than those observed in Case 2, discussed in \cref{subsec:case2}. The highest standard deviation value across all scenarios and investigations was \qty{15.8}{\percent} for the $k_u$ parameter of the first experimental investigation, while in Case 2, it reached \qty{56.5}{\percent}. In contrast, the PF consistently exhibited high standard deviation values across all data scenarios and experimental investigations, except for the simulated scenario of the first investigation, where it was similar to that of the PINN. Notably, although the PF achieved small MAPE values for the $k_u$ parameter in most scenarios, the standard deviation for this parameter remained consistently high. Thus, despite achieving MAPE values similar to those of the PINN, the PF results were less consistent.

The difficulty encountered by the PINN and the high standard deviation associated with the $k_u$ parameter in the PF is likely due to its relatively high correlation with other parameters, such as $k_d$, as shown in \cref{fig:practical_corr_8}. Moreover, it is noteworthy that in Case 3, besides the fluid bulk modulus and density, other parameters are also subject to bounded output scaling transformations. This could have contributed to the observed variations in standard deviation and MAPE values.

\subsection{Estimated unknown parameters mean and standard deviation}\label{apdx:pinn_val}

\noindent In this section, we present additional results from the PINN and PF methods for parameter estimation in the ESP system model. These results, obtained from averaging the parameter estimations over 30 runs for each method, complement the findings discussed in \cref{subsec:inv}. The standard deviations are presented alongside mean estimates, denoted by the $\pm$ symbol. The \cref{tab:exp1_meanstd} and \cref{tab:exp3_meanstd} showcase the results from the first and second experimental investigations, respectively.

The tables illustrate how effectively PINN and PF estimated the unknown parameters across different data scenarios and experimental investigations for each of the three cases. In cases where certain parameters were not considered, the corresponding cells are left blank. While these tables are not the focus of detailed discussion within this work, they are included for completeness and may provide a valuable reference for future research.


\begin{landscape}
  \begin{table}[H]
    \caption{\textbf{Estimation results of ESP system model parameters using PINN and PF for Investigation 1.}}
    \sisetup{
      detect-all,
      round-mode = places,  
      round-precision = 2,   
      separate-uncertainty=true,
      scientific-notation=true
    }
    \sisetup{output-exponent-marker=\ensuremath{\mathrm{E}}}
    \begin{adjustbox}{max width=\linewidth}
      \begin{tabular}{cccccccccccc}
        \toprule
        \multirow{2}{*}{\textbf{Params.}} & \multirow{2}{*}{\textbf{True}} & & \multicolumn{3}{c}{\textbf{Case 1}} & \multicolumn{3}{c}{\textbf{Case 2}} & \multicolumn{3}{c}{\textbf{Case 3}} \\
        \cmidrule(l){4-6} \cmidrule(l){7-9} \cmidrule(l){10-12}
        & & & \textbf{Simulated} & \textbf{Noisy} & \textbf{Experimental} & \textbf{Simulated} & \textbf{Noisy} & \textbf{Experimental} & \textbf{Simulated} & \textbf{Noisy} & \textbf{Experimental} \\
        \midrule
        \multirow{2}{*}{$B$} & \multirow{2}{*}{\num{1310000000.0}} & PINN &       \num{1305912431.090448} $\pm$ \num{2536730.0825917125} &       \num{1302456398.0399926} $\pm$ \num{2757001.281866439} &         N/A &       \num{1335169614.256528} $\pm$ \num{61460480.30452725} &      \num{1311763579.820222} $\pm$ \num{29450497.01152904} &    N/A &         \num{1114176041.062754} $\pm$ \num{1822588.8906043724} &          \num{1120175139.9626386} $\pm$ \num{36444153.614120096} &          N/A  \\
                 &                              & PF &        \num{1279472010.749351} $\pm$ \num{308349352.3414486} &       \num{1236387809.369192} $\pm$ \num{280594129.22623354} &      \num{1206412321.8807626} $\pm$ \num{338321876.32997787} &     \num{1347184589.2652671} $\pm$ \num{117802754.31642234} &    \num{1347250051.4348373} $\pm$ \num{234870518.13727567} &    \num{1341147324.3152232} $\pm$ \num{271850764.18749493} &         \num{1316076200.1938286} $\pm$ \num{50397809.06812913} &          \num{1375784428.4159722} $\pm$ \num{202891145.09515977} &          \num{1358745005.8730352} $\pm$ \num{210118267.87722465} \\
        \cmidrule{1-12}
        \multirow{2}{*}{$\mu$} & \multirow{2}{*}{\num{0.14293628356925767}} & PINN &  \num{0.14297384425423093} $\pm$ \num{3.861040970937555e-05} &  \num{0.1429233305164508} $\pm$ \num{0.00021040215264327353} &  \num{0.14551232530506222} $\pm$ \num{0.0005729711811068929} &   \num{0.16116370364367488} $\pm$ \num{0.03911315951115801} &  \num{0.19423614667418096} $\pm$ \num{0.05706512720025046} &  \num{0.31472721071386217} $\pm$ \num{0.04144639773637725} &     \num{0.12737711495138695} $\pm$ \num{0.011334018806977935} &        \num{0.12811983352974904} $\pm$ \num{0.00790890156081042} &        \num{0.12427966369247198} $\pm$ \num{0.00916758663928755} \\
                 &                              & PF &   \num{0.14555008655609178} $\pm$ \num{0.007598794445389201} &    \num{0.1434601423071313} $\pm$ \num{0.004258339818011782} &     \num{0.146453624093867} $\pm$ \num{0.021681429555690208} &  \num{0.14453665930834522} $\pm$ \num{0.007304271227081017} &   \num{0.1469254479138905} $\pm$ \num{0.01670443528415077} &  \num{0.13883671817615556} $\pm$ \num{0.03220230099650161} &      \num{0.1412246798350288} $\pm$ \num{0.010126094015683757} &       \num{0.14075835274080417} $\pm$ \num{0.019373752609350335} &       \num{0.13603585907270077} $\pm$ \num{0.018815016715836273} \\
        \cmidrule{1-12}
        \multirow{2}{*}{$\rho$} & \multirow{2}{*}{\num{882.2053015499059}} & PINN &       \num{882.041999975824} $\pm$ \num{0.12592760499141092} &       \num{882.2105995796587} $\pm$ \num{0.9680714762487345} &       \num{875.1839354840548} $\pm$ \num{1.8414608048397414} &       \num{891.5039854175892} $\pm$ \num{25.00057431433288} &      \num{927.3635492487643} $\pm$ \num{54.24107277587715} &      \num{936.6688531620358} $\pm$ \num{74.29337202296773} &         \num{906.4443724515028} $\pm$ \num{21.553213069569694} &            \num{879.8098168685591} $\pm$ \num{20.88915003213918} &            \num{800.5556743850642} $\pm$ \num{18.69721053177292} \\
                 &                              & PF &        \num{863.0468318674306} $\pm$ \num{52.13808508246922} &        \num{878.0743480094588} $\pm$ \num{34.83044195920028} &        \num{867.6327587728952} $\pm$ \num{88.59302938764762} &      \num{875.1858547413002} $\pm$ \num{33.429101667784636} &      \num{861.4569353140613} $\pm$ \num{63.84861792257252} &      \num{770.9836096510376} $\pm$ \num{157.1202176614434} &         \num{877.7375959451481} $\pm$ \num{15.807736979404579} &            \num{888.1824354500346} $\pm$ \num{102.4199002764289} &             \num{829.948043364596} $\pm$ \num{97.22830952116395} \\
        \cmidrule{1-12}
        \multirow{2}{*}{$k_u$} & \multirow{2}{*}{\num{51.024099349975586}} & PINN &                                                              &                                                              &                                                              &       \num{59.98088029093978} $\pm$ \num{26.14729221765321} &      \num{50.76228961991519} $\pm$ \num{29.58239957733802} &     \num{21.533104565956375} $\pm$ \num{39.26537252693291} &         \num{39.23497897030772} $\pm$ \num{10.303873453418325} &          \num{40.023161022846566} $\pm$ \num{10.462903683855272} &          \num{36.699570991693996} $\pm$ \num{13.175343836126162} \\
                 &                              & PF &                                                              &                                                              &                                                              &      \num{51.46146902947204} $\pm$ \num{2.2987237391611304} &     \num{51.738357779798896} $\pm$ \num{6.082147447869667} &     \num{51.394892316393516} $\pm$ \num{4.432478548092219} &           \num{52.3757378836272} $\pm$ \num{5.076817809294529} &            \num{51.74133407870649} $\pm$ \num{7.775204786468458} &          \num{53.488070441721426} $\pm$ \num{7.4031891499411255} \\
        \cmidrule{1-12}
        \multirow{2}{*}{$k_d$} & \multirow{2}{*}{\num{30.669759377918858}} & PINN &                                                              &                                                              &                                                              &       \num{28.79930494534356} $\pm$ \num{4.069679390568045} &      \num{24.83643076632115} $\pm$ \num{6.365846852291596} &    \num{13.401129822495177} $\pm$ \num{3.2664100916191634} &        \num{31.240421511035542} $\pm$ \num{1.1819876995630012} &           \num{32.09401495319657} $\pm$ \num{1.1013020869980192} &          \num{35.578972422081556} $\pm$ \num{0.6164098886742462} \\
                 &                              & PF &                                                              &                                                              &                                                              &     \num{31.154065736810313} $\pm$ \num{1.5157298220113329} &    \num{31.843247164469638} $\pm$ \num{2.8721733127243034} &     \num{30.871227375262503} $\pm$ \num{6.364340845969781} &          \num{30.94144702609245} $\pm$ \num{2.415218888966917} &            \num{30.88343254535254} $\pm$ \num{4.453402469713384} &          \num{29.968142516921525} $\pm$ \num{3.3195480613527057} \\
        \cmidrule{1-12}
        \multirow{2}{*}{$k_{3p}$} & \multirow{2}{*}{\num{-63059695.62392379}} & PINN &                                                              &                                                              &                                                              &      \num{-62330730.51383852} $\pm$ \num{4326842.543347514} &     \num{-58144916.80146923} $\pm$ \num{6947522.573239048} &      \num{-45607184.45779439} $\pm$ \num{3080953.35878099} &                                                                &                                                                  &                                                                  \\
                 &                              & PF &                                                              &                                                              &                                                              &       \num{-63196666.82382781} $\pm$ \num{2924024.94548367} &    \num{-66079229.38174006} $\pm$ \num{10784075.501739899} &    \num{-65000744.88467144} $\pm$ \num{11037548.185901385} &                                                                &                                                                  &                                                                  \\
        \cmidrule{1-12}
        \multirow{2}{*}{$k_{4p}$} & \multirow{2}{*}{\num{-2738928.7728844346}} & PINN &                                                              &                                                              &                                                              &      \num{-2635020.81851213} $\pm$ \num{197533.89864339965} &    \num{-2451957.9778547054} $\pm$ \num{289055.4377191409} &    \num{-1937460.863517611} $\pm$ \num{121318.73419942951} &          \num{-2776041.45040252} $\pm$ \num{56956.08415639395} &         \num{-2788488.0958259804} $\pm$ \num{43484.997658918786} &          \num{-2899833.769698136} $\pm$ \num{32717.149136617558} \\
                 &                              & PF &                                                              &                                                              &                                                              &      \num{-2737279.632147623} $\pm$ \num{89798.36207957922} &    \num{-2713241.183412259} $\pm$ \num{154096.77327665125} &    \num{-2922723.438272043} $\pm$ \num{442818.45584028313} &       \num{-2758111.8761087363} $\pm$ \num{152592.82027845472} &          \num{-2763411.355858637} $\pm$ \num{162227.37089220283} &         \num{-2861049.6122050495} $\pm$ \num{350362.86636520043} \\
        \cmidrule{1-12}
        \multirow{2}{*}{$k_{1s}$} & \multirow{2}{*}{\num{-61.26078341712889}} & PINN &                                                              &                                                              &                                                              &                                                             &                                                            &                                                            &         \num{-64.25942555244028} $\pm$ \num{4.127387588928234} &           \num{-58.68053111441437} $\pm$ \num{4.332532630364442} &         \num{-53.735749920157865} $\pm$ \num{3.8381819644743658} \\
                 &                              & PF &                                                              &                                                              &                                                              &                                                             &                                                            &                                                            &        \num{-61.42258268761751} $\pm$ \num{1.5008498097140548} &          \num{-63.17269024670768} $\pm$ \num{10.456520531847302} &           \num{-62.48418829824325} $\pm$ \num{7.751505008078811} \\
        \cmidrule{1-12}
        \multirow{2}{*}{$k_{5s}$} & \multirow{2}{*}{\num{0.00014230344668551565}} & PINN &                                                              &                                                              &                                                              &                                                             &                                                            &                                                            &  \num{0.0001590218531141323} $\pm$ \num{1.664110550054707e-05} &  \num{0.00014204682869629876} $\pm$ \num{1.1387993059442725e-05} &   \num{0.00013126197721502884} $\pm$ \num{5.131571995505879e-06} \\
                 &                              & PF &                                                              &                                                              &                                                              &                                                             &                                                            &                                                            &  \num{0.0001413479970236867} $\pm$ \num{6.411931450317124e-06} &   \num{0.0001432459979319357} $\pm$ \num{2.0915875593329238e-05} &  \num{0.00014376105763946144} $\pm$ \num{2.0978085097801908e-05} \\
        \bottomrule
      \end{tabular}
    \end{adjustbox}
    \label{tab:exp1_meanstd}
  \end{table}
  \begin{table}[H]
    \caption{\textbf{Estimation results of ESP system model parameters using PINN and PF for Investigation 2.}}
    \sisetup{
      detect-all,
      round-mode = places,  
      round-precision = 2,   
      separate-uncertainty=true,
      scientific-notation=true
    }
    \sisetup{output-exponent-marker=\ensuremath{\mathrm{E}}}
    \begin{adjustbox}{max width=\linewidth}
      \begin{tabular}{cccccccccccc}
        \toprule
        \multirow{2}{*}{\textbf{Params.}} & \multirow{2}{*}{\textbf{True}} & & \multicolumn{3}{c}{\textbf{Case 1}} & \multicolumn{3}{c}{\textbf{Case 2}} & \multicolumn{3}{c}{\textbf{Case 3}} \\
        \cmidrule(l){4-6} \cmidrule(l){7-9} \cmidrule(l){10-12}
        & & & \textbf{Simulated} & \textbf{Noisy} & \textbf{Experimental} & \textbf{Simulated} & \textbf{Noisy} & \textbf{Experimental} & \textbf{Simulated} & \textbf{Noisy} & \textbf{Experimental} \\
        \midrule
        \multirow{2}{*}{$B$} & \multirow{2}{*}{\num{1310000000.0}} & PINN &      \num{1309474919.6886563} $\pm$ \num{17284190.143877678} &     \num{1304274584.2288857} $\pm$ \num{3742654.3100340255} &      N/A &     \num{1318493936.0506938} $\pm$ \num{18536542.053426526} &    \num{1312027542.7765584} $\pm$ \num{24136251.335142203} &       N/A                   &           \num{1113511404.3865001} $\pm$ \num{52743.29288401336} &          \num{1113502788.2573762} $\pm$ \num{10281.651955787573} &           N/A \\
                 &                              & PF &      \num{1376722327.2759852} $\pm$ \num{267489622.90693396} &       \num{1347360516.141581} $\pm$ \num{349494411.2735946} &       \num{1354173274.943563} $\pm$ \num{396934727.3572507} &      \num{1333065932.9496248} $\pm$ \num{163537131.5388472} &    \num{1324874235.5820768} $\pm$ \num{254543840.54077786} &      \num{1260021667.1601532} $\pm$ \num{130575850.8906198} &           \num{1331977730.577774} $\pm$ \num{219922642.27222517} &           \num{1329452337.380149} $\pm$ \num{231698180.10260046} &           \num{1428994055.6692817} $\pm$ \num{302684183.6902443} \\
        \cmidrule{1-12}
        \multirow{2}{*}{$\mu$} & \multirow{2}{*}{\num{0.22811801215471225}} & PINN &  \num{0.22881163676647098} $\pm$ \num{0.0030373405547033236} &  \num{0.2284339223858988} $\pm$ \num{0.0014286960411368985} &  \num{0.23793278046639096} $\pm$ \num{0.000594051878555039} &  \num{0.23115181426401307} $\pm$ \num{0.005715951383552383} &   \num{0.2573185696453281} $\pm$ \num{0.05526249449294543} &  \num{0.24684531471134744} $\pm$ \num{0.055226471136155614} &       \num{0.19557397883679445} $\pm$ \num{0.032816110154393154} &        \num{0.2571927888485142} $\pm$ \num{0.021519635812986906} &        \num{0.2288308143352755} $\pm$ \num{0.004582018420725209} \\
                 &                              & PF &   \num{0.22696943429511415} $\pm$ \num{0.006433356522986973} &    \num{0.2288165195715843} $\pm$ \num{0.00847145272625976} &  \num{0.24518227250440366} $\pm$ \num{0.026610472569850784} &  \num{0.22206932045230088} $\pm$ \num{0.025325202196360683} &  \num{0.23168806757720484} $\pm$ \num{0.04452408879388263} &      \num{0.22224277769544} $\pm$ \num{0.02357702147905815} &       \num{0.23752238624294988} $\pm$ \num{0.026260883788861992} &        \num{0.23535715589471357} $\pm$ \num{0.03026448404787998} &       \num{0.25193214054325985} $\pm$ \num{0.041899801381947314} \\
        \cmidrule{1-12}
        \multirow{2}{*}{$\rho$} & \multirow{2}{*}{\num{872.5911354844369}} & PINN &       \num{870.1128498117015} $\pm$ \num{10.545825857679224} &       \num{871.4497855512296} $\pm$ \num{5.294229904739264} &        \num{851.6633005134411} $\pm$ \num{1.94370343259329} &      \num{872.7290971170529} $\pm$ \num{1.9989378496344787} &       \num{899.2544633157602} $\pm$ \num{43.2658642501075} &        \num{828.6673802152416} $\pm$ \num{70.9499711807339} &            \num{906.9238728309327} $\pm$ \num{26.89821381899316} &            \num{894.4142978928563} $\pm$ \num{36.59697816907617} &            \num{783.9514683566254} $\pm$ \num{10.93726779738504} \\
                 &                              & PF &        \num{881.2216271702265} $\pm$ \num{36.22988189289657} &       \num{868.6238554063442} $\pm$ \num{52.86265202976269} &       \num{828.1808230360057} $\pm$ \num{69.04871285342503} &      \num{878.4978086456115} $\pm$ \num{50.891633847055424} &     \num{878.1348273569292} $\pm$ \num{105.31620436475222} &      \num{808.6661657456367} $\pm$ \num{114.72361510083648} &           \num{826.6908392579127} $\pm$ \num{104.85901996835408} &           \num{825.7473530077235} $\pm$ \num{109.11017371666144} &            \num{787.5908526305028} $\pm$ \num{166.7469258858161} \\
        \cmidrule{1-12}
        \multirow{2}{*}{$k_u$} & \multirow{2}{*}{\num{656.553858757019}} & PINN &                                                              &                                                             &                                                             &       \num{658.0143943472841} $\pm$ \num{2.094003170478333} &     \num{661.6414380372211} $\pm$ \num{51.992039527656175} &       \num{629.0504335117142} $\pm$ \num{32.60903735432731} &            \num{626.1996818137158} $\pm$ \num{62.97475549428411} &             \num{655.079089366513} $\pm$ \num{65.19803421008412} &            \num{643.6849749737685} $\pm$ \num{3.505959453145839} \\
                 &                              & PF &                                                              &                                                             &                                                             &       \num{661.9266952357997} $\pm$ \num{35.21843898057801} &     \num{694.9796840477757} $\pm$ \num{101.34665924338837} &       \num{687.4344487125611} $\pm$ \num{84.16037499689749} &            \num{697.0642840798333} $\pm$ \num{152.0368278747145} &           \num{696.3698492817662} $\pm$ \num{164.55743609946657} &           \num{752.0010995630096} $\pm$ \num{246.55360889601656} \\
        \cmidrule{1-12}
        \multirow{2}{*}{$k_d$} & \multirow{2}{*}{\num{27.025931881022082}} & PINN &                                                              &                                                             &                                                             &     \num{26.728532118188742} $\pm$ \num{0.5958923653012335} &      \num{23.63752684996178} $\pm$ \num{6.015996331409745} &      \num{26.663098405819962} $\pm$ \num{5.819445816701238} &           \num{28.677475141804525} $\pm$ \num{2.195217575125783} &          \num{23.511647465320817} $\pm$ \num{2.4630912713171127} &          \num{30.283205468534973} $\pm$ \num{0.5472390366649226} \\
                 &                              & PF &                                                              &                                                             &                                                             &      \num{27.762795209029793} $\pm$ \num{3.033241849542803} &     \num{27.559123991323688} $\pm$ \num{5.347695033113092} &     \num{26.017447284888668} $\pm$ \num{3.5358301236151704} &          \num{27.942444902999856} $\pm$ \num{2.5478088771509095} &           \num{27.65338299780544} $\pm$ \num{2.6606341476135325} &           \num{29.297959631113397} $\pm$ \num{5.053549071359416} \\
        \cmidrule{1-12}
        \multirow{2}{*}{$k_{3p}$} & \multirow{2}{*}{\num{-63059695.62392379}} & PINN &                                                              &                                                             &                                                             &      \num{-65566408.90534092} $\pm$ \num{800035.8255002708} &      \num{-62349257.6609007} $\pm$ \num{5776758.692992654} &      \num{-65847939.07659952} $\pm$ \num{6695363.107199376} &                                                                  &                                                                  &                                                                  \\
                 &                              & PF &                                                              &                                                             &                                                             &     \num{-61959162.407940894} $\pm$ \num{5293772.290548208} &     \num{-63725141.65937119} $\pm$ \num{9338498.133418888} &     \num{-64692796.963937305} $\pm$ \num{6642417.672676725} &                                                                  &                                                                  &                                                                  \\
        \cmidrule{1-12}
        \multirow{2}{*}{$k_{4p}$} & \multirow{2}{*}{\num{-2738928.7728844346}} & PINN &                                                              &                                                             &                                                             &     \num{-2676877.820125957} $\pm$ \num{34054.833879164464} &     \num{-2559044.578139135} $\pm$ \num{237761.3131623675} &     \num{-2639722.1620853087} $\pm$ \num{251401.4785272263} &          \num{-2729588.6905680075} $\pm$ \num{134875.1700073167} &         \num{-2546050.8566617533} $\pm$ \num{109575.02178663021} &          \num{-2883594.6905449997} $\pm$ \num{24447.14714591223} \\
                 &                              & PF &                                                              &                                                             &                                                             &    \num{-2814457.9364319635} $\pm$ \num{229490.02840256985} &   \num{-2754652.2190875458} $\pm$ \num{324197.79627850995} &     \num{-2808969.671145508} $\pm$ \num{271507.44327728084} &          \num{-2716792.4757535304} $\pm$ \num{298289.0079801192} &         \num{-2727370.5858155815} $\pm$ \num{354508.52404609765} &          \num{-2818321.5905948784} $\pm$ \num{541336.7133101906} \\
        \cmidrule{1-12}
        \multirow{2}{*}{$k_{1s}$} & \multirow{2}{*}{\num{-61.26078341712889}} & PINN &                                                              &                                                             &                                                             &                                                             &                                                            &                                                             &          \num{-54.731328869272474} $\pm$ \num{8.036643333850847} &           \num{-65.12503579190665} $\pm$ \num{8.429798660434741} &          \num{-57.15907389823763} $\pm$ \num{2.4243409941071206} \\
                 &                              & PF &                                                              &                                                             &                                                             &                                                             &                                                            &                                                             &          \num{-64.44613669769178} $\pm$ \num{12.811759654021232} &            \num{-65.6608077497108} $\pm$ \num{13.88142786443245} &           \num{-69.2128581851811} $\pm$ \num{12.550733370260506} \\
        \cmidrule{1-12}
        \multirow{2}{*}{$k_{5s}$} & \multirow{2}{*}{\num{0.00014230344668551565}} & PINN &                                                              &                                                             &                                                             &                                                             &                                                            &                                                             &   \num{0.0001479219687331831} $\pm$ \num{1.0011258120057928e-05} &  \num{0.00014928893702113099} $\pm$ \num{2.0691157044475828e-05} &  \num{0.00012914472479852847} $\pm$ \num{4.3493772156115394e-06} \\
                 &                              & PF &                                                              &                                                             &                                                             &                                                             &                                                            &                                                             &  \num{0.00014757854453138744} $\pm$ \num{2.2835288842961624e-05} &  \num{0.00015035095804912739} $\pm$ \num{2.9976815815118614e-05} &  \num{0.00014887552753858188} $\pm$ \num{2.8405044852007308e-05} \\
        \bottomrule
        \end{tabular}
    \end{adjustbox}
    \label{tab:exp3_meanstd}
  \end{table}
\end{landscape}

\section{Summary}\label{sec:conclusion}

\noindent The usage of multiphase flow meters (MPFMs) in the oil and gas industry encounters significant challenges due to high operational costs and degradation over time. This study proposed a physics-informed neural network (PINN) model coupled with a lumped-element model of an electrical submersible pump (ESP) system to estimate the fluid properties, system parameters, and states with limited state measurements. This approach provides an alternative method to virtual flow meters (VFM), which often rely on complex multiphase flow models and detailed knowledge of production systems. The proposed PINN method was capable of estimating the volumetric flow rates, fluid properties, and system parameters with both simulated and experimental data with relative accuracy. These estimations may provide future strategy for further research in monitoring, and oil production optimization cost-effectively.

The identifiability analysis revealed that the ESP system model is not locally structurally identifiable if only the intake and discharge pressures and geometrical parameters are considered. For local structural identifiability, we had to include additional parameters such as shaft inertia ($I_s$), fluid-impeller disk friction ($k_{3s}$), and shaft viscous damping ($k_{4s}$) as known parameters. However, to achieve practical identifiability, additional parameters were required. Thus, we selected the most critical parameters in the context of the ESP's actual operation, such as the fluid properties and pipeline equivalent resistance, as unknown. We thus considered three sets of unknown parameters that were used to assess the PINN and particle filter (PF) capabilities.

In this study, the proposed PINN model could estimate the states and unknown parameters of the ESP system under both simulated and experimental data conditions with acceptable error. However, for Case 2 (refer \cref{fig:study}), the PINN faced some challenges in the simulated and experimental data conditions for the first experimental investigation (high water fraction). In these scenarios, we observed that the PINN estimations yielded relatively high MAPE values for both state and parameter estimations. Furthermore, in the first experimental investigation, the PINN showed high MAPE values for Cases 2 and 3 across all data scenarios when estimating the upstream equivalent resistance ($k_u$). This issue, however, was not present in the second experimental investigation (low water fraction) and the PF method. It suggests that specific experimental conditions affect its accuracy, indicating areas for further refinement.

We observed that the performance of the PINN is better than the PF in state estimation, where the increase of unknown parameters significantly affected the PF performance in state estimation across all cases and data scenarios. However, for estimating the unknown parameter, the PINN performed significantly better for Case 1, while for Cases 2 and 3, the PINN and PF results were similar, except for the upstream equivalent resistance ($k_u$) in the first experimental investigation. As previously discussed, we observed that the PF might be suffering from sample impoverishment and degeneracy in Cases 2 and 3, suggesting the need to increase the number of particles to improve its performance. However, this would result in significantly higher computational costs.

Although the results of PINNs for solving the inverse problem in the context of ESP systems are promising, the PINN has some limitations. One limitation is that PINN needs to be retrained again if any operating condition or/and measured data changes. Furthermore, the accuracy of the properties estimated using PINNs depends on the accuracy of the measured states and known parameters. In the event of a faulty reading or instrument failure, the PINN's ability to provide accurate estimations is compromised. Regarding the experimental setup used, it is common to have multiple ESPs operating in parallel in actual oil production systems. 

Moreover, in our experimental apparatus, the number of ESP stages and the pipeline length are smaller than those in actual oil production systems. These differences in scale and configuration can lead to more complex behavior that was not observed in our experiments. Additionally, despite evaluating the PINN under two experimental investigations, the range of water fractions studied was limited. 

For further studies, we recommend using the Friedman test, as demonstrated in \citet{manoharam2023, zamri2022}, to assess the impact of different architectural components, optimization algorithms, and activation functions on the performance of the PINN in estimating unknown parameters and states. Furthermore, an ablation study could reveal a more effective architecture and an optimal number of data points for the known states while reducing the computational cost of training. Additionally, it could reveal the hyperparameters that most affect the accuracy of the unknown parameter and state. Moreover, we suggest investigating the transfer learning strategy for new operational conditions and measured data to reduce the computational cost and the number of data points. Furthermore, we suggest incorporating missing physics in the PINN formulation as a future study, as discussed in \citet{zou2024a}. Additional experimental investigations across a broader range of water fractions and multiple ESP are also suggested; these would enable a comprehensive evaluation of the method’s performance under different flow and system conditions.

\section*{Acknowledgment}

\noindent
We gratefully acknowledge the support from the Energy Production Innovation Center (EPIC) at the University of Campinas (UNICAMP), financially backed by the São Paulo Research Foundation (FAPESP, grants 2017/15736-3 and 2019/14597-5). We also acknowledge the financial sponsorship from Equinor Brazil and the regulatory support offered by Brazil's National Oil, Natural Gas, and Biofuels Agency (ANP) through the R\&D levy regulation. Additional acknowledgments are extended to the Center for Energy and Petroleum Studies (CEPETRO), the School of Mechanical Engineering (FEM), and the Artificial Lift and Flow Assurance Research Group (ALFA).

The work of K. Nath and G. E. Karniadakis was supported by OSD/AFOSR MURI grant FA9550-20-1-0358.

\printbibliography

\appendix

\section{Experimental procedure} \label{apdx:exp_proc}

\counterwithin{figure}{section}
\counterwithin{table}{section}

\subsection{Fluid characterization}\label{apdx:oil_visc}

\noindent The oil viscosity measurement used a rotational rheometer model HAAKE MARS III. The measurements ranged from \SIrange{10}{60}{\celsius} with a step of approximately \SI{0.5}{\celsius}. For each temperature, the oil viscosity was measured three times. The \cref{fig:visc_rheology} shows the measured viscosity as a function of the temperature. Then, using least squares, we obtained a 4\textsuperscript{th} order polynomial, \cref{eq:Tvisc}, that fits viscosity measurements to the temperature.

\begin{equation}\label{eq:Tvisc}
  \sisetup{scientific-notation=true,round-mode=figures,round-precision=4}
  \sisetup{output-exponent-marker=\ensuremath{\mathrm{E}}}
    \mu(T) = \frac{\num{0.00026436} \, T^4 - \num{0.04864}\, T^3 + \num{3.436} \, T^2 - \num{114.28} \, T + \num{1610.3}}{\num{1000}}
\end{equation}

\begin{figure*}[!htb]
  \centering
  \input{figs/appendix-a/rheology_p100l_fase1.pgf}
  \caption{The oil viscosity for a temperature range of \SIrange{10}{60}{\celsius}.}
  \label{fig:visc_rheology}
\end{figure*}

\subsection{Experimental apparatus}\label{apdx:exp_apparatus}

\noindent The equipments specifications are presented in \cref{tab:device_spec}, and the instrumentation characteristics are shown in \cref{tab:sensor_spec}.

\begin{table}[H]
    \begin{center}
    \caption{\textbf{Experimental setup components specification.} The component's specifications are separated by the flow line where they are installed.} \label{tab:device_spec}
        \begin{tabular}{lll}
          \toprule 
          Line         & Device & Model                      \\  
          \midrule 
          Oil           & Pump   & NETZSCH two-screw pump\\
                        & Motor  & WEG, \SI{45}{\kilo\watt}, \qty{1775}{RPM}\\
                        & VSD-1  & CFW 700 WEG Vectrue inverter\\
          \midrule
          Water         & Pump   & IMBIL 65160\\
                        & Motor  & WEG, \SI{22}{\kilo\watt}, \qty{3535}{RPM}\\
                        & VSD-2  & CFW 700 WEG Vectrue inverter\\
          \midrule
          Emulsion      & Pump   & Baker Hughes P100L 538 series\\
                        & Motor  & WEG, \SI{37}{\kilo\watt}, \qty{3555}{RPM}\\
                        & VSD-3  & CFW 09 WEG Vectrue inverter\\
                        & Tank   & Intelfibra, capacity \qty{12000}{\litre} \\
          \midrule
          Heat control  & Heat exchanger  & FYTERM T3480 shell and tube\\
                        & Heater          & Mecalor TMR-M-18-380/C \\
                        & Chiller         & Mecalor MSA-60-CA-380/C\\
          \midrule
          Valve         & Choke  & Fisher 657 glob valve\\
          \bottomrule 
        \end{tabular}
    \end{center}
  \end{table}


\sisetup{range-units = single}
\begin{table}[H]
  \centering
  \caption{Instrumentation range and uncertainty.}
  \begin{threeparttable}
    \begin{tabular}{lllr}
      \toprule 
        Instrument    & Model & Range & Uncertainty \\
      \midrule 
      Flow meters             & Micro Motion\treg F300S355   & \SIrange{0}{100}{\cubic\meter\per\hour} & \qty{0.2}{\percent} O.V.*\\
      Temperature sensors     & ECIL\treg PT100              & \SIrange{0}{100}{\celsius}              & 1/10 DIN \\
      Pressure transducers    & Emerson Rosemount 2088       & \SIrange{0}{200}{\kPa}                  & \qty{0.065}{\percent} F.S.* \\
      Water cut meter         & Roxar Nemko 05 ATEX 112      & \SIrange{0}{100}{\percent}              & \qty{1}{\percent} F.S. \\
      Torque  transmitter     & HBM T21WN                    & \SIrange{0}{100}{\newton\meter}         & \qty{0.1}{\percent} F.S. \\
      Encoder                 & Minipa MDT-2238A             & \SIrange{0}{166}{\Hz}                   & \qty{0.05}{\percent}F.S.\\
      \bottomrule
    \end{tabular}
    \label{tab:sensor_spec}
    \begin{tablenotes}
        \item[*]{*\small O.V. : of value; \small F.S. : of full scale.}
    \end{tablenotes}
  \end{threeparttable}
\end{table}
\sisetup{range-units = repeat}
\section{Electrical submersible pump system model}\label{apdx:model}

\noindent \citet{fctc2024} relied on experimental data collected under steady-state conditions to determine the parameters for the ESP model. Notably, due to the experimental setup's constraints related to temperature control, \citet{fctc2024} used the heat exchanger bypass valve for fine temperature control. Consequently, it was necessary to individually distinguish the equivalent resistance of the upstream pipeline and the twin-screw pump parameters for both experimental investigations 1 and 2. Therefore, the \cref{tab:model-ps} presents the parameters that are constant for both experimental investigations, and \cref{tab:model-ps2} presents the parameters that change with the experimental investigation.


\begin{table}[!htb]
  \sisetup{round-mode = figures, round-precision = 4, scientific-notation = engineering}
  \sisetup{output-exponent-marker=\ensuremath{\mathrm{E}}}
  \centering
  \caption{\textbf{Parameters for the ESP model.} The table presents the parameters used in the ESP model, including the impeller, shaft, oil pump, and pipeline coefficients. They were obtained using steady-state conditions, while fluid parameters varied based on the specific investigation conducted. Geometrical parameters, such as pipeline diameter, length, and cross-sectional area, were directly measured from the experimental setup.}
  \label{tab:model-ps}
  \begin{threeparttable}
  \begin{tabular}{lllll}
    \toprule
    \#  & Description & Symbol    & Value     & Unit \\
    \midrule
    1  & 1$^{st}$ impeller-fluid coupling and shock loss coefficient  & $k_{1p}$  &  \num{4.86767028064}  & \qty{}{\per\meter} \\
    2  & 2$^{nd}$ impeller-fluid coupling and shock loss coefficient  & $k_{2p}$  &  \num{0.00993000571}  & \qty{}{\meter\squared} \\
    3  & Pump viscous flow loss coefficient                                & $k_{3p}$  &  \num{-63059695.623}$^{*}$  & \qty{}{\per\cubic\meter} \\
    4  & Pump equivalent resistance                                   & $k_{4p}$  &  \num{-2227387.757}$^{*}$  & \qty{}{\per\meter\tothe{4}} \\
    5  & Shaft 1$^{st}$ impeller-fluid coupling coefficient                 & $k_{1s}$  &  \num{-61.26078341712889}$^{*}$  & \qty{}{\second\per\square\meter} \\
    6  & Shaft 2$^{nd}$ impeller-fluid coupling coefficient                 & $k_{2s}$  &  \num{0.007812519509782526}  & \qty{}{\meter\second} \\
    7  & Fluid-impeller disk friction constant                        & $k_{3s}$  &  \num{0.11037160345516595}  & \qty{}{\meter\tothe{2}\second} \\
    8  & Shaft viscous damping coefficient                            & $k_{4s}$  &  \num{0.0673205282579523}  & \qty{}{\kg\meter} \\
    9  & Shaft second-order friction coefficient                      & $k_{5s}$  &  \num{0.00014230344668551565}  & \qty{}{\kg\meter\second} \\
    10 & Shaft moment of inertia                                      & $I_{s}$   &  \num{0.0005084508657503108}  & \qty{}{\kg\meter\tothe{2}} \\
    11 & Fluid bulk modulus                                           & $B$       &  \num{1.31e9}  & \qty{}{\pascal} \\
    12 & Downstream pipeline diameter                                 & $d_d$     &  \num{0.0762}  & \qty{}{\meter} \\
    13 & Downstream pipeline cross-sectional area                     & $A_d$     &  \num{0.004560367311877479}  & \qty{}{\square\meter} \\
    14 & Downstream pipeline length                                   & $L_d$     &  \num{28}  & \qty{}{\meter} \\
    15 & Upstream pipeline diameter                                   & $d_u$     &  \num{0.0762}  & \qty{}{\meter} \\
    16 & Upstream pipeline cross-sectional area                       & $A_u$     &  \num{0.004560367311877479}  & \qty{}{\square\meter} \\
    17 & Upstream pipeline length                                     & $L_u$     &  \num{31.5}  & \qty{}{\meter} \\
    18 & Cross-sectional area of all impeller channels$^\dagger$      & $A_p$     &  \num{0.004896564849701392}  & \qty{}{\square\meter} \\
    19 & Impeller channel length$^\ddagger$                           & $L_p$     &  \num{0.0475}  & \qty{}{\meter} \\
    \bottomrule
  \end{tabular}
  \begin{itemize}[noitemsep,leftmargin=*]
    \item[$*$]{\small The negative sign indicates that the obtained parameter is opposite in sign to what is considered in the equations.}
    \item[$\dagger$]{\small The ESP P100L has a mixed impeller whose channel height varies from the inlet to the outlet. In this study, we assume a linear variation in channel height and approximate it using its arithmetic mean, obtained by the impeller inlet and outlet heights.}
    \item[$\ddagger$]{\small We approximate the channel length as the difference between the inner and outer diameters of the impeller.}
  \end{itemize}
  \end{threeparttable}
\end{table}
\begin{table}[!htb]
  \sisetup{round-mode = figures, round-precision = 4, scientific-notation = engineering}
  \sisetup{output-exponent-marker=\ensuremath{\mathrm{E}}}
  \centering
  \caption{\textbf{Parameters for the twin-screw pump and pipeline.} It is important to mention that the value of the downstream equivalent resistance ($k_d$) provided takes into account the losses from the valve, as depicted in the experimental setup schematics in \cref{fig:explayout}. Additionally, the values for the pipeline equivalent resistances ($k_u$ and $k_d$) also reflect manual fine-tuning adjustments made to the model.}
  \label{tab:model-ps2}
  \begin{tabular}{llllll}
    \toprule
    \#  & Description & Symbol & Inv. 1 & Inv. 2 & Unit \\
    \midrule
    20 & Fluid effective viscosity                                    & $\mu$      &  \num{0.14293628356925767}     &  \num{0.22811801215471225}  & \qty{}{\pascal\meter} \\
    21 & Fluid density                                                & $\rho$     &  \num{882.2053015499059}       & \num{872.5911354844369}     & \qty{}{\kilo\gram\per\cubic\meter} \\
    22 & Upstream pipeline resistance coefficient                     & $k_{u}$    &  \num{51.024099349975586}      &  \num{656.553858757019}     & 1 \\
    23 & Downstream pipeline resistance coefficient                   & $k_{d}$    &  \num{30.669759377918858}      &  \num{27.025931881022082}   & 1 \\
    24 & Twin-screw pump angular velocity                             & $\omega_t$ &  \num{1280.0}                  &  \num{1160.0}               & \qty{}{RPM} \\
    25 & Twin-screw pump flow-rate coefficient                         & $k_{bd}$  &  \num{162186.57714429626}      &  \num{133454.74619668655}   & \qty{}{\meter} \\
    26 & Twin-screw pump leaking coefficient                          & $k_{bl}$   &  \num{1.1441462507651596e+10}  &  \num{9.56458543221449e+9}  & \qty{}{\meter} \\
    \bottomrule
  \end{tabular}
\end{table}

\section{Detail physics loss function for the PINN model for the ESP system}\label{apdx:loss-ode}

\noindent As presented in \cref{subsec:pinn}, the PINN physics loss function for the ESP system is defined as
\begin{equation}\label{eq:apdx:physics-loss}
  \mathcal{L}^{ode}(\bm{\theta}, \bm{\Lambda}, \bm{\lambda}_r) = \sum_{s \in \Phi} m(\lambda_r^s) \mathcal{L}^{ode}_s, \quad \Phi =\{Q_p,\, \omega, \, Q_1, \, Q_2, \, P_1, \, P_2 \},
\end{equation}
where, $\bm{\theta}$ represents the neural network hyperparameters, $\bm{\Lambda}$ are the unknown parameters, and $\bm{\lambda}_r$ are the physics loss self-adaptive weights for each ESP system equation. The function $m(\cdot)$ acts as a mask and is considered to be a softplus function. Expanding \cref{eq:apdx:physics-loss} for every state, we get:
\begin{subequations}
  \begin{align}
    \begin{split}
    \mathcal{L}_{Q_p} = m(\lambda^{Q_p}_r)\,\frac{1}{N} \sum^{N}_{i = 1} r(Q_p)^2 = m(\lambda^{Q_p}_r)\,\frac{1}{N} \sum^{N}_{i = 1} \biggl( \frac{dQ_p}{dt} - \frac{(P_1 - P_2 + k_3 \, \mu \, Q_p)\,A_p}{\rho \, L_p} \\ + \frac{A_p\,(k_{1p} \, \omega \, Q_p + k_{2p} \, \omega^2 + k_{4p} \, {Q_p}^2)}{L_p} \biggr)_i^2,
    \end{split}\\
    \begin{split}
    \mathcal{L}_{\omega} = m(\lambda^{\omega}_r)\,\frac{1}{N} \sum^{N}_{i = 1} r(\omega)^2 = m(\lambda^{\omega}_r)\,\frac{1}{N} \sum^{N}_{i = 1} \biggl(\frac{d\omega}{dt} - \frac{\gamma(t) - k_{1s}\,\rho\,{Q_p}^2 - k_{2s}\,\rho\,\omega\,Q_p}{I_s} \\ - \frac{k_{3s}\,\mu\,\omega - k_{4s}\,\omega - k_{5s}\,\omega^2}{I_s} \biggr)_i^2, 
    \end{split}\\
    \begin{split}
    \mathcal{L}_{Q_1} = m(\lambda^{Q_1}_r)\,\frac{1}{N} \sum^{N}_{i = 1} r(Q_1)^2 = m(\lambda^{Q_1}_r)\,\frac{1}{N} \sum^{N}_{i = 1} \biggl(\frac{dQ_1}{dt} - \frac{(k_{bd}\omega_t - Q_1)\,k_{bl}\,\mu + P_{in} - P_1}{\rho L_u} \\ - \frac{f_f(Q_1, \mu, L_u, d_u)\, {Q_1}^2\,A_u}{\rho\,L_u} - \frac{k_u\,{Q_1}^2}{2\,L_u\,A_u} \biggr)_i^2, 
    \end{split}\\
    \begin{split}
    \mathcal{L}_{Q_2} = m(\lambda^{Q_2}_r)\,\frac{1}{N} \sum^{N}_{i = 1} r(Q_2)^2 = m(\lambda^{Q_2}_r)\,\frac{1}{N} \sum^{N}_{i = 1} \biggl( \frac{dQ_2}{dt} - \frac{(P_2 - P_{out})\,A_d}{\rho\,L_d} \\ - \frac{(f_f(Q_2, \mu, L_d, d_d)\, {Q_2}^2)\,A_d}{\rho\,L_d} \\ - \frac{k_d\,{Q_2}^2}{2\,L_d\,A_d} - \frac{Q_2\, A_d}{L_d\,C_v(a)^2\,1000} \biggr)_i^2,
    \end{split}\\
    \begin{split}
    \mathcal{L}_{P_1} = m(\lambda^{P_1}_r)\,\frac{1}{N} \sum^{N}_{i = 1} r(P_1)^2 = m(\lambda^{P_1}_r)\,\frac{1}{N} \sum^{N}_{i = 1} \left(\frac{dP_1}{dt} - \frac{(Q_1 - Q_p) \, B}{A_u \, L_u} \right)_i^2,
    \end{split}\\
    \begin{split}
    \mathcal{L}_{P_2} = m(\lambda^{P_2}_r)\,\frac{1}{N} \sum^{N}_{i = 1} r(P_2)^2 = m(\lambda^{P_2}_r)\,\frac{1}{N} \sum^{N}_{i = 1} \left(\frac{dP_2}{dt} - \frac{(Q_p - Q_2) \, B}{A_d \, L_d} \right)_i^2, \\
    \end{split}
  \end{align}
\end{subequations}
where $N$ denotes the collocation point, $r(\cdot)$ represents the residual, and the subscript $i$ indicates that the state variables are evaluated at \(t = t_i\).
\section{Particle filter}\label{subsec:pf}

\noindent The PF is a widely used method for state estimation in non-linear dynamic systems. It conducts a Sequential Monte Carlo (SMC) estimation using point mass representation to approximate probability densities and estimate unknown states and parameters. For a detailed discussion on PF, we refer readers to \citet{simon2006, ristic2004}.

To estimate the unknown parameters alongside system states, we augment the state vector, $\bm{X}_t$, with a set of unknown parameters, $\bm{\Lambda}_t$, at each time instant denoted by the subscript $t$. This augmented state vector is represented as:
\begin{equation}
  \bm{U}_t = [\bm{X}_t, \bm{\Lambda}_t]^T.
\end{equation}
The set of measured states is denoted by $\bm{Y}_t$. For the PF, we employed the Sequential Importance Resampling (SIR) method, which triggers resampling when the effective number of samples, $N_{\text{eff}}$, falls below a predefined threshold of $50$. The $N_{\text{eff}}$ is defined as:
\begin{equation}
  N_{\text{eff}} = \frac{1}{\sum_{i=1}^{N} {w_i}^2},
\end{equation}
where $w_i$ is the normalized weight for the $i$-th particle. For the present study, we considered $N = 200$ particles to keep the computational cost similar to the training of the PINN. The particles were initialized with a uniform distribution across a range of $\pm 50\%$ of the true value. The PF implementation is detailed in the \cref{alg:pf}.
\begin{algorithm}
\caption{Particle filter procedure}\label{alg:pf}
\begin{algorithmic}[1]
\State \textbf{Initialization:} At $k=0$, initialize particles $\{\bm{U}_0^i \}_{i=1}^N$ from the initial distribution $p(\bm{U}_0)$.
\For{$k = 1, 2, \ldots$}
    \State \textbf{Prediction step:}
    \For{$i = 1$ to $N$}
        \State Propagate each particle using \cref{eq:esp-model}:
        \State $\bm{U}_k^i = f(\bm{U}_{k-1}^i, 
        \gamma_{k-1}, z_{k-1}^i)$ where $z_{k-1}^i \sim p(z_{k-1})$ represents the process noise
    \EndFor
    \State \textbf{Update step:}
    \For{$i = 1$ to $N$}
        \State Calculate likelihoods based on measurements $\bm{Y}_k$:
        \State $q_i = p(\bm{Y}_k \mid \bm{U}_k^i)$
    \EndFor
    \State Normalize the likelihoods:
    \State $w_i = \frac{q_i}{\sum_{j=1}^N q_j}$
    \If{$N_{\text{eff}} < 50$}
      \State \textbf{Resampling step:}
      \State Resample $N$ particles from $\{\bm{U}_k^i \}$ according to normalized weights $\{ w_i \}$
    \EndIf
    \State \textbf{Estimation:}
    Compute the mean
    $\hat{\bm{U}}_k = \sum_{i=1}^{N} w_i U_k^i$
    \State \textbf{Output:} $\hat{\bm{U}}_k$, the estimated state
\EndFor
\end{algorithmic}
\end{algorithm}
We observed that for the present study, the best results were obtained when we considered the model process noise as zero, and for the resampling, we considered a normal distribution with a standard deviation of \num{0.03}.

\section{Training settings}\label{apdx:trn-cfg}

\noindent The training algorithm settings employed for each analyzed case are detailed in \crefrange{tab:adam_case1}{tab:adam_case3}. The training procedure involved multiple steps characterized by specific Adam optimization settings. The tables present the number of epochs alongside the learning rates denoted as NN for the neural network weights, PS for the ESP model's unknown parameters, and SA for the self-adaptive weights. The maximum number of epochs employing self-adaptive weights for each step is also indicated as Max. SA. It is important to note that the training process is continuous.
\begin{table}[!htb]
  \sisetup{output-exponent-marker=\ensuremath{\mathrm{E}},scientific-notation=true}
  \centering
  \caption{\textbf{Adam training settings for neural network training in Case 1 (three unknown parameters).} We applied the same training procedure to the simulated and simulated-with-noise (Sim. noisy) data scenarios. However, we had to adapt it to suit the experimental data scenario. Importantly, we maintained the settings across the investigations.}
  \label{tab:adam_case1}
  \begin{adjustbox}{max width=\textwidth}
  \begin{tabular}{cclcccc}
    \toprule
    \multirow{2}{*}{\textbf{Scenario}} & & \multicolumn{5}{c}{\textbf{Adam}} \\
    \cmidrule{3-7}
    & & \textbf{Epoch (\#)} & \textbf{NN} & \textbf{PS} & \textbf{SA} & \textbf{Max. SA}\\
    \cmidrule{1-1} \cmidrule{3-7}
    \multirow{3}{*}{\pbox{2cm}{Simulated\\ Sim. noisy}} & & 0 to 32k & \num{1.0e-3} & \num{1.0e-3} & \num{8.0e-4} & 12k \\
    & & 32k to 38k & \num{2.0e-5} & \num{1.0e-4} & \num{1.0e-6} & 12k \\
    & & 38k to 60k & \num{1.0e-5} & \num{1.0e-6} & \num{1.0e-5} & 12k \\
    \cmidrule{1-1} \cmidrule{3-7}
    \multirow{3}{*}{Experimental} & & 0 to 32k & \num{1.0e-3} & \num{5.0e-3} & \num{5.0e-4} & 32k \\
    & & 32k to 38k & \num{2.0e-5} & \num{1.0e-4} & \num{1.0e-6} & 38k \\
    & & 38k to 60k & \num{1.0e-5} & \num{1.0e-6} & \num{1.0e-5} & 60k \\
    \bottomrule
  \end{tabular}
  \end{adjustbox}
\end{table}
\begin{table}[!htb]
  \sisetup{output-exponent-marker=\ensuremath{\mathrm{E}},scientific-notation=true}
  \centering
  \caption{\textbf{Adam training settings for neural network training in Case 2 (seven unknown parameters).} We employed a consistent training process for both the simulated data and the simulated with noise data (Sim. noisy). However, adjustments were necessary for the experimental data. In the initial step of each scenario, we employed a linear learning rate schedule for both the neural network weights (NN) and the unknown ESP parameters (PS). For the learning rate of the neural network weights, denoted as $\chi_{NN}$, we initiated it at \num{1.0E-3} and linearly transitioned it to \num{1.0E-4} between epochs 4000 and 6000. Similarly, for the learning rate of the unknown ESP parameters, denoted as $\chi_{PS}$, we started at \num{1.0E-3} and transitioned it to \num{1.0E-4} between epochs 4000 and 8000.}
  \label{tab:adam_case2}
  \begin{adjustbox}{max width=\textwidth}
  \begin{tabular}{cclccccc}
    \toprule
    \multirow{2}{*}{\textbf{Scenario}} & & \multicolumn{5}{c}{\textbf{Adam}} \\
    \cmidrule{3-7}
    & & \textbf{Epoch (\#)} & \textbf{NN} & \textbf{PS} & \textbf{SA} & \textbf{Max. SA}\\
    \cmidrule{1-1} \cmidrule{3-7}
    \multirow{3}{*}{\pbox{2cm}{Simulated\\ Sim. noisy}} & &  0 to 52k & $\chi_{NN}$ & $\chi_{PS}$ & \num{5.0E-3} & 24k\\
    & & 52k to 63k & \num{2.0E-5} & \num{1.0E-4} & \num{1.0E-6} & 24k \\
    & & 63k to 83k & \num{1.0E-5} & \num{1.0E-6} & \num{1.0E-5} & 24k \\
    \cmidrule{1-1} \cmidrule{3-7}
    \multirow{3}{*}{Experimental} & & 0 to 48k & $\chi_{NN}$ & $\chi_{PS}$ & \num{5.0E-3} & 8k \\
    & & 48k to 59k & \num{2.0E-5} & \num{1.0E-4} & \num{1.0E-6} & 8k \\
    & & 59k to 79k & \num{1.0E-5} & \num{1.0E-6} & \num{1.0E-5} & 8k \\
    \bottomrule
  \end{tabular}
  \end{adjustbox}
\end{table}
\begin{table}[!htb]
  \sisetup{output-exponent-marker=\ensuremath{\mathrm{E}},scientific-notation=true}
  \centering
  \caption{\textbf{Adam training settings for neural network training in Case 3 (eight unknown parameters).} We used the same training approach for both the simulated and simulated-with-noise (Sim. noisy) data scenarios. However, we had to adjust the training settings for the experimental data scenario. It is worth noting that we kept the training settings consistent throughout our experimental investigations.}
  \label{tab:adam_case3}
  \begin{adjustbox}{max width=\textwidth}
  \begin{tabular}{cclccccccccccc}
    \toprule
    \multirow{2}{*}{\textbf{Scenario}} & & \multicolumn{5}{c}{\textbf{Adam}} \\
    \cmidrule{3-7} \cmidrule{8-9}
    & & \textbf{Epoch (\#)} & \textbf{NN} & \textbf{PS} & \textbf{SA} & \textbf{Max. SA}\\
    \cmidrule{1-1} \cmidrule{3-7} \cmidrule{8-9}
    \multirow{4}{*}{\pbox{2cm}{Simulated\\ Sim. noisy}} & & 0 to 44k & \num{5.0E-4} & \num{1.0E-2} & \num{1.0E-3} & 16k\\
    & & 44k to 70k & \num{5.0E-4} & \num{5.0E-4} & \num{5.0E-4} & 16k \\
    & & 70k to 80k & \num{2.0E-5} & \num{1.0E-5} & \num{2.0E-5} & 16k \\
    & & 80k to 90k & \num{1.0E-5} & \num{1.0E-6} & \num{1.0E-5} & 16k \\
    \cmidrule{1-1} \cmidrule{3-7} \cmidrule{8-9}
    \multirow{5}{*}{Experimental} & & 0 to 16k & \num{5.0E-4} & \num{2.0E-3} & \num{1.0E-3} & 6k\\
    & & 16k to 36.3k & \num{5.0E-4} & \num{5.0E-4} & \num{5.0E-4} & 6k \\
    & & 36300 to 41k & \num{5.0E-4} & \num{5.0E-4} & \num{5.0E-4} & 6k \\
    & & 41k to 51k & \num{2.0E-5} & \num{1.0E-5} & \num{2.0E-5} & 6k \\
    & & 51k to 61k & \num{1.0E-5} & \num{1.0E-6} & \num{1.0E-5} & 6k \\
    \bottomrule
  \end{tabular}
  \end{adjustbox}
\end{table}
The initial self-adaptive weight values significantly impacted the experimental data scenario of Case 2 and all scenarios in Case 3. These cases were more challenging due to their higher number of unknown parameters, noisy data, and model uncertainties. Although the self-adaptive weights could adapt and achieve satisfactory state and parameter estimation results, they needed further adjustment of the initial weights to improve performance and accuracy in these scenarios. The initial weights are shown in \cref{tab:initial_weights}.
\begin{table}[!htb]
  \sisetup{output-exponent-marker=\ensuremath{\mathrm{E}}, scientific-notation=true}
  \centering
  \caption{\textbf{Initial weights for the data, physics, and initial conditions (I.C.) losses.} The table presents the initial weights for data, physics, and initial conditions losses for the cases and scenarios evaluated in this work. The $Q_p$, $\omega$, $Q_1$, $Q_2$, $P_1$, and $P_2$ represent initial weights for the state corresponding ODE. The ``Data'' column is the initial weight for data loss, and the ``I.C.'' column is the initial condition weights. The ``Sim.'' case is for the simulated and simulated with noise scenarios, and ``Exp.'' is for the experimental data scenario.}
  \label{tab:initial_weights}
  \begin{adjustbox}{max width=\textwidth}
  \begin{tabular}{lcccccccc}
    \toprule
    \multicolumn{1}{c}{\textbf{Case}} & \textbf{Data} & \multicolumn{6}{c}{\textbf{Physics}} & \textbf{I.C.} \\
    \cmidrule{3-8}
     & & $Q_p$ & $\omega$ & $Q_1$ & $Q_2$ & $P_1$ & $P_2$ & \\
    \cmidrule{1-1} \cmidrule{2-2} \cmidrule{3-8} \cmidrule{9-9}
    \textbf{Case 1: Sim. } & \num{400} & \num{1.0E-6} & \num{1.0E-6} & \num{1.0E-6} & \num{1.0E-6} & \num{1.0E-6} & \num{1.0E-6} & \num{1.0E0} \\
    \textbf{Case 1: Exp.} & \num{1.0E1} & \num{1.0E-6} & \num{1.0E-6} & \num{1.0E-6} & \num{1.0E-6} & \num{1.0E-6} & \num{1.0E-6} & \num{1.0E0} \\
    \textbf{Case 2: Sim. } & \num{1.0E1} & \num{1.0E-6} & \num{1.0E-6} & \num{1.0E-6} & \num{1.0E-6} & \num{1.0E-6} & \num{1.0E-6} & \num{1.0E0} \\
    \textbf{Case 2: Exp.} & \num{1.0E-4} & \num{1.0E-6} & \num{1.0E-6} & \num{1.0E-4} & \num{1.0E-4} & \num{1.0E-6} & \num{1.0E-6} & \num{1.0E0} \\
    \textbf{Case 3: Sim. } & \num{1.0E1} & \num{1.0E-6} & \num{1.0E-6} & \num{1.0E-4} & \num{1.0E-4} & \num{1.0E-6} & \num{1.0E-6} & \num{1.0E0} \\
    \textbf{Case 3: Exp.} & \num{1.0E1} & \num{1.0E-6} & \num{1.0E-6} & \num{1.0E-4} & \num{1.0E-4} & \num{1.0E-6} & \num{1.0E-6} & \num{1.0E0} \\
    \bottomrule
  \end{tabular}
  \end{adjustbox}
\end{table}

\section{Hyperparameter optimization study}\label{apdx:ho-results}

\noindent In this section, we present a detailed numerical study on the performance of hyperparameter optimization (HO) using the Tree-structured Parzen Estimator (TPE), as described by \citet{bergstra2011algorithms}, and implemented via the Optuna Python library \cite{optuna_2019}. The study was conducted across all cases considered, which include simulated data, simulated data with added Gaussian noise, experimental data, and different water fractions. The numerical results suggest that while the HO with TPE method was effective for the less complex Case 1, it faced challenges in Cases 2 and 3, particularly in achieving satisfactory accuracy across different data conditions. The findings for each case are summarized as follows:

\begin{itemize}
    \item \textbf{Case 1:} The HO approach using TPE yielded performance comparable to manual hyperparameter tuning, particularly for simulated data. However, manual tuning performed marginally better than TPE in simulated noisy and experimental data scenarios. The optimization history using TPE showed progressive improvement in loss values with iterations, though its performance with experimental data was inferior to that with simulated data.
    \item \textbf{Cases 2 and 3:} As the complexity of the problem increased, the performance of the PINN with HO using TPE showed significantly higher error. Manual tuning of hyperparameters outperformed the HO approach in terms of accuracy across different data conditions and experimental investigations. The optimization histories using TPE indicated limited improvement with iterations, as the cost function remained relatively unchanged.
\end{itemize}

\noindent The detailed numerical setup and the resulting comparative analysis of hyperparameter optimization are presented in the sections below.

\subsection{Numerical setup}\label{apdx:ho-results:mtd}

\noindent We conducted 200 trials using TPE, where each trial consisted of five random initializations of the neural network parameters (along with unknown parameters and self-adaptive weights). The target objective optimal parameters considered include the number of layers, the number of neurons per layer, the learning rates scheduler (for neural network weights, self-adaptive weights, and unknown parameters), and the activation function (evaluating hyperbolic tangent, rectified linear unit, and Swish). We also considered the initial value of the weights (for the loss functions) as adjustable parameters for each loss term for each equation of the model, the initial condition, and the data. This resulted in a total of \num{30} independent hyperparameters.

The loss function considered in the hyperparameter optimization process was the Mean Absolute Percentage Error (MAPE) between the predicted and true values of the measured states. The trial loss value was the average between the MAPE of the five random initializations. The MAPE is calculated as follows:
\begin{align}
\text{MAPE} = \frac{1}{N^{mape}} \sum_{i=1}^{N^{mape}} \left| \frac{y_s(t_i) - \bar{y}_s(t_i; \bm{\theta})}{y_s(t_i)} \right|,
\label{eq:ho:mape}
\end{align}
where $N^{mape}$ represents the total number of temporal data points. The data samples used for calculating the MAPE are identical to those employed for training the PINN, with a sampling interval of \(\Delta t = \qty{0.5}{\second}\). Here, $y_s(t_i)$ denotes the true value of state $s$ at time $t_i$, and $\bar{y}_s(t_i; \bm{\theta})$ denotes the predicted mean of state $s$ from the PINN.

The hyperparameter search ranges were determined manually based on the values obtained from the manual hyperparameter tuning. Then, we performed the HO process using TPE with \num{200} trials to obtain the optimal architecture and the other parameters, and we selected the best-performing parameters with the lowest MAPE. Thus, we conducted \num{30} random initialization using the selected hyperparameters (optimal using TPE) and performed the inverse problem using PINN for each case and data scenario. The results are compared with results from the manual tuning of hyperparameters and are discussed below.

\subsection{Numerical results}

\subsubsection{Inverse problem - Parameters estimation}\label{apdx:ho-results:inv:ps}

\noindent This section presents the results comparing the unknown parameters obtained for each case using TPE and manual tuning. The results indicate that the HO method using TPE successfully identified a set of hyperparameters with comparable accuracy to the manual approach, particularly in Case 1, shown in \cref{tab:ho:inv:ps:case1}. However, as discussed earlier, in the more complex scenarios of Case 2 (with seven unknown parameters) and Case 3 (with eight unknown parameters), the performance of the PINN using HO with TPE did not achieve the same level of accuracy as manual tuning. The results for Case 2 are shown in shown \cref{tab:ho:inv:ps:case2} and that of Case 3 are shown in \cref{tab:ho:inv:ps:case3}.

\begin{table}[H]

\sisetup{round-mode=places, round-precision=2}
\caption{ \textbf{Parameter estimation comparison for Case 1:} Comparative mean absolute percentage error (MAPE) with standard deviations (in parentheses) for the parameter estimates in Case 1, using PINN with manual hyperparameter tuning and PINN with hyperparameter tuning via TPE for the two experimental investigations.}
\centering
\begin{tabular}{llllll}
\toprule
  &              &    & \multicolumn{3}{c}{MAPE and std. ($\%$)} \\
  \cmidrule{4-6}
  Investigation & Data type & Method &                                        $B$ &                                                   $\mu$ &                                                   $\rho$ \\
\midrule
\multirow[t]{6}{*}{1} & \multirow[t]{2}{*}{Sim.} & Manual & \num{0.3123914334679204} (\num{0.19303666922700458}) & \num{0.03048262039075415} (\num{0.021972139301937466}) & \num{0.019583041849378462} (\num{0.012707899791126381}) \\
 &  & TPE & \num{5.536130605597964} (\num{1.9697784605792497}) & \num{1.6618036135166792} (\num{0.3996269951991529}) & \num{1.5576282637815402} (\num{0.36147945233595047}) \\
\cmidrule{2-6}
 & \multirow[t]{2}{*}{Noisy} & Manual & \num{0.5758474778631687} (\num{0.21045811311958046}) & \num{0.1246478464937965} (\num{0.0753653424894653}) & \num{0.09468425026741538} (\num{0.05260687862254404}) \\
 &  & TPE & \num{6.331758412213742} (\num{1.6236315351727324}) & \num{3.1989926632735526} (\num{1.1038380291101755}) & \num{3.4010301895757036} (\num{1.2628149383914}) \\
\cmidrule{2-6}
 & \multirow[t]{2}{*}{Exp.} & Manual & N/A & \num{1.802230806257376} (\num{0.4008577575960764}) & \num{0.7958879926832861} (\num{0.20873381758243328}) \\
 &  & TPE & N/A & \num{6.912597403670164} (\num{0.5577240205901203}) & \num{6.577383395680661} (\num{0.727792132635347}) \\
\cmidrule{1-6} \cmidrule{2-6}
\multirow[t]{6}{*}{2} & \multirow[t]{2}{*}{Sim.} & Manual & \num{0.6118124586223436} (\num{1.164159174886659}) & \num{0.3155249354959103} (\num{1.3287162986502057}) & \num{0.29123451847482407} (\num{1.2067855711457436}) \\
 &  & TPE & \num{5.20594792875318} (\num{1.5280526508704289}) & \num{0.14156749399559246} (\num{0.08223667130618315}) & \num{0.26438987521569035} (\num{0.16109442042773844}) \\
\cmidrule{2-6}
 & \multirow[t]{2}{*}{Noisy} & Manual & \num{0.48019770476369555} (\num{0.20170805277503845}) & \num{0.2143052567605567} (\num{0.6038020530427}) & \num{0.21391718172044438} (\num{0.5817864869230065}) \\
 &  & TPE & \num{4.939356824427481} (\num{1.6511414026879272}) & \num{0.2084327880721944} (\num{0.16622250071142752}) & \num{0.4348727022902847} (\num{0.3430288365039486}) \\
\cmidrule{2-6}
 & \multirow[t]{2}{*}{Exp.} & Manual & N/A & \num{4.302495983974393} (\num{0.26041427984745963}) & \num{2.3983552112728317} (\num{0.22275076534145866}) \\
 &  & TPE & N/A & \num{4.926654028532647} (\num{0.31847902392092436}) & \num{5.657700515066611} (\num{0.9360375039285032}) \\
\bottomrule
\end{tabular}

\label{tab:ho:inv:ps:case1}
\end{table}

\begin{table}[H]
\sisetup{round-mode=places, round-precision=1}
\setlength{\tabcolsep}{4pt}
\small
\caption{\textbf{Parameter estimation comparison for Case 2:} Comparative mean absolute percentage error (MAPE) with standard deviations (in parentheses) for the parameter estimates in Case 2, using PINN with manual hyperparameter tuning and PINN with hyperparameter tuning via TPE for the two experimental investigations.}
\centering
\begin{tabular}{lllccccccc}
\toprule
  &      &    & \multicolumn{7}{c}{MAPE and standard deviation ($\%$)} \\
  \cmidrule{4-10}
  Inv. & Data T. & Mtd. &                                                  $B$ &                                                $\mu$ &                                                $\rho$ &                                                $k_u$ &                                                $k_d$ &                                             $k_{3p}$ &                                             $k_{4p}$ \\
\midrule
\multirow[t]{6}{*}{1} & \multirow[t]{2}{*}{Sim.} & Manual & \num{2.5213836327651036} (\num{4.387909342162803}) & \num{13.102961533897536} (\num{27.19205369881111}) & \num{1.3105662092164576} (\num{2.720901664315951}) & \num{24.223342389290032} (\num{48.35094650639312}) & \num{6.151542597588386} (\num{13.244094835165122}) & \num{4.280495346543093} (\num{5.432141161902463}) & \num{3.80066181276497} (\num{7.2083173224836825}) \\
 &  & TPE & \num{6.543120936386771} (\num{0.48560046841397125}) & \num{225.2375002303746} (\num{36.99013573265702}) & \num{13.134903198824013} (\num{3.403511751896077}) & \num{271.7147895726595} (\num{111.25249977858067}) & \num{95.83732217208578} (\num{17.969348642870752}) & \num{41.275007922972094} (\num{4.971626378683817}) & \num{42.58206888726678} (\num{4.4773167415529045}) \\
\cmidrule{2-10}
 & \multirow[t]{2}{*}{Noisy} & Manual & \num{1.3429040186514873} (\num{1.7908819122742459}) & \num{36.42589635565102} (\num{39.41820724248472}) & \num{5.681855304227679} (\num{5.613470473518714}) & \num{40.57632313663549} (\num{40.72381315607422}) & \num{19.066711108882664} (\num{20.71153646571213}) & \num{9.788220859167755} (\num{9.2254125819255}) & \num{10.505141326103331} (\num{10.5251128539542}) \\
 &  & TPE & \num{6.605718473282441} (\num{0.33807970599650145}) & \num{239.68116120474707} (\num{27.832826461120156}) & \num{13.88472317464535} (\num{3.387687404256968}) & \num{242.82608004385355} (\num{123.92377314839877}) & \num{102.9719714363998} (\num{13.237229901531112}) & \num{43.66327075010575} (\num{3.4428615882196643}) & \num{44.536202620222504} (\num{3.0496902014086698}) \\
\cmidrule{2-10}
 & \multirow[t]{2}{*}{Exp.} & Manual & N/A & \num{120.18706717064303} (\num{28.996414837031224}) & \num{9.718875912541977} (\num{3.5541846452975263}) & \num{77.35400849627331} (\num{56.46106547734819}) & \num{56.30507022450421} (\num{10.65026318390637}) & \num{27.676174129055294} (\num{4.885772644947763}) & \num{29.262093899680988} (\num{4.42942275098544}) \\
 &  & TPE & N/A & \num{231.88369261195808} (\num{46.532063354726176}) & \num{10.08156503094187} (\num{7.288838075879531}) & \num{442.5220647226708} (\num{156.73000890778823}) & \num{96.7530634985958} (\num{20.82267645438385}) & \num{43.50148407241914} (\num{5.2066644594700655}) & \num{43.50949185375313} (\num{4.546284331703261}) \\
\cmidrule{1-10} \cmidrule{2-10}
\multirow[t]{6}{*}{2} & \multirow[t]{2}{*}{Sim.} & Manual & \num{0.9823345510017085} (\num{1.1995361381298189}) & \num{1.3793030201642582} (\num{2.477931388377458}) & \num{0.17681157226145608} (\num{0.1428153207295647}) & \num{0.2789054854944502} (\num{0.2691537002417715}) & \num{1.2046161021431805} (\num{2.1478117911958234}) & \num{3.9751433250910284} (\num{1.2686959833608065}) & \num{2.2655190369602156} (\num{1.2433632526814615}) \\
 &  & TPE & \num{6.662253760814251} (\num{0.3456814980545522}) & \num{146.43198840488384} (\num{19.26580107336151}) & \num{12.454819244490135} (\num{3.783698685219558}) & \num{72.59366349675292} (\num{54.72520625130038}) & \num{132.69759748252915} (\num{17.18216517023189}) & \num{44.59308251799127} (\num{3.7453558988046405}) & \num{46.119814005467056} (\num{3.0092531509183904}) \\
\cmidrule{2-10}
 & \multirow[t]{2}{*}{Noisy} & Manual & \num{1.2875478059261767} (\num{1.3055661331966981}) & \num{13.010727545257499} (\num{24.10933721489435}) & \num{3.1028807904373883} (\num{4.927883570221072}) & \num{3.624120697748016} (\num{7.052904086646576}) & \num{12.618720004236595} (\num{22.212623875796833}) & \num{6.448300466498422} (\num{6.497575101824763}) & \num{6.56768429052131} (\num{8.680814029054691}) \\
 &  & TPE & \num{6.563367409669211} (\num{0.38779987720510634}) & \num{134.5983315021704} (\num{17.29631479627239}) & \num{12.81071508359876} (\num{3.696324850838837}) & \num{61.87430526114965} (\num{44.27765953593691}) & \num{120.43313956942157} (\num{16.80009497589442}) & \num{42.776626691173824} (\num{5.193061241535699}) & \num{44.63763184115187} (\num{4.373296277729056}) \\
\cmidrule{2-10}
 & \multirow[t]{2}{*}{Exp.} & Manual & N/A & \num{14.49782816158458} (\num{20.93775780898413}) & \num{7.994798599578138} (\num{5.118943537830622}) & \num{4.818419157758497} (\num{4.3363164923481055}) & \num{15.393800325496779} (\num{14.845421842893636}) & \num{9.72735121273275} (\num{5.922143286957163}) & \num{5.734447186165114} (\num{7.987802584070322}) \\
 &  & TPE & N/A & \num{188.6101489177803} (\num{43.603023752991}) & \num{14.684426138902571} (\num{9.723826308247311}) & \num{88.90694181588424} (\num{60.4600861223141}) & \num{169.6791017627438} (\num{38.986282056273694}) & \num{53.01954516949464} (\num{7.262831329495235}) & \num{50.629596907063295} (\num{5.144158125301205}) \\
\bottomrule
\end{tabular}

\label{tab:ho:inv:ps:case2}
\end{table}

\begin{table}[H]
\sisetup{round-mode=places, round-precision=1}
\setlength{\tabcolsep}{4pt}
\small
\caption{\textbf{Parameter estimation comparison for Case 3:} Comparative mean absolute percentage error (MAPE) with standard deviations (in parentheses) for the parameter estimates in Case 3, using PINN with manual hyperparameter tuning and PINN with hyperparameter tuning via TPE for the two experimental investigations.}
\centering
\begin{tabular}{lllcccccccc}
\toprule
  &      &    & \multicolumn{8}{c}{MAPE and standard deviation ($\%$)} \\
  \cmidrule{4-11}
  Inv. & Data T. & Mtd. &                                                   $B$ &                                                $\mu$ &                                               $\rho$ &                                                $k_u$ &                                                $k_d$ &                                             $k_{4p}$ &                                             $k_{1s}$ &                                             $k_{5s}$ \\
\midrule
\multirow[t]{6}{*}{1} & \multirow[t]{2}{*}{Sim.} & Manual & \num{14.948393812003522} (\num{0.1391289229468987}) & \num{11.307134906412633} (\num{7.293375557134734}) & \num{3.024805096842583} (\num{2.0768225473905195}) & \num{26.26571308093323} (\num{15.696354879782545}) & \num{3.2450799503453087} (\num{2.745995319765503}) & \num{1.7870117025733796} (\num{1.7088466565532832}) & \num{6.915994745528418} (\num{4.549545752180401}) & \num{12.238812343580292} (\num{11.161682685434615}) \\
 &  & TPE & \num{14.936683155216286} (\num{0.004353943887392071}) & \num{16.611755689367314} (\num{6.36521984038623}) & \num{3.9759038409164673} (\num{1.850270068637068}) & \num{35.570670654307456} (\num{13.672362190069533}) & \num{13.53688071837192} (\num{4.442608107832205}) & \num{11.537700915043084} (\num{4.240283094150259}) & \num{32.09494877295431} (\num{5.934697032917006}) & \num{31.98065606776775} (\num{7.23231658989074}) \\
\cmidrule{2-11}
 & \multirow[t]{2}{*}{Noisy} & Manual & \num{14.506395405134997} (\num{2.694645461986955}) & \num{10.410154579252955} (\num{5.446287735472595}) & \num{1.8768204707887575} (\num{1.4279313717772126}) & \num{25.96860009670603} (\num{14.273776146505766}) & \num{4.988127891202737} (\num{3.0763038908609475}) & \num{2.057263822405098} (\num{1.2366811858478661}) & \num{6.878481075721019} (\num{4.4072746908221205}) & \num{6.000178140461784} (\num{5.17994719438864}) \\
 &  & TPE & \num{14.937261923664122} (\num{0.004293207172111158}) & \num{14.71416447087154} (\num{6.313726058133057}) & \num{4.1525453484006905} (\num{2.1725562686669098}) & \num{39.73605604477954} (\num{16.488702427324917}) & \num{12.700294935368941} (\num{4.774895106019818}) & \num{10.733004789713526} (\num{4.157840562609105}) & \num{31.86367268394108} (\num{8.308170857646004}) & \num{26.664509946285055} (\num{10.292731825596123}) \\
\cmidrule{2-11}
 & \multirow[t]{2}{*}{Exp.} & Manual & N/A & \num{13.052403078428917} (\num{6.413757522137849}) & \num{9.255173033011172} (\num{2.1193718172997436}) & \num{34.50834145300853} (\num{15.817923386210257}) & \num{16.00668914180382} (\num{2.009829555813336}) & \num{5.87474192124567} (\num{1.1945235473269455}) & \num{12.3686559173505} (\num{6.0897642254101125}) & \num{7.759101924557011} (\num{3.6060770944265523}) \\
 &  & TPE & N/A & \num{8.251773887905156} (\num{3.060611447332285}) & \num{5.0573090352690455} (\num{1.499279496134628}) & \num{32.99897516595099} (\num{18.430773404015646}) & \num{8.88674851804518} (\num{2.3502817122998403}) & \num{8.79384074996396} (\num{3.387748139060911}) & \num{28.60780547504564} (\num{12.15856733143574}) & \num{6.329095675638884} (\num{7.440430259476702}) \\
\cmidrule{1-11} \cmidrule{2-11}
\multirow[t]{6}{*}{2} & \multirow[t]{2}{*}{Sim.} & Manual & \num{14.999129436145031} (\num{0.004026205563665935}) & \num{14.336275224657454} (\num{14.3134550186948}) & \num{4.119317099646765} (\num{2.8218806983709612}) & \num{6.900439504079831} (\num{8.05322948904196}) & \num{6.655014737929068} (\num{7.667611831514188}) & \num{3.4281873280701864} (\num{3.494575683664538}) & \num{12.706454478948592} (\num{11.072490429870568}) & \num{6.96440678514521} (\num{3.929910827578125}) \\
 &  & TPE & \num{14.936148356234098} (\num{0.00535115786244873}) & \num{15.371066592260469} (\num{6.031405355348095}) & \num{6.298855801547605} (\num{3.07450146597834}) & \num{31.27042046910926} (\num{12.800720186501598}) & \num{12.544651468084899} (\num{9.056482973840327}) & \num{18.680809479917535} (\num{7.994136350733993}) & \num{30.226634778493995} (\num{11.629407034006837}) & \num{35.799051926027346} (\num{8.9440474873023}) \\
\cmidrule{2-11}
 & \multirow[t]{2}{*}{Noisy} & Manual & \num{14.99978715592548} (\num{0.0007848589279220654}) & \num{13.658673159887488} (\num{8.003053918764916}) & \num{3.7653651802310963} (\num{3.0649208275370228}) & \num{4.91554888888324} (\num{8.583009052116708}) & \num{13.63428123347513} (\num{8.10408638898512}) & \num{7.042093176433951} (\num{4.000652476670067}) & \num{11.863419587322072} (\num{9.215161224716065}) & \num{12.215725252776759} (\num{9.053991032948176}) \\
 &  & TPE & \num{14.936791938931298} (\num{0.0038533388811222377}) & \num{12.028126859155668} (\num{4.941322789056808}) & \num{7.362889905207272} (\num{3.923859706107914}) & \num{31.740328807299047} (\num{16.162096082217875}) & \num{9.805174405278013} (\num{6.325473360409552}) & \num{20.349568086261517} (\num{6.170486534603317}) & \num{22.303097462895092} (\num{13.779618662353727}) & \num{30.762390983612924} (\num{12.58364774912856}) \\
\cmidrule{2-11}
 & \multirow[t]{2}{*}{Exp.} & Manual & N/A & \num{1.5967748423505703} (\num{1.2239031997119865}) & \num{10.158213110725832} (\num{1.2534241241532886}) & \num{1.9600652119559312} (\num{0.5339941889585855}) & \num{12.052400641918984} (\num{2.0248664840645176}) & \num{5.281842999816811} (\num{0.8925806099063434}) & \num{7.00733773341248} (\num{3.3526932721643163}) & \num{9.246945308406612} (\num{3.0564103097400515}) \\
 &  & TPE & N/A & \num{8.731925226366293} (\num{5.121974889225708}) & \num{4.873456073548835} (\num{1.7324094175373113}) & \num{13.448776571012546} (\num{13.668169961082105}) & \num{10.521898462609748} (\num{2.620843310928286}) & \num{6.519349982447037} (\num{4.447885098862208}) & \num{15.479705862410885} (\num{15.612485937257851}) & \num{18.40692101586912} (\num{11.04698820639792}) \\
\bottomrule
\end{tabular}

\label{tab:ho:inv:ps:case3}
\end{table}

\subsubsection{Inverse problem - States estimation}\label{apdx:ho-results:inv:states}

\noindent This section presents the results comparing the state variables obtained for each case of Investigation 2 using TPE and manual tuning. We used MAPE (\cref{eq:state:mape}) as an error metric. Similar to the parameters estimation, the results for the states also indicate that the HO with TPE  successfully identified a set of hyperparameters with comparable accuracy to the manual approach, particularly in Case 1 with simulated data, as shown in \cref{tab:ho:inv:state:sim}. However, for the other data sources simulated with added Gaussian noise, shown in \cref{tab:ho:inv:state:noi}, and for experimental data, shown in \cref{tab:ho:inv:state:exp}, the PINN using HO with TPE did not achieve the same level of accuracy as the manual tuning.

\begin{table}[H]

\sisetup{round-mode=places, round-precision=3}
\caption{\textbf{State estimation comparison for simulated data:} Comparison of mean absolute percentage error in the predicted state variables when predicted using PINN with manual hyperparameter tuning and PINN with hyperparameter tuning with TPE for simulated data.}
\centering
\begin{tabular}{llllllll}
\toprule
 & Method & $P_1$ & $P_2$ & $Q_1$ & $Q_2$ & $Q_p$ & $\omega$ \\
\midrule
\multirow[t]{2}{*}{Case 1} & Manual & \qty{0.29631732549629414}{\percent} & \qty{0.007272611329997738}{\percent} & \qty{0.004299792600604389}{\percent} & \qty{0.005461655850932174}{\percent} & \qty{0.0067292711009511895}{\percent} & \qty{0.03598979695997091}{\percent} \\
 & TPE & \qty{0.03100025261872425}{\percent} & \qty{0.0008079034659301367}{\percent} & \qty{0.016605940476137716}{\percent} & \qty{0.02372173857562553}{\percent} & \qty{0.025457803585392166}{\percent} & \qty{0.057779151306578076}{\percent} \\
\cmidrule{1-8}
\multirow[t]{2}{*}{Case 2} & Manual & \qty{0.1288012343173085}{\percent} & \qty{0.007380847118950712}{\percent} & \qty{0.013398790927977982}{\percent} & \qty{0.013614996113345683}{\percent} & \qty{0.013454745265758808}{\percent} & \qty{0.12058742146603693}{\percent} \\
 & TPE & \qty{15.242451917946736}{\percent} & \qty{0.06759053310864371}{\percent} & \qty{2.811798960744442}{\percent} & \qty{2.8059717214555437}{\percent} & \qty{2.80605437543181}{\percent} & \qty{9.769511939787611}{\percent} \\
\cmidrule{1-8}
\multirow[t]{2}{*}{Case 3} & Manual & \qty{0.4326674963014063}{\percent} & \qty{0.008941093954183105}{\percent} & \qty{0.5002355077151027}{\percent} & \qty{0.5100637282502447}{\percent} & \qty{0.5047619957165506}{\percent} & \qty{2.0799682198773475}{\percent} \\
 & TPE & \qty{10.108700391054217}{\percent} & \qty{0.05730276314315529}{\percent} & \qty{2.765225718328779}{\percent} & \qty{2.759045004523327}{\percent} & \qty{2.7641227362334453}{\percent} & \qty{7.335496964130073}{\percent} \\
\bottomrule
\end{tabular}

\label{tab:ho:inv:state:sim}
\end{table}

\begin{table}[H]
\sisetup{round-mode=places, round-precision=3}
\caption{\textbf{State estimation comparison for simulated data with added noise:} Comparison of mean absolute percentage error in the predicted state variables when predicted using PINN with manual hyperparameter tuning and PINN with hyperparameter tuning  with TPE for simulated data with added noise.}
\centering
\begin{tabular}{llllllll}
\toprule
 & Method & $P_1$ & $P_2$ & $Q_1$ & $Q_2$ & $Q_p$ & $\omega$ \\
\midrule
\multirow[t]{2}{*}{Case 1} & Manual & \qty{0.27317262460427566}{\percent} & \qty{0.11733860600207056}{\percent} & \qty{0.009490576577285604}{\percent} & \qty{0.010592401058418091}{\percent} & \qty{0.01118455688588485}{\percent} & \qty{0.023573964525749794}{\percent} \\
 & TPE & \qty{0.18654384669389873}{\percent} & \qty{0.11055622137405617}{\percent} & \qty{0.01316811676087331}{\percent} & \qty{0.018829064242508988}{\percent} & \qty{0.018545749454866752}{\percent} & \qty{0.0406808275389825}{\percent} \\
\cmidrule{1-8}
\multirow[t]{2}{*}{Case 2} & Manual & \qty{0.2688615466545799}{\percent} & \qty{0.11670623061216924}{\percent} & \qty{0.029428832866138996}{\percent} & \qty{0.026569915576298307}{\percent} & \qty{0.029456166994718223}{\percent} & \qty{1.5865982986949283}{\percent} \\
 & TPE & \qty{12.272048243759476}{\percent} & \qty{0.18848542101124002}{\percent} & \qty{2.2010822519446265}{\percent} & \qty{2.196033183784638}{\percent} & \qty{2.196069318660018}{\percent} & \qty{8.797847913244148}{\percent} \\
\cmidrule{1-8}
\multirow[t]{2}{*}{Case 3} & Manual & \qty{0.5958818838616401}{\percent} & \qty{0.12111597619960086}{\percent} & \qty{0.07713426731390464}{\percent} & \qty{0.07964753987023127}{\percent} & \qty{0.08079546971942186}{\percent} & \qty{1.0066037499004197}{\percent} \\
 & TPE & \qty{17.745106844956698}{\percent} & \qty{0.1469086524731162}{\percent} & \qty{2.2396730228968385}{\percent} & \qty{2.237886998829297}{\percent} & \qty{2.2377834986825063}{\percent} & \qty{6.1469029756163005}{\percent} \\
\bottomrule
\end{tabular}

\label{tab:ho:inv:state:noi}
\end{table}

\begin{table}[H]
\sisetup{round-mode=places, round-precision=3}
\caption{\textbf{State estimation comparison for experimental data:} Comparison of mean absolute percentage error in the predicted state variables when predicted using PINN with manual hyperparameter tuning and PINN with hyperparameter tuning with TPE for experimental data.}
\centering
\begin{tabular}{llllll}
\toprule
 & Method & $P_1$ & $P_2$ & $Q_1$ & $\omega$ \\
\midrule
\multirow[t]{2}{*}{Case 1} & Manual & \qty{6.957757107190568}{\percent} & \qty{0.27105642131521374}{\percent} & \qty{0.10102834310170199}{\percent} & \qty{0.6731478323222528}{\percent} \\
 & TPE & \qty{1.2554549639202768}{\percent} & \qty{0.3550938408377011}{\percent} & \qty{5.347594478778005}{\percent} & \qty{4.51472340723653}{\percent} \\
\cmidrule{1-6}
\multirow[t]{2}{*}{Case 2} & Manual & \qty{7.375238970746743}{\percent} & \qty{0.2641143585773164}{\percent} & \qty{0.32217517776100907}{\percent} & \qty{0.7243781001194229}{\percent} \\
 & TPE & \qty{22.283341435250172}{\percent} & \qty{0.4907376422220645}{\percent} & \qty{0.9460945760796633}{\percent} & \qty{10.983408520621108}{\percent} \\
\cmidrule{1-6}
\multirow[t]{2}{*}{Case 3} & Manual & \qty{5.438674287229841}{\percent} & \qty{0.2667440457248636}{\percent} & \qty{0.1259121808825852}{\percent} & \qty{2.154050825121562}{\percent} \\
 & TPE & \qty{8.604719603375075}{\percent} & \qty{0.2919540338451358}{\percent} & \qty{3.227378159213419}{\percent} & \qty{5.703385148285287}{\percent} \\
\bottomrule
\end{tabular}

\label{tab:ho:inv:state:exp}
\end{table}

\subsubsection{Optimization history}\label{apdx:ho-results:ho:history}

\begin{figure}[H]
\centering
\begin{subfigure}[b]{0.475\textwidth}
  \centering
  \includegraphics{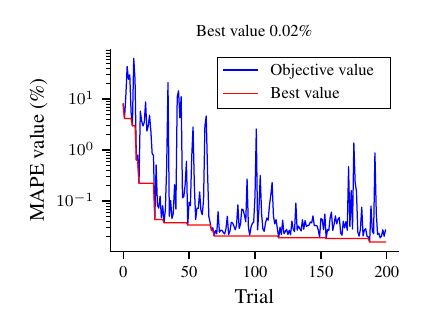}
  \caption{Simulated data.}
  \label{fig:ho:case1:sim}
\end{subfigure}
\hfil
\begin{subfigure}[b]{0.475\textwidth}
  \centering
  \includegraphics[width=\textwidth, keepaspectratio]{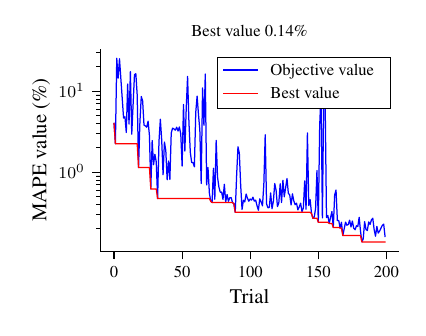}
  \caption{Experimental data.}
  \label{fig:ho:case1:exp}
\end{subfigure}
\hfil
\caption{Optimization history for Case 1.}
\label{fig:ho:case1}
\end{figure}

\begin{figure}[H]
\centering
\begin{subfigure}[b]{0.475\textwidth}
  \centering
  \includegraphics[width=\textwidth, keepaspectratio]{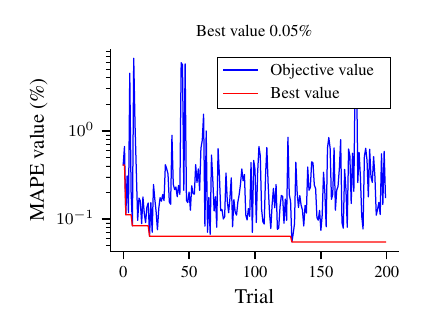}
  \caption{Simulated data.}
  \label{fig:ho:case2:sim}
\end{subfigure}
\hfil
\begin{subfigure}[b]{0.475\textwidth}
  \centering
  \includegraphics[width=\textwidth, keepaspectratio]{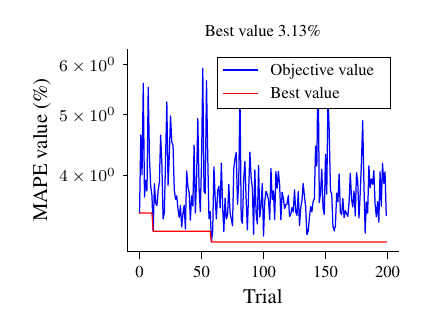}
  \caption{Experimental data.}
  \label{fig:ho:case2:exp}
\end{subfigure}
\hfil
\caption{Optimization history for Case 2.}
\label{fig:ho:case2}
\end{figure}

\begin{figure}[H]
\centering
\begin{subfigure}[b]{0.475\textwidth}
  \centering
  \includegraphics[width=\textwidth, keepaspectratio]{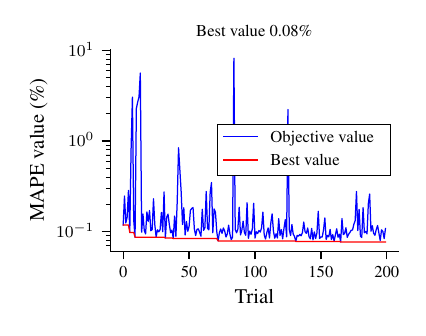}
  \caption{Simulated data.}
  \label{fig:ho:case3:sim}
\end{subfigure}
\hfil
\begin{subfigure}[b]{0.475\textwidth}
  \centering
  \includegraphics[width=\textwidth, keepaspectratio]{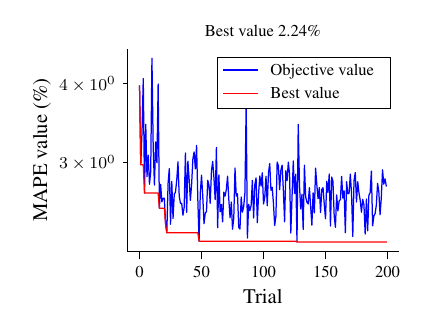}
  \caption{Experimental data.}
  \label{fig:ho:case3:exp}
\end{subfigure}
\hfil
\caption{Optimization history for Case 3.}
\label{fig:ho:case3}
\end{figure}

\noindent In the previous section, we observed that the HO using TPE performed well in Case 1; however, it did not in Case 2 and Case 3. In this section, we present the optimization history of the HO MAPE value across the different trials using TPE. The optimization history for the different cases is shown in \cref{fig:ho:case1} for Case 1, \cref{fig:ho:case2} for Case 2, and \cref{fig:ho:case3} and Case 3, respectively.

From the \cref{fig:ho:case1}, we observed that for Case 1, the optimization history using TPE shows that HO progressively improved loss values with iteration, though its performance with experimental data was inferior to that with simulated data. Thus, we observe higher errors in the experimental data, as discussed in the previous section. For Case 2, presented in \cref{fig:ho:case2}, and Case 3, shown in \cref{fig:ho:case3}, the optimization histories using TPE indicate that HO does not improve with iteration, as the cost function remains the same with marginal improvement. Thus, we observed higher errors in the parameter and state estimation, as discussed in the previous section.

\section{PINN additional results}\label{apdx:pinn_results}

\subsection{Investigation 1}
In this section, we present additional results pertaining to the state estimation for the PINN. These results are included for the sake of completeness and have been previously discussed in \cref{subsec:states}.

\subsubsection{State estimation results for simulated data}

\begin{figure}[H]
  \centering
  \includegraphics{./figs/results/exp-1/fw_sim_lgd.pdf}\\
  \begin{subfigure}[b]{0.47\textwidth}
    \centering
    \includegraphics{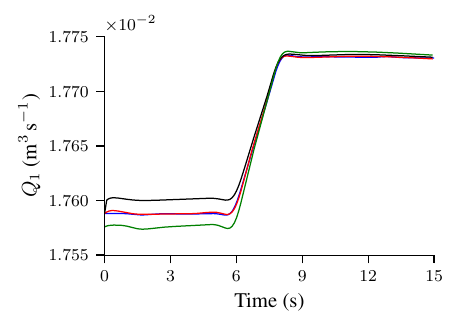}
    \caption{Upstream volumetric flow rate.}
    \label{fig:sim1_time:q1}
  \end{subfigure}
  \begin{subfigure}[b]{0.47\textwidth}
    \centering
    \includegraphics{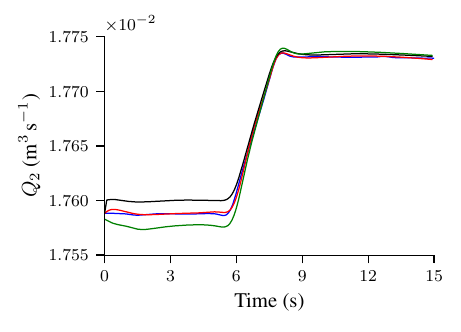}
    \caption{Downstream volumetric flow rate.}
    \label{fig:sim1_time:q2}
  \end{subfigure}\\
  \caption{\textbf{Additional predicted states for the simulated and unknown flow parameters for the first investigation.} Across all cases, the training dataset comprises \num{30} data points and by \num{100} collocation points.}
  \label{fig:sim1_time:additional}
\end{figure}

\subsubsection{State estimation results for simulated data with noise}

\begin{figure}[H]
  \centering
  \includegraphics{./figs/results/exp-1/fw_sim_noi_lgd.pdf}\\
  \begin{subfigure}[b]{0.47\textwidth}
    \centering
    \includegraphics{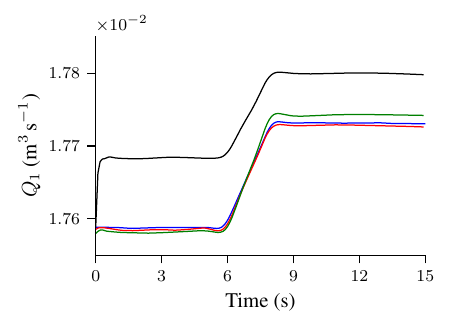}
    \caption{Upstream volumetric flow rate.}
    \label{fig:sim1_noi_time:q1}
  \end{subfigure}
  \begin{subfigure}[b]{0.47\textwidth}
    \centering
    \includegraphics{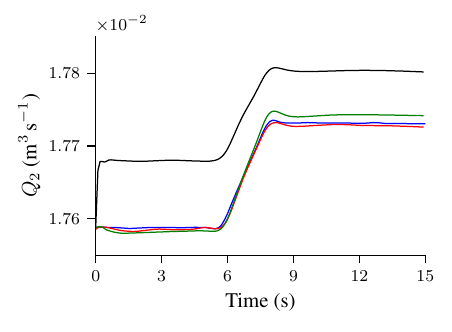}
    \caption{Downstream volumetric flow rate.}
    \label{fig:sim1_noi_time:q2}
  \end{subfigure}
  \caption{\textbf{Additional predicted states for the simulated with noise and unknown flow parameters for the first investigation.} Across all cases, the training dataset comprises \num{37} data points and by \num{100} collocation points.}
  \label{fig:sim1_noi_time:additional}
\end{figure}

\subsection{Investigation 2}
In this section, we present the results pertaining to the state estimation for the PINN with the second investigation. It is important to note, that these results are included for the sake of completeness and have been previously discussed in \cref{subsec:states}.
\subsubsection{State estimation results for simulated data}
\begin{table}[H]
  \sisetup{round-mode = places, round-precision = 3}
  \small
  \caption{\textbf{Mean absolute percentage error for the simulated case.}}
  \centering
  \begin{tabular}{lllllll}
    \toprule
     & $P_1$ & $P_2$ & $Q_1$ & $Q_2$ & $Q_p$ & $\omega$ \\
     \midrule
     Case 1 & \qty{0.2953220179088705}{\percent} & \qty{0.007201254721717595}{\percent} & \qty{0.00431684080736977}{\percent} & \qty{0.005452382131029972}{\percent} & \qty{0.006651820766369669}{\percent} & \qty{0.03676927198124851}{\percent} \\
     Case 2 & \qty{0.12869831870500634}{\percent} & \qty{0.00718160065522572}{\percent} & \qty{0.013717160927496558}{\percent} & \qty{0.013907312532191173}{\percent} & \qty{0.01366268668163769}{\percent} & \qty{0.12100436005928719}{\percent} \\
     Case 3 & \qty{0.4368526992191205}{\percent} & \qty{0.008534703473467865}{\percent} & \qty{0.5091731785957018}{\percent} & \qty{0.5187950994773766}{\percent} & \qty{0.5137306435783965}{\percent} & \qty{2.113168937135578}{\percent} \\
     \bottomrule
    \end{tabular}
    \label{tab:mae_sim3}
\end{table}

\begin{figure}[H]
  \centering
  \includegraphics{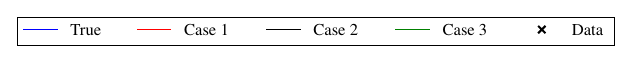}\\
  \begin{subfigure}[b]{0.47\textwidth}
    \centering
    \includegraphics{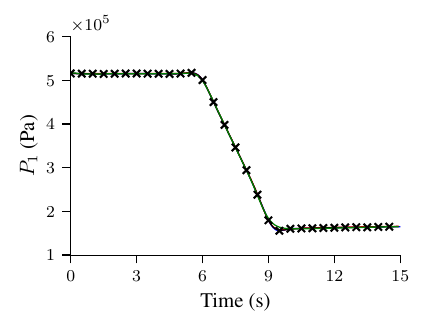}
    \caption{Intake pressure.}
    \label{fig:sim3_time:p1}
  \end{subfigure}
  \begin{subfigure}[b]{0.47\textwidth}
    \centering
    \includegraphics{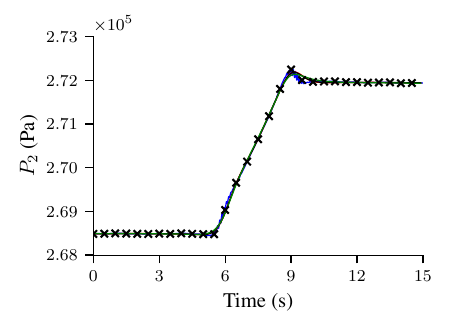}
    \caption{Discharge pressure.}
    \label{fig:sim3_time:p2}
  \end{subfigure}\\
  \begin{subfigure}[b]{0.47\textwidth}
    \centering
    \includegraphics{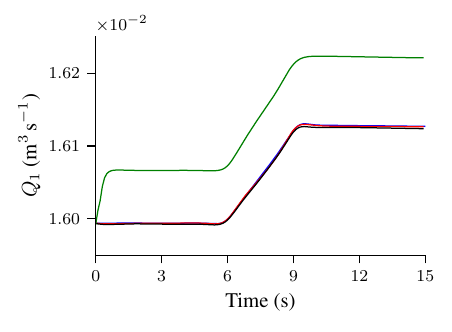}
    \caption{Upstream volumetric flow rate.}
    \label{fig:sim3_time:q1}
  \end{subfigure}
  \begin{subfigure}[b]{0.47\textwidth}
    \centering
    \includegraphics{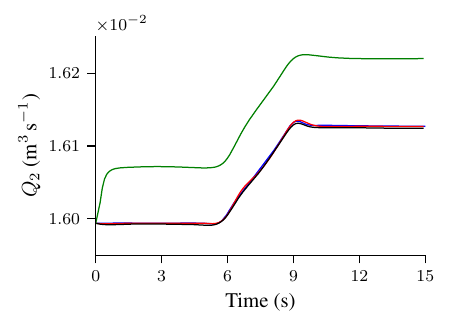}
    \caption{Downstream volumetric flow rate.}
    \label{fig:sim3_time:q2}
  \end{subfigure}\\
  \begin{subfigure}[b]{0.47\textwidth}
    \centering
    \includegraphics{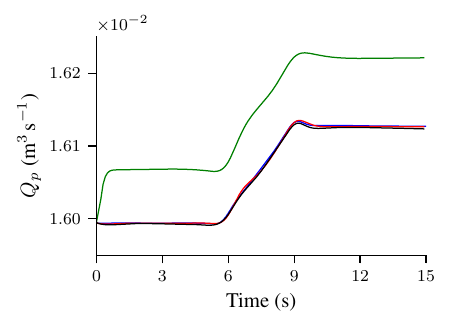}
    \caption{Impeller volumetric flow rate.}
    \label{fig:sim3_time:qi}
  \end{subfigure}
  \begin{subfigure}[b]{0.47\textwidth}
    \centering
    \includegraphics{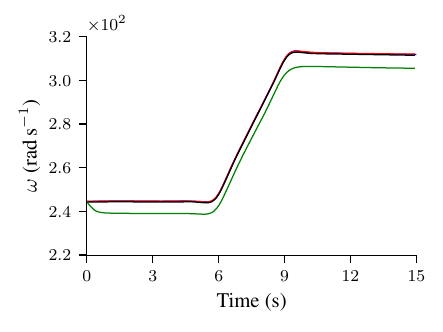}
    \caption{ESP angular velocity.}
    \label{fig:sim3_time:omega}
  \end{subfigure}\\
  \caption{\textbf{Predicted states for the simulated and unknown flow parameters for the second investigation.} Across all cases, the training dataset comprises \num{30} data points and by \num{100} collocation points.}
  \label{fig:sim3_time}
\end{figure}

\subsubsection{State estimation results for simulated data with noise}

\begin{table}[H]
  \small
  \sisetup{round-mode = places, round-precision = 3}
  \caption{\textbf{Mean absolute percentage error for the simulation with added noise case.}}
  \centering
  \begin{tabular}{lllllll}
    \toprule
     & $P_1$ & $P_2$ & $Q_1$ & $Q_2$ & $Q_p$ & $\omega$ \\
     \midrule
     Case 1 & \qty{0.2599347073369137}{\percent} & \qty{0.029908423147014734}{\percent} & \qty{0.009542642898642865}{\percent} & \qty{0.010517174084855772}{\percent} & \qty{0.010703753491727672}{\percent} & \qty{0.02554045837985367}{\percent} \\
     Case 2 & \qty{0.27755142957239837}{\percent} & \qty{0.026457354013453376}{\percent} & \qty{0.030153125673741597}{\percent} & \qty{0.027388373323869664}{\percent} & \qty{0.0304392975682618}{\percent} & \qty{1.6219662070529945}{\percent} \\
     Case 3 & \qty{0.5443767228377588}{\percent} & \qty{0.029139405665778436}{\percent} & \qty{0.0755574988515187}{\percent} & \qty{0.07811259323416116}{\percent} & \qty{0.07896161871581812}{\percent} & \qty{1.0122102158323114}{\percent} \\
     \bottomrule
    \end{tabular}
    \label{tab:mae_sim_noi3}
\end{table}

\begin{figure}[H]
  \centering
  \includegraphics{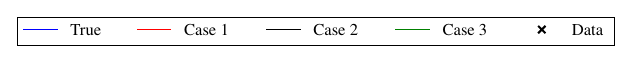}\\
  \begin{subfigure}[b]{0.47\textwidth}
    \centering
    \includegraphics{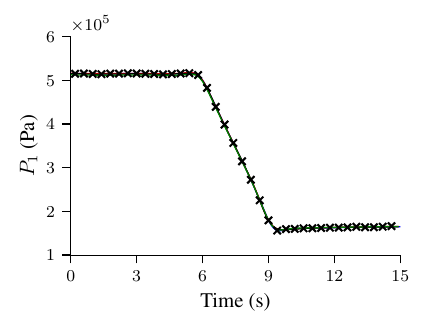}
    \caption{Intake pressure.}
    \label{fig:sim3_noi_time:p1}
  \end{subfigure}
  \begin{subfigure}[b]{0.47\textwidth}
    \centering
    \includegraphics{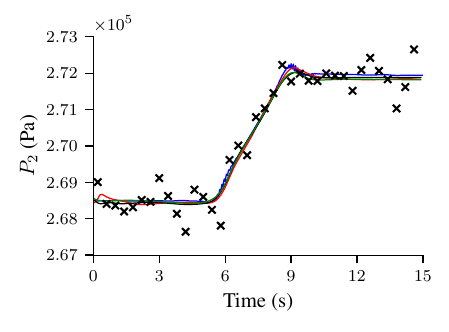}
    \caption{Discharge pressure.}
    \label{fig:sim3_noi_time:p2}
  \end{subfigure}\\
  \begin{subfigure}[b]{0.47\textwidth}
    \centering
    \includegraphics{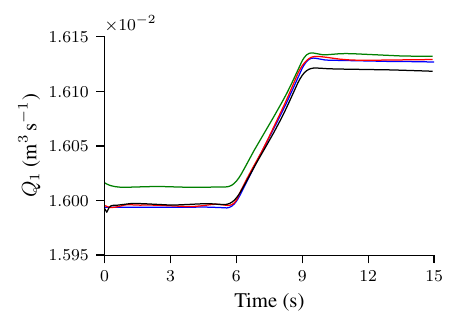}
    \caption{Upstream volumetric flow rate.}
    \label{fig:sim3_noi_time:q1}
  \end{subfigure}
  \begin{subfigure}[b]{0.47\textwidth}
    \centering
    \includegraphics{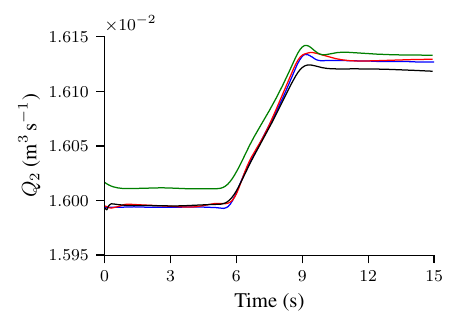}
    \caption{Downstream volumetric flow rate.}
    \label{fig:sim3_noi_time:q2}
  \end{subfigure}\\
  \begin{subfigure}[b]{0.47\textwidth}
    \centering
    \includegraphics{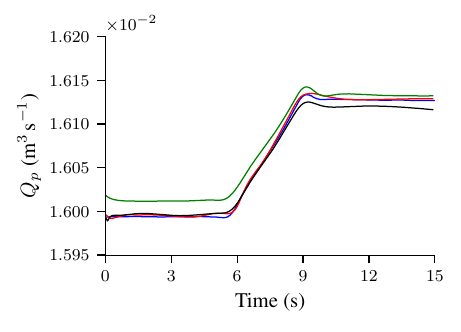}
    \caption{Impeller volumetric flow rate.}
    \label{fig:sim3_noi_time:qi}
  \end{subfigure}
  \begin{subfigure}[b]{0.47\textwidth}
    \centering
    \includegraphics{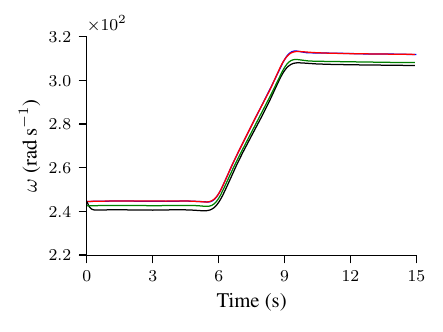}
    \caption{ESP angular velocity.}
    \label{fig:sim3_noi_time:omega}
  \end{subfigure}\\
  \caption{\textbf{Predicted states for the simulated with noise and unknown flow parameters for the second investigation.} Across all cases, the training dataset comprises \num{37} data points and by \num{100} collocation points.}
  \label{fig:sim3_noi_time}
\end{figure}

\break

\subsubsection{State estimation results for experimental data}

\begin{table}[H]
  \small
  \sisetup{round-mode = places, round-precision = 3}
  \caption{\textbf{Mean absolute percentage error for the experimental case.}}
  \centering
  \begin{tabular}{lllll}
    \toprule
     & $P_1$ & $P_2$ & $Q_1$ & $\omega$ \\
     \midrule
     Case 1 & \qty{6.978218580436645}{\percent} & \qty{0.26434722951053674}{\percent} & \qty{0.09047458390235757}{\percent} & \qty{0.6626513291829126}{\percent} \\
     Case 2 & \qty{7.391333742267528}{\percent} & \qty{0.2532021274071301}{\percent} & \qty{0.33600542774150305}{\percent} & \qty{0.7506326322386131}{\percent} \\
     Case 3 & \qty{5.460018790378688}{\percent} & \qty{0.25160447841658956}{\percent} & \qty{0.1228899789251732}{\percent} & \qty{2.1666920722604632}{\percent} \\
     \bottomrule                    
    \end{tabular}
    \label{tab:mae_exp3}
\end{table}

\begin{figure}[H]
  \centering
  \includegraphics{./figs/results/exp-1/fw_exp_lgd.pdf}\\
  \begin{subfigure}[b]{0.47\textwidth}
    \centering
    \includegraphics[width=\textwidth, keepaspectratio]{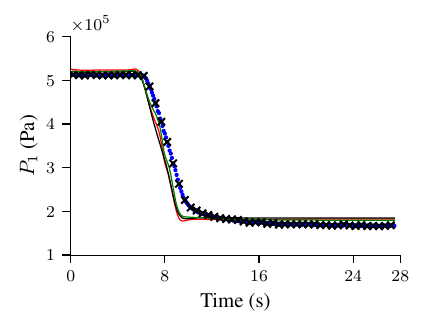}
    \caption{Intake pressure.}
    \label{fig:exp3_time:p1}
  \end{subfigure}
  \begin{subfigure}[b]{0.47\textwidth}
    \centering
    \includegraphics[width=\textwidth, keepaspectratio]{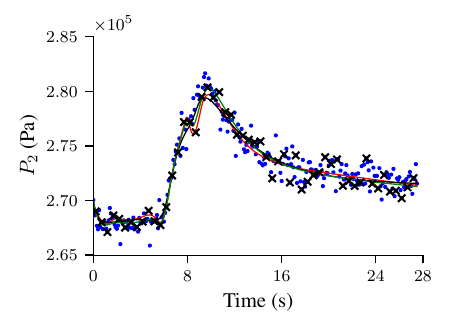}
    \caption{Discharge pressure.}
    \label{fig:exp_3time:p2}
  \end{subfigure}\\
  \begin{subfigure}[b]{0.47\textwidth}
    \centering
    \includegraphics[width=\textwidth, keepaspectratio]{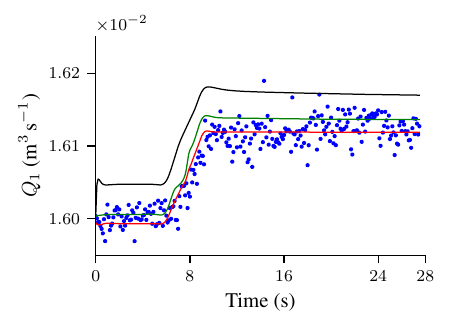}
    \caption{Upstream flow rate.}
    \label{fig:exp3_time:q1}
  \end{subfigure}
  \begin{subfigure}[b]{0.47\textwidth}
    \centering
    \includegraphics[width=\textwidth, keepaspectratio]{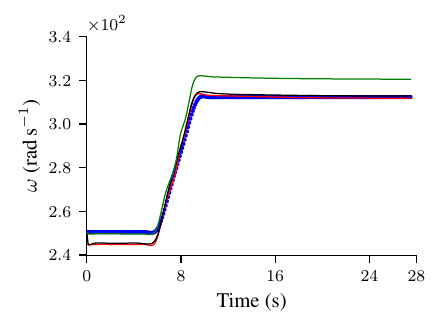}
    \caption{ESP angular velocity.}
    \label{fig:exp3_time:omega}
  \end{subfigure}\\
  \caption{\textbf{Predicted states for the experimental data and unknown flow parameters for the second investigation.} Across all cases, the training dataset comprises \num{55} data points and by \num{100} collocation points.}
  \label{fig:exp3_time}
\end{figure}

\end{document}